%% file: sn-article.tex
\documentclass[pdflatex,sn-mathphys-num]{sn-jnl}

\usepackage{CJKutf8}
\usepackage{graphicx}%
\usepackage{multirow}%
\usepackage{amsmath,amssymb,amsfonts}%
\usepackage{amsthm}%
\usepackage{mathrsfs}%
\usepackage[title]{appendix}%
\usepackage{xcolor}%
\usepackage{textcomp}%
\usepackage{manyfoot}%
\usepackage{booktabs}%
\usepackage{algorithm}%
\usepackage{rotating}
\usepackage{algorithmicx}%
\usepackage{algpseudocode}%
\usepackage{listings}%
\usepackage{booktabs}
\usepackage{graphicx}
\usepackage{booktabs}
\usepackage{lineno}


\theoremstyle{thmstyleone}

\theoremstyle{thmstyletwo}%
\usepackage{booktabs}

\theoremstyle{thmstylethree}%

\begin{document}

\begin{CJK*}{UTF8}{gbsn}

\title[Article Title]{Knowledge-structured and evidence-aware semantic self-distillation for glomerular pathology foundation model}

\author[1,6]{Qiming He+*}
\author[2]{Jing Li+}
\author[3]{Zihan Wang+}
\author[3]{Tian Guan+}
\author[4]{Yifei Ma}
\author[4]{Zimo Zhao}
\author[2]{Yanxia Wang}
\author[7]{Hongjing Chen}
\author[8]{Yexing Zhang}
\author[3]{Yizhi Wang}
\author[3]{Xinrui Chen}
\author[3]{Xitong Ling}
\author[3]{Minxi Ouyang}
\author[3]{Yujie Xie}
\author[3]{Lianghui Zhu}
\author[3]{Yiqing Liu}
\author[2]{Qingxia Hou}
\author[2]{Shuyan Zhao}
\author[2]{Xiaoqin Wang}
\author[2]{Lili Ma}
\author[3]{Qiang Huang}
\author[3]{Zhiyuan Shen}
\author[3]{Junru Cheng}
\author[3]{Siqi Zeng}
\author[4]{Chao He*}
\author[3,5,6]{Yonghong He*}
\author[2,5]{Zhe Wang*}

\affil*[1]{Fuzhou University Affiliated Provincial Hospital, Interdisciplinary Institute for Medical Engineering, Fuzhou University, Fuzhou, China}

\affil*[2]{Department of Pathology, School of Basic Medicine and Xijing Hospital, Fourth Military Medical University, Xi'an, China}

\affil*[3]{Institute of Biopharmaceutical and Health Engineering, Tsinghua Shenzhen International Graduate School, Tsinghua University, Shenzhen, China}

\affil*[4]{Department of Engineering Science, University of Oxford, Oxford, UK}

\affil[5]{Jinfeng Laboratory, Chongqing, China}

\affil[6]{Medical Optical Technology R\&D Center, Research Institute of Tsinghua, Pearl River Delta, Guangzhou, China}

\affil[7]{Department of Pathology, Ankang Central Hospital, Ankang, China}

\affil[8]{Jinan Inspur Data Technology Co., Ltd., Jinan, China

+ Qiming He, Jing Li, Zihan Wang and Tian Guan contributed equally to this work.

* Qiming He (he-qm23@tsinghua.org.cn), Chao He (chao.he@eng.ox.ac.uk), Yonghong He (heyh@sz.tsinghua.edu.cn), and Zhe Wang (zhwang@fmmu.edu.cn) are the corresponding authors}

\abstract{
Visual foundation models learn rich pathological morphology but lack explicit concept-level supervision. Vision--language pretraining can provide this supervision but requires costly image--text pairs. Here, we introduce GloSAGE, a semantic self-distillation framework that uses lesion profiles predicted by a pretrained visual teacher to construct structured pathology descriptions. Reliability-guided selection identifies image--text anchors, while Knowledge-Structured Pathology Text Synthesis and Evidence-Guided Negative-pair Construction account for partially observed lesions. Image--text alignment trains projection layers while retaining the frozen visual encoder; an intermediate Semantic Transfer Layer supplies features for downstream tasks. Developed on GloMillion and evaluated across seven glomerulus- and case-level cohorts, GloSAGE supported zero-shot lesion recognition, image--text retrieval, hard-negative discrimination and concept-intervention analyses. It outperformed its source visual model in most lesion-recognition tasks and improved lesion grading, while supporting competitive case-level disease classification, injury grading and baseline renal function estimation. Ablations supported the combined contributions of knowledge-structured text and evidence-guided negatives, and the intermediate transfer layer improved mean lesion-recognition ROC-AUC relative to the complete projection output. A frozen language model conditioned on GloSAGE features generated glomerular descriptions. These findings suggest that teacher-derived lesion profiles can provide annotation-efficient semantic supervision for extending a glomerular visual foundation model.
}

\keywords{Computational pathology, Foundation model, Semantic self-distillation, Vision–language alignment, Glomerular pathology}

\maketitle
\input{Section/introduction}
\input{Section/results}

\input{Section/discussion}
\input{Section/methods}
\input{Section/appendix}

\bibliography{glosage_references}

\input{Section/supplementary}

\end{CJK*}
\end{document}

%% file: Section/introduction.tex
\section{Introduction}\label{introduction}

Self-supervised pretraining has shifted computational pathology from task-specific discriminative pipelines towards foundation representations that support heterogeneous diagnostic, prognostic and biomarker-prediction tasks \cite{chen2024uni,xu2024gigapath,vorontsov2024virchow,wang2024chief}. Nevertheless, independent evaluations indicate that their performance still remains dependent on tissue type, staining protocol, acquisition domain, morphological granularity and downstream readout \cite{campanella2025benchmark,neidlinger2025benchmark,yan2025pathorchestra}. These limitations are particularly relevant when foundation modeling is extended from cancer-dominated, tissue-level analysis to renal pathology, where diagnostically meaningful phenotypes are expressed within repeated anatomical entities and are defined by subtle, coexisting alterations of glomerular cells, capillary loops, basement membranes and mesangial architecture \cite{zhu2026subspecialty, wang2025generic, he2026glopath}. Our previous research introduced GloPath, an entity-centric visual foundation model for glomerular pathology \cite{he2026glopath}. However, morphology-focused visual pretraining primarily structures feature spaces through visual invariance, instance discrimination or reconstruction objectives, without explicitly constraining representations by pathological concepts. Consequently, pathology-relevant semantics may be present in the learned representations but remain insufficiently organized and underexploited for downstream tasks \cite{nakhli2024volta,alfasly2024rotation,huang2025knowledgeadapt,yang2025vlmbias}.

Vision--language pretraining provides a mechanism for introducing concept-level supervision by aligning histopathology images with captions, reports, diagnostic descriptions or instructional responses in a shared embedding space. Contrastive and generative pathology models have consequently expanded computational pathology towards zero-shot recognition, cross-modal retrieval, report generation and interactive visual question answering \cite{lu2024conch,lu2024pathchat,xiang2025musk,ding2025titan,xu2025mstar,tran2025histogpt,chen2025clover,javed2024cplip,shi2024vilamil,seyfioglu2024quiltllava,watawana2024hlss}. Yet the scalability of this paradigm remains coupled to the availability and fidelity of image--text pairs. Web-derived captions may be weakly matched to local morphology, whereas clinical reports are typically written at the case or slide level and do not exhaustively describe every visual entity. Constructing lesion-level descriptions for individual glomeruli would therefore require substantial expert annotation and cross-stain reconciliation. At the same time, a domain-specific visual foundation model already contains predictive knowledge acquired from large-scale morphological pretraining and downstream supervision. Existing knowledge-distillation strategies in computational pathology have largely focused on representation compression, domain adaptation or the aggregation of multiple visual teachers \cite{ma2025unifiedkd,vray2024distillsoda,vuong2025moma}. Whether a visual foundation model can instead reuse its own predictions as concept-level supervision, thereby converting latent morphological knowledge into cross-modal representations without manually paired corpora, remains insufficiently explored.

Repurposing model-derived predictions as semantic supervision introduces two further challenges. First, directly converting prediction into isolated text tokens produces a flat semantic codebook that does not encode the hierarchical, relational and compositional organization through which pathologists interpret lesions. Knowledge-enhanced pathology models have shown that disease ontologies and graph-structured priors can connect diagnostic entities with definitions, attributes and inter-concept relations \cite{zhou2026keep,xu2025mstar,huang2025knowledgeadapt}; however, these structures must be coupled to image-specific evidence if they are to organize rather than merely append textual knowledge. Second, pseudo-label-derived supervision is intrinsically asymmetric in partially observed multi-label pathology data. A glomerulus may contain multiple concurrent lesions, while stain-dependent assessment and annotation workflows reveal only a subset of possible findings. Standard image--text contrastive learning nevertheless treats off-diagonal image--text combinations as negatives, implicitly conflating verified absence with unexamined concepts and potentially separating samples that share positive lesions. More generally, studies of vision--language learning, multi-label contrastive objectives and partial-label disambiguation have demonstrated that semantic overlap and incomplete observations invalidate the conventional one-positive-versus-many-negative assumption \cite{zhang2024vlmsurvey,zhang2024mulsupcon,wang2024picoplus}. Thus, scalable semantic self-distillation requires more than pseudo-text generation: it must jointly establish a structured semantic target and determine which pairwise contrastive relations are supported by the available pathological evidence.

\begin{figure}[htbp]
    \centering
    \includegraphics[width=\textwidth]{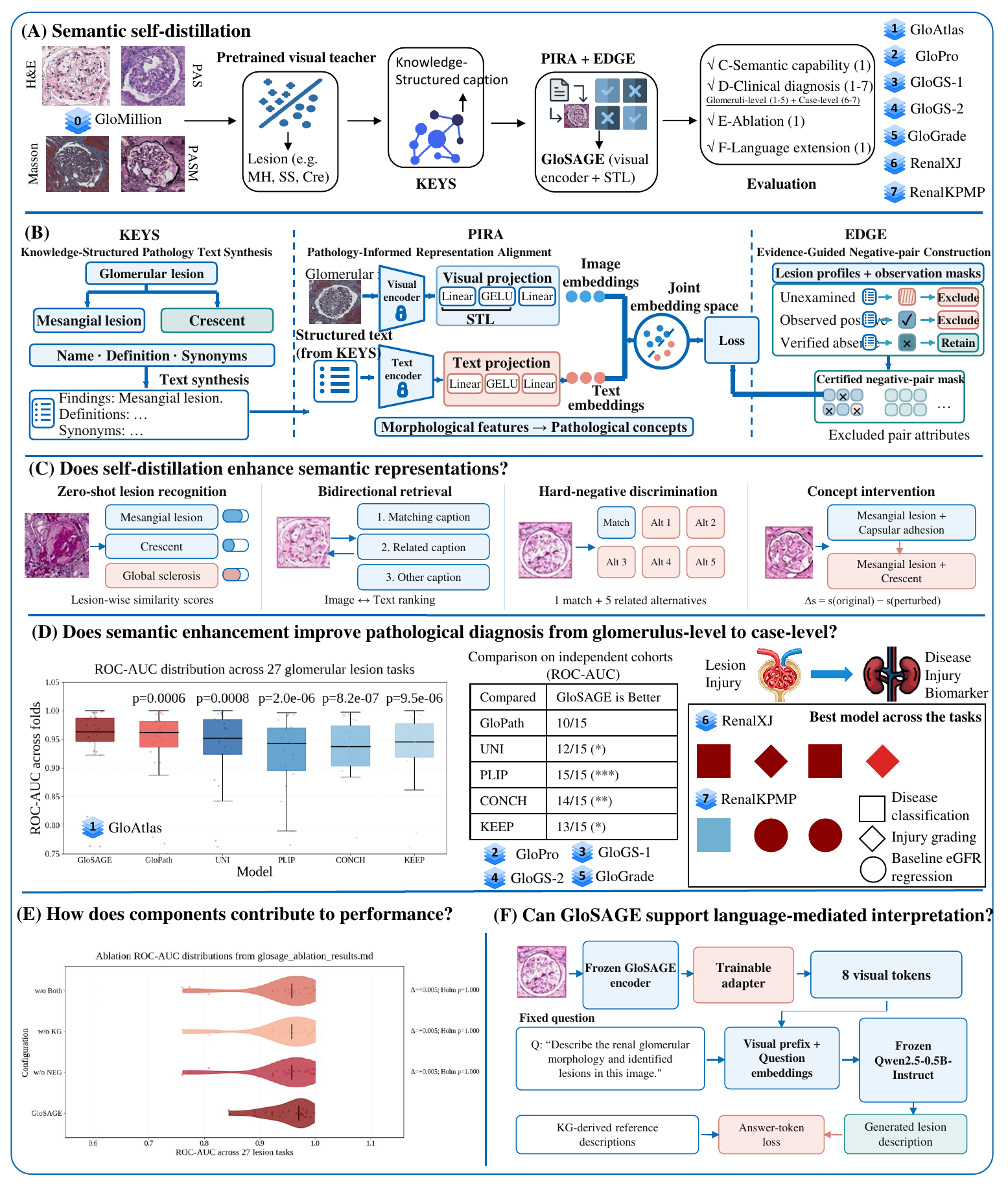}
    \caption{Overview of GloSAGE. \textbf{(A)} Development of GloSAGE through semantic self-distillation. A pretrained visual teacher predicts lesion profiles for glomerular images in GloMillion, and KEYS converts these profiles into knowledge-structured captions. Paired images and captions undergo image--text contrastive pretraining through PIRA with EDGE-guided negative construction. STL transforms pretrained visual features for downstream tasks. GloSAGE is evaluated across seven cohorts for semantic capabilities, clinical diagnosis, component contributions and language-related extensions. \textbf{(B)} Framework of KEYS, PIRA and EDGE. Reliability-guided selection identifies over 40,000 trustworthy visual--semantic anchors. KEYS transforms their lesion profiles into structured descriptions incorporating lesion hierarchies, subtypes and severity grades. PIRA aligns images encoded by the frozen visual encoder with descriptions encoded by PubMedBERT through trainable projection heads, while EDGE excludes unsupported or contradicted negative pairs. STL comprises the first visual projection layer and GELU activation and provides the downstream visual representation. \textbf{(C)} Evaluation of semantic enhancement through zero-shot lesion recognition, bidirectional retrieval, hard-negative discrimination and concept intervention. \textbf{(D)} Evaluation of the diagnostic benefits of semantic enhancement at the glomerular and case levels. Glomerular evaluations include the distribution of ROC-AUC across 27 lesion-recognition tasks (left) and performance on 15 tasks across four independent cohorts (middle). Case-level evaluations include kidney disease diagnosis, subtyping, injury grading and biomarker prediction. \textbf{(E)} Ablation results assessing the contributions of individual components. \textbf{(F)} Exploratory evaluation of language-related capabilities.}
    \label{overview}
\end{figure}

Here, we introduce GloSAGE (Semantic Alignment Guided by Evidence), a knowledge-structured and evidence-aware semantic self-distillation framework for glomerular pathology foundation models (Fig. \ref{overview}). GloSAGE first deploys a pretrained visual teacher across GloMillion to derive knowledge-structured caption using KEYS, and then applies PIRA with EDGE-guided negative construction to preserve diagnostic utility during self-distillation. GloSAGE is then evaluated for semantic capability, clinical diagnosis, ablation, and language extension tasks on seven cohorts. (Fig. \ref{overview}A). In the framework, Knowledge-Structured Pathology Text Synthesis (\textbf{KEYS}) subsequently converts these profiles into structured pathological descriptions containing standardized lesion names, definitions, and synonyms organized by the knowledge graph. To prevent partially observed labels from introducing fabricated negative supervision, GloSAGE further incorporates Evidence-Guided Negative-pair Construction (\textbf{EDGE}), which distinguishes verified absence from unexamined concepts and excludes negative pairs that are unsupported by or inconsistent with the observed lesion profiles. The resulting descriptions are aligned with representations produced by a frozen visual encoder through the Semantic Transfer Layer (\textbf{STL}) and Pathology-Informed Representation Alignment (\textbf{PIRA}), thereby reorganizing latent morphological representations within a cross-modal semantic space (Fig. \ref{overview}B). We evaluated this process through a progressive chain of evidence. Zero-shot lesion recognition, bidirectional image--text retrieval, hard-negative discrimination, and concept intervention were used to determine whether concept-level alignment had been established (Fig. \ref{overview}C), whereas lesion recognition, lesion grading, and case-level disease classification, injury grading, and biomarker regression assessed whether semantic reorganization improved conventional morphological assessment (Fig. \ref{overview}D). Ablation analyses further disentangled the respective contributions of KEYS, EDGE, and STL (Fig. \ref{overview}E). Finally, we examined the extensibility of the learned representations to higher-level pathology applications through fixed-prompt, image-conditioned glomerular description generation (Fig. \ref{overview}F). Collectively, this framework investigates whether morphological knowledge already encoded by a domain-specific visual foundation model can be converted into reliable semantic supervision and subsequently reused to improve both representation quality and downstream capabilities.

%% file: Section/results.tex
\section{Results}\label{results}

\subsection{Overview}
We developed GloSAGE to reorganize the latent morphological representations of a pretrained glomerular visual model through semantic self-distillation on GloMillion (Secs. \ref{sec_glosage} and \ref{dataset_glomillion}). Using the GloAtlas (Secs. \ref{sec:dataset_description} and \ref{dataset_gloatlas}), we evaluated the resulting semantic space in four settings: zero-shot lesion recognition (Fig. \ref{fig:semantic}A), bidirectional image--text retrieval (Fig. \ref{fig:semantic}B), hard-negative discrimination (Fig. \ref{fig:semantic}C), and concept intervention (Fig. \ref{fig:semantic}D). In zero-shot lesion recognition, GloSAGE achieved the best performance on more of the 27 tasks than any competing model, including 80\% of the lesions under PASM staining. In bidirectional retrieval, GloSAGE ranked first in the majority of image-to-text tasks and in nearly all text-to-image tasks. In six-candidate hard-negative discrimination, GloSAGE achieved the highest mean top-1 accuracy on PAS and Masson, whereas its
PASM accuracy was close to chance and slightly below KEEP. In concept intervention, GloSAGE outperformed the second-best model by 12.2\%, 7.7\%, and 5.4\% across the three staining protocols, respectively. We next assessed conventional diagnosis tasks, including glomerulus-level tasks (lesion recognition, and lesion and injury grading on GloAtlas, GloPro, GloGS-1, GloGS-2 and GloGrade (Sec. \ref{dataset_gloatlas}-\ref{dataset_glograde}), as well as case-level tasks on RenalXJ (Sec. \ref{dataset_renalxj}) and RenalKPMP (Sec. \ref{dataset_renalkpmp}). For both lesion recognition and grading on GloAtlas, GloSAGE achieved significantly higher ROC-AUC than each competing model across tasks (all Holm-adjusted $p<0.05$; Fig. \ref{fig:lesion}A-D). Evaluation across four independent cohorts further demonstrated that GloSAGE significantly to highly significantly outperformed most competing models (with Holm-adjusted $p$ values ranging from $<0.05$ to $<0.001$; Fig. \ref{fig:lesion}E-F). At the case level, GloSAGE achieved the best performance across 5 of 7 tasks on disease classification, injury grading, and biomarker regression on RenalXJ and RenalKPMP (Fig. \ref{fig:case}). Component-wise ablations further quantified the contributions of EDGE, KEYS, and STL, which yielded mean gains of 0.92 percentage points (95\% CI, 0.49--1.36; $p<0.001$), 0.98 percentage points (95\% CI, 0.41--1.55; $p<0.01$), and 2.26 percentage points (95\% CI, 0.99--3.53; $p<0.01$), respectively (Fig. \ref{fig:ablation}). Finally, fixed-prompt image-conditioned description generation was used to examine whether the learned representations could support language output. GloSAGE achieved the highest average scores on all the three ROUGE metrics across stains (Fig. \ref{fig:vqa}) and demonstrated performance comparable to that of GPT-6-Astra (\hyperlink{https://chatgpt.com/}{https://chatgpt.com/}). Collectively, these experiments distinguish direct evidence of semantic alignment from its potential utility across downstream pathological tasks.

\subsection{Knowledge-guided alignment reveals lesion-level semantic understanding}
To determine whether knowledge-guided alignment produced a genuinely lesion-aware semantic space, rather than merely improving similarity for familiar image--text pairs, we designed four complementary evaluations that progressively interrogated different properties of the learned representations. 

\subsubsection{Zero-shot lesion recognition}
To assess whether GloSAGE acquired interpretable lesion-level semantics, we performed zero-shot lesion recognition by measuring image–text similarity in a shared embedding space, without training additional classifiers (Fig.~\ref{fig:semantic}A). GloSAGE achieved mean lesion-wise ROC-AUCs of 0.714, 0.689, and 0.759 across PAS, Masson, and PASM stains, respectively, outperforming the strongest overall comparator, CONCH (0.628, 0.602, and 0.622), by 8.6, 8.7, and 13.7 percentage points. At the individual lesion level, GloSAGE achieved the highest ROC-AUC in 7 of 11 PAS tasks, 2 of 6 Masson tasks, and 8 of 10 PASM tasks (Fig.~\ref{fig:semantic}A; Supplementary Tables~\ref{sm_zeroshot_auc_pas}--\ref{sm_zeroshot_auc_pasm}). Notable improvements were observed for PAS(+) (0.828 versus 0.458 for the next-best model), Masson-MH (0.778 versus 0.502), and PASM-MH (0.776 versus 0.513), although CONCH retained an advantage for certain lesions, particularly GS in Masson (0.865 versus 0.596) and PASM (0.952 versus 0.626). We additionally included GloPath as an unaligned visual-only negative control: its image features were compared directly with independently encoded PubMedBERT lesion descriptions, without a trained cross-modal mapping, to illustrate why visual representations alone do not provide a reliable text-similarity readout. Because GloPath has no jointly trained image--text embedding space, it was excluded from retrieval, hard-negative discrimination, and concept-intervention analyses, for which direct cross-modal matching is the primary evaluation. Compared with GloPath, GloSAGE improved mean ROC-AUC by 16.98, 11.40, and 19.10 percentage points across PAS, Masson, and PASM, respectively, with statistically significant improvements for PAS (P = 0.0098) and PASM (P = 0.0020; two-sided Wilcoxon signed-rank test). These results show that the aligned GloSAGE representation supports
classifier-free recognition of the evaluated lesion concepts. The
GloPath image--text comparison is an unaligned negative control,
whereas the comparisons with pretrained vision--language models
should be interpreted in light of their different supervision
histories.

\subsubsection{Bidirectional image--text retrieval}
\begin{figure}[htbp]
    \centering
    \includegraphics[width=\textwidth]{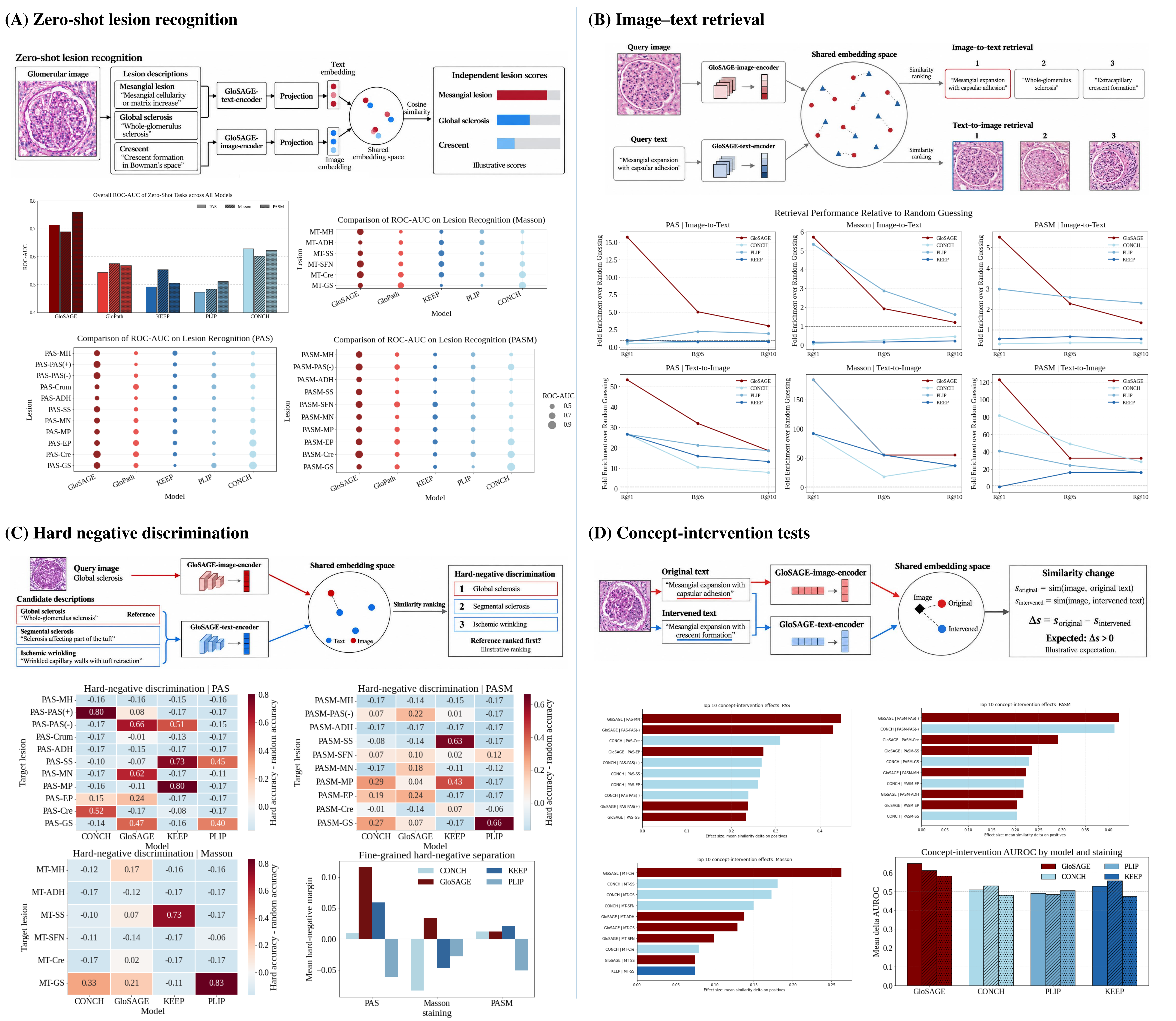}
    \caption{
    Semantic assessment of GloSAGE across three histological stains.
    \textbf{(A)} Experimental setup and performance comparison for zero-shot lesion recognition. The bar chart shows the overall ROC-AUC of each model across all evaluated lesions for the three stains. The three bubble plots illustrate the lesion-wise ROC-AUC rankings of the models for PAS, Masson and PASM, respectively.
    \textbf{(B)} Experimental setup and performance comparison for bidirectional image--text retrieval, evaluated using R@1, R@5 and R@10. The upper and lower rows of line plots represent image-to-text and text-to-image retrieval, respectively, while the three columns correspond to PAS, Masson and PASM.
    \textbf{(C)} Ranking a matched lesion description against five hard negatives. Heatmaps show the difference between hard-negative accuracy and the six-candidate chance level (1/6) for individual lesions, while bar charts summarize the mean difference across all evaluated stress-test targets for each stain.
    \textbf{(D)} Experimental setup and performance comparison for concept intervention. The three horizontal bar charts illustrate the distribution of the best-performing model--lesion combinations across the three stains, while the vertical bar chart compares the overall ROC-AUC across models.
    }
    \label{fig:semantic}
\end{figure}

We next asked whether the learned space could recover the correspondence between an image and a pathological description, rather than merely rank prespecified lesion names. Images were ranked against candidate descriptions (image-to-text), and descriptions against candidate images (text-to-image), with retrieval measured by recall at ranks 1, 5 and 10 (Fig.~\ref{fig:semantic}B). In image-to-text retrieval, GloSAGE consistently achieved the highest R@1 across PAS, Masson, and PASM, reaching 0.1694, 0.2867, and 0.1083, respectively, compared with 0.0109, 0.2671, and 0.0585 for the next-best models. The improvement was particularly pronounced in PAS, where GloSAGE achieved a 15.5-fold higher R@1 than KEEP. This advantage extended to text-to-image retrieval, with GloSAGE achieving the highest or joint-highest R@1 across all three stains (0.0215, 0.1000, and 0.0588, respectively). At broader retrieval ranks, GloSAGE maintained competitive performance, achieving the highest R@5 and R@10 in PAS image-to-text retrieval and the highest or joint-highest R@10 across all three stains in text-to-image retrieval. Although performance varied across models at higher retrieval ranks, GloSAGE demonstrated consistent advantages in top-ranked semantic matching across staining modalities and retrieval directions (Supplementary Tables~\ref{sm_ret_pas}--\ref{sm_ret_pasm}). These findings suggest that semantic alignment enables GloSAGE to establish more precise image–text correspondences, particularly in identifying the most relevant pathological concepts.

\subsubsection{Discrimination of hard-negative descriptions}
Because familiar lesion names can differ by subtle morphology, we tested whether an image ranked its matching lesion description above five closely related but incorrect alternatives. Hard-negative accuracy measures the fraction of lesion-positive images for which the reference description ranked first; chance accuracy with six candidates is 1/6 (Fig.~\ref{fig:semantic}C; Supplementary Table~\ref{sm_hard_negative}). Across 12 PAS and 8 Masson stress-test lesion targets, GloSAGE's mean hard-negative accuracy was 0.283 and 0.201, corresponding to margins above chance of 0.116 and 0.034 and exceeding the other evaluated models in both stains. Across 11 PASM targets, its mean accuracy was 0.179 (margin 0.012), slightly below KEEP (0.188; margin 0.021)(Supplementary Table~\ref{sm_hard_negative}). On PAS, GloSAGE correctly distinguished PAS(-) in 0.827 of positive images, compared with 0.680 for KEEP; for MN, the corresponding accuracies were 0.782 and at most 0.059 among the other models. In Masson, GloSAGE obtained 0.333 for MH versus 0.046 for CONCH, but PLIP surpassed GloSAGE for GS (1.000 versus 0.375). In PASM, GloSAGE achieved 0.406 for EP versus 0.359 for CONCH, while KEEP performed better for MP (0.598 versus 0.205 for GloSAGE). Overall, these results demonstrate GloSAGE's enhanced ability to discriminate morphologically similar lesion concepts, particularly in PAS and Masson stains, while highlighting opportunities for further refinement in challenging lesion-specific distinctions (Fig.~\ref{fig:semantic}C; Supplementary Tables~\ref{sm_hard_negative_pas}--\ref{sm_hard_negative_pasm}).

\subsubsection{Concept-intervention analysis}
Finally, we perturbed lesion-bearing descriptions while leaving the image unchanged to assess whether image–text similarity responded to targeted changes in pathological concepts. For each intervention, we calculated the similarity difference between the original and perturbed descriptions and evaluated its ability to distinguish images positive and negative for the original lesion. The mean similarity difference on positive images was used as an effect size, while its ROC-AUC quantified intervention-discrimination performance (Fig.~\ref{fig:semantic}D; Supplementary Tables~\ref{sm_concept_pas_delta_auc}--\ref{sm_concept_pasm_delta_auc}). Across PAS, Masson and PASM, GloSAGE achieved mean ROC-AUCs of 0.651, 0.584 and 0.613, respectively, outperforming the strongest comparators, KEEP (0.529), PLIP (0.507) and KEEP (0.559). The corresponding mean similarity differences on positive images were 0.095, 0.062 and 0.064, indicating that perturbing matched lesion descriptions generally reduced image–text similarity. Beyond these aggregate improvements, GloSAGE was prominently represented among the ten highest-performing model–lesion combinations for each stain, accounting for 5 of 10 in PAS, 5 of 10 in Masson and 4 of 10 in PASM (Fig.~\ref{fig:semantic}D). This consistent representation among top-performing combinations further supports its ability to capture lesion-specific semantic distinctions beyond improvements in average performance. Nevertheless, intervention sensitivity varied across individual lesions and stains (Supplementary Fig.~\ref{sm_concept_all_size}). For PAS-negative mesangial changes, GloSAGE achieved intervention-discrimination ROC-AUCs of 0.884 in PAS and 0.840 in PASM, compared with 0.529 and 0.653 for CONCH, respectively. On Masson, GloSAGE achieved 0.683 for Cre versus 0.637 for CONCH. In contrast, its performance was lower than KEEP for PAS MH (0.523 versus 0.589) and PASM GS (0.549 versus 0.559). Together, these findings provide further evidence that GloSAGE learns lesion-sensitive semantic representations that respond meaningfully to targeted textual perturbations, supporting its ability to capture pathological concepts beyond superficial image–text associations.

\subsection{Knowledge-guided representations improve glomerulus-level diagnostic tasks}

Glomeruli are fundamental functional units of the kidney, and assessing their structural and functional alterations is essential for understanding the onset and progression of kidney diseases. Previous studies have explored automated glomerular lesion recognition using deep learning, including graph-based modeling, noise-robust learning, and multi-lesion identification approaches \cite{He2024GNN,Li2023Noisy,jayapandian2021development,Yang2022Glomerular}. First, we evaluated GloSAGE on GloAtlas across three stains—periodic acid–Schiff (PAS), Masson's trichrome (hereafter referred to as Masson), and periodic acid–Schiff methenamine silver (PASM)—covering 27 lesion recognition and 10 lesion grading tasks. For lesion recognition, GloSAGE achieved the highest or tied-highest ROC-AUC at the displayed precision in 19 of 27 tasks and mean values of 0.952, 0.937 and 0.949 across PAS, Masson and PASM, respectively (Fig.~\ref{fig:lesion}A,B). For grading, it reached the best result in 6 of 10 tasks, with a mean ROC-AUC of 0.912 across tasks (Fig.~\ref{fig:lesion}C,D). Next, we assessed its performance in identifying global glomerulosclerosis (GS) on GloGS-1 and GloGS-2, as well as grading glomerular injury on GloGrade, a cross-species dataset of mouse kidney tissues (Fig.~\ref{fig:lesion}E). Finally, we evaluated its performance on approximately 14,000 glomeruli from GloPro, representing real-world clinical settings (Fig.~\ref{fig:lesion}F).

\subsubsection{Glomerular lesion recognition}
\begin{figure}[htbp]
    \centering
    \includegraphics[width=\textwidth]{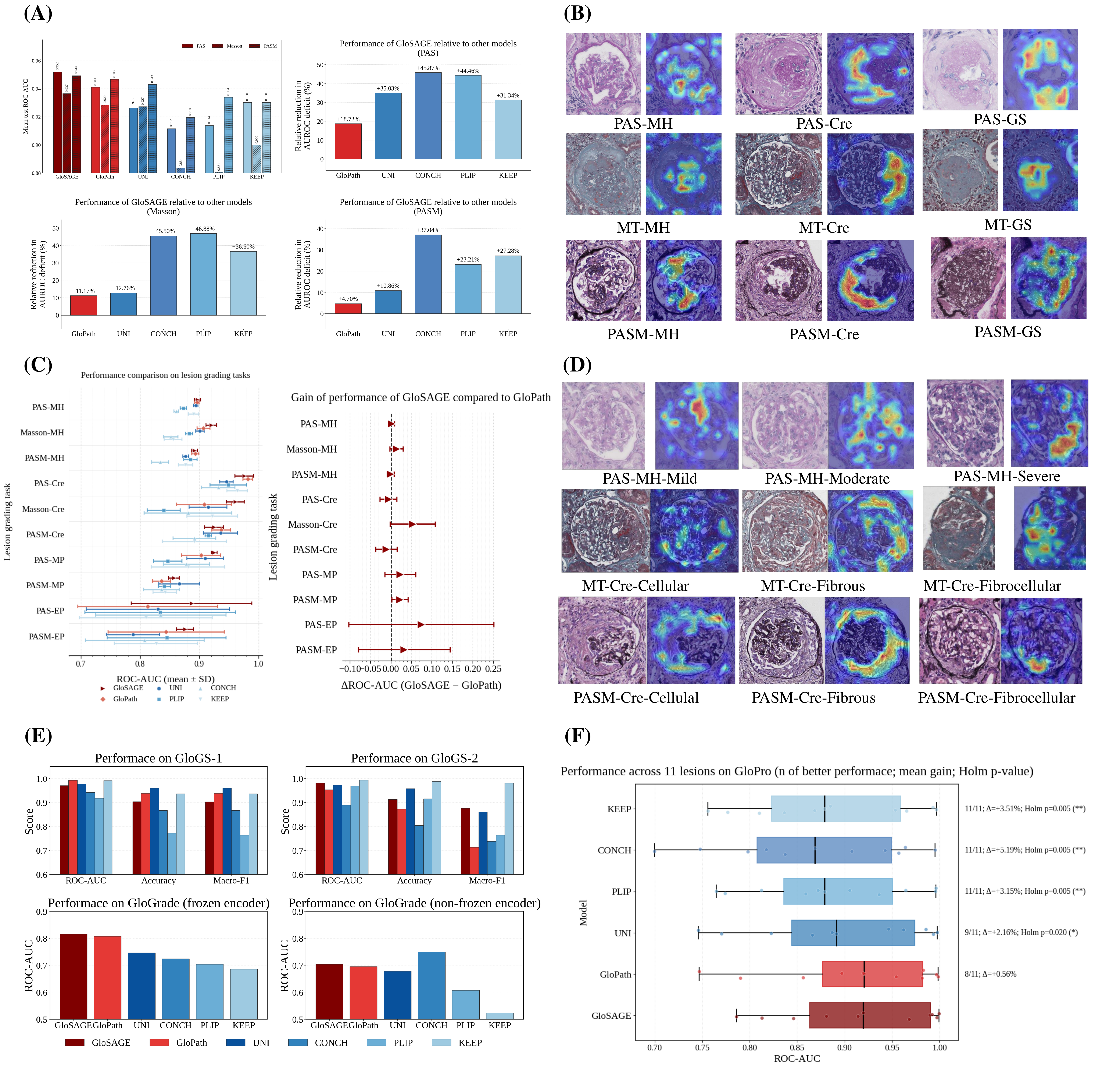}
    \caption{
    GloSAGE enhances diagnostic performance across multiple levels of glomerular pathology analysis.
    \textbf{(A)} Performance comparison of glomerular lesion recognition across different foundation models. The upper-left panel shows the mean ROC-AUC across all lesions for each of the three stains. The upper-right, lower-left, and lower-right panels show the relative reduction in ROC-AUC deficit achieved by GloSAGE compared with other models on PAS, Masson, and PASM stains, respectively.
    \textbf{(B)} Representative visualizations of glomerular lesion recognition by GloSAGE. The three rows correspond to PAS, Masson, and PASM stains, respectively, while the three columns represent mesangial hypercellularity (MH), crescent formation (Cre), and global sclerosis (GS).
    \textbf{(C)} Performance comparison of glomerular lesion grading. The left panel shows the distributions of ROC-AUC across multiple cross-validation folds for different models, while the right panel illustrates the distribution of ROC-AUC improvements achieved by GloSAGE over GloPath.
    \textbf{(D)} Representative visualizations of glomerular lesion grading by GloSAGE. The three rows correspond to PAS, Masson, and PASM stains, respectively, and the three columns illustrate different severity grades or morphological subtypes.
    \textbf{(E)} Performance on GloGS-1 (recognition of GS), GloGS-2 (recognition of GS), GloGrade (injury grading of glomeruli).
    \textbf{(F)} Performance on GloPro (lesion recognition on PAS).
    }
    \label{fig:lesion}
\end{figure}

Glomerular lesion recognition evaluates whether learned representations preserve the morphological distinctions essential for routine pathological diagnosis. We fine-tuned task-specific classifiers for individual lesion targets and evaluated their performance on held-out folds across PAS, Masson and PASM stains. GloSAGE achieved mean ROC-AUCs of 0.952, 0.937 and 0.949 for PAS, Masson and PASM, respectively, outperforming GloPath by 1.1, 0.8 and 0.3 percentage points (Fig.~\ref{fig:lesion}A). Notably, GloSAGE achieved the highest ROC-AUC in 9 of 11 PAS tasks, 3 of 6 Masson tasks and 7 of 10 PASM tasks. These findings highlight the value of semantic enhancement even when conventional ROC-AUC metrics approach saturation. Nevertheless, performance varied across lesions and stains. GloSAGE performed strongly on PAS-MH (0.930), PASM-GS (0.996) and Masson-GS (0.997), whereas ADH remained comparatively challenging across stains (0.815, 0.763 and 0.762 for PAS, Masson and PASM, respectively). Several exceptions were observed: KEEP marginally outperformed GloSAGE on PAS-GS (0.998 versus 0.997), UNI led on PASM-MP (0.969 versus 0.953), and PLIP achieved higher performance on PASM-ADH (0.790 versus 0.762). Lesion-wise bubble plots further illustrate the consistently leading or near-leading performance of GloSAGE across most lesion categories (Supplementary Fig.~\ref{sm_lesion_rec_bubble}; Supplementary Tables~\ref{sm_lesion_rec_pas}--\ref{sm_lesion_rec_pasm}). 

To better characterize performance gains in the high-ROC-AUC regime, we further quantified the relative reduction in ROC-AUC deficit, defined as the proportion of the remaining gap to perfect discrimination (ROC-AUC = 1) eliminated by GloSAGE (Supplementary Table~\ref{tab:relative_auc_deficit_pas}-\ref{tab:relative_auc_deficit_pasm}). Compared with GloPath, GloSAGE reduced the overall ROC-AUC deficit by 18.72\%, 11.17\% and 4.70\% across PAS, Masson and PASM, respectively. Notably, these improvements varied across lesion types and staining modalities. In PAS, substantial reductions were observed for crescent formation (Cre, 45.42\%) and crumpling (Crum, 30.81\%), indicating improved discrimination of lesions characterized by structural alterations. In Masson, the largest gains occurred in glomerulosclerosis (GS, 48.72\%) and segmental sclerosis (SS, 41.96\%), suggesting enhanced sensitivity to sclerotic and fibrotic morphological changes. In PASM, GloSAGE showed pronounced improvements in Cre (34.16\%), PAS-negative mesangial changes (32.15\%), SS (28.37\%) and GS (26.93\%). By contrast, improvements were limited or slightly negative for several already well-recognized lesions, including PAS-GS, PASM-MH and PASM-ADH, while Masson-Cre showed a more substantial decline. Relative to general-purpose foundation models, GloSAGE further reduced the overall ROC-AUC deficit by up to 45.87\%, 46.88\% and 37.04\% across PAS, Masson and PASM, respectively. These findings suggest that knowledge-guided semantic enhancement provides meaningful gains beyond conventional ROC-AUC improvements, particularly for lesion categories involving complex architectural remodeling, although the benefits remain lesion- and stain-dependent.

To investigate the morphological evidence underlying these predictions, we visualized lesion-specific activation patterns using Grad-CAM (Fig.~\ref{fig:lesion}B). Across different stains, GloSAGE preferentially highlighted mesangial regions exhibiting matrix expansion and increased mesangial cellularity in MH, suggesting sensitivity to the structural and cellular alterations characteristic of mesangial hypercellularity. For crescent formation (Cre), the activation maps emphasized extracapillary regions containing proliferating parietal epithelial cells and associated fibrous components, consistent with the morphological organization of crescentic lesions. In glomerulosclerosis (GS), GloSAGE localized regions characterized by sclerosis and obliteration of glomerular capillary loops, reflecting its sensitivity to the architectural disruption associated with sclerotic injury. Together, these quantitative and qualitative findings suggest that knowledge-guided semantic enhancement improves lesion-level discrimination while preserving sensitivity to diagnostically relevant morphological structures across histological stains.

\subsubsection{Glomerular lesion grading}

Glomerular lesion grading provides clinically relevant insights into pathological severity and morphological heterogeneity. We therefore evaluated GloSAGE on ten grading and morphological subtype classification tasks across PAS, Masson trichrome (MT), and PASM stains, covering mesangial hypercellularity (MH), crescent formation (Cre), membranoproliferative lesions (MP), and endocapillary proliferation (EP). Specifically, MH was assessed using four ordinal severity grades (absent, mild, moderate, and severe), whereas Cre was classified into cellular, fibrocellular, and fibrous subtypes. MP and EP were further categorized according to segmental or global involvement (Supplementary Table~\ref{sm_lesion_grading}).

Across the ten tasks, GloSAGE achieved the highest mean ROC-AUC of 0.912, outperforming GloPath (0.892, $p=0.0484$), UNI (0.886, $p=0.0295$), KEEP (0.876, $p=0.0010$), PLIP (0.871, $p=0.0056$), and CONCH (0.861, $p<0.01$). GloSAGE achieved the best performance in six tasks, including PAS-MH (0.897), MT-MH (0.920), MT-Cre (0.961), PAS-MP (0.925), PAS-EP (0.887), and PASM-EP (0.876). Notably, GloSAGE outperformed its visual predecessor, GloPath, in 70\% of the tasks, with an average absolute ROC-AUC improvement of 2.85 percentage points across these tasks. The most pronounced improvements were observed in PAS-EP (0.887 versus 0.813), MT-Cre (0.961 versus 0.908), and PASM-EP (0.876 versus 0.844). Nevertheless, GloPath remained marginally superior in PASM-MH (0.893 versus 0.892) and PASM-Cre (0.937 versus 0.925), while UNI achieved the highest ROC-AUC in PASM-MP (0.866 versus 0.857). Collectively, these results suggest that knowledge-guided semantic reorganization enhances the discrimination of lesion severity and morphological subtypes, although the magnitude of improvement varies across stains and lesion categories.

The grading heatmaps in Fig.~\ref{fig:lesion}D further illustrate the morphological distinctions underlying these tasks. For PAS-MH, the visualizations span mild, moderate, and severe mesangial hypercellularity, characterized by progressively increased mesangial cellularity and expansion of mesangial regions. For MT-Cre, the heatmaps highlight distinct crescentic morphologies, including cellular crescents dominated by proliferating cells, fibrocellular crescents containing both cellular and extracellular matrix components, and fibrous crescents characterized by collagen-rich fibrotic tissue. Similarly, the PASM-Cre examples illustrate these three crescent subtypes, with differences in cellular composition and extracellular matrix deposition reflected in their glomerular architecture. Together, these visualizations provide qualitative examples of the diverse pathological patterns involved in lesion grading and subtype classification, complementing the quantitative results. Full task-wise means and standard deviations are provided in Supplementary Table~\ref{sm_lesion_grading}.

\subsubsection{Generalizability evaluation on independent cohorts}
GS represents an advanced and irreversible form of glomerular injury and serves as an important histopathological indicator of chronic kidney damage. To assess the generalizability of the GloSAGE framework, we externally validated the GloAtlas-derived lesion recognition models on GloGS-1 and GloGS-2 and further evaluated GloSAGE for glomerular injury grading on GloGrade. For external GS recognition, GloSAGE maintained strong discriminative performance, achieving ROC-AUCs of 0.9717 and 0.9813 on the two cohorts, respectively (Fig. \ref{fig:lesion}E). Although GloSAGE did not consistently outperform all competing models, its performance remained competitive, with several models achieving similarly high ROC-AUCs. This may be attributed to the relatively distinctive morphological characteristics of global glomerulosclerosis, which make GS recognition a comparatively straightforward task and limit the additional benefits of semantic enhancement. Nevertheless, these results support the cross-cohort generalizability of GloSAGE for glomerular lesion recognition.

To further assess cross-species generalizability, we evaluated glomerular injury grading on GloGrade, an independent mouse glomerular pathology dataset. With the encoder frozen, GloSAGE achieved the highest ROC-AUC of 0.8156, outperforming GloPath by 0.80 percentage points and four general-purpose pathology foundation models (UNI, CONCH, PLIP, and KEEP) by an average of 9.99 percentage points, highlighting the strong cross-species transferability of its semantically enriched representations. Interestingly, end-to-end fine-tuning reduced GloSAGE's ROC-AUC to 0.7043, with similar declines observed in most competing models. This may reflect overfitting and representation drift caused by limited training data and cross-species distribution shifts, whereas freezing the encoder preserves generalizable morphological representations learned during large-scale pretraining.

\subsubsection{Large-scale real-world deployment}
To assess the robustness and clinical applicability of GloSAGE under real-world conditions, we constructed GloPro, a large-scale dataset comprising approximately 14,000 PAS-stained glomeruli collected consecutively from routine clinical practice, covering 11 glomerular lesion recognition tasks. Across these 11 tasks, GloSAGE demonstrated superior overall performance, achieving a mean ROC-AUC of 0.9188 and outperforming all five competing models (Fig. \ref{fig:lesion}F). Notably, GloSAGE achieved higher ROC-AUCs than CONCH, KEEP, and PLIP across all 11 tasks, with mean improvements of 5.19, 3.51, and 3.15 percentage points, respectively (all Holm-adjusted $p=0.005$). It also outperformed UNI in 9 of 11 tasks, with an average improvement of 2.16 percentage points (Holm-adjusted $p=0.020$). Compared with its visual-only predecessor GloPath, GloSAGE achieved superior performance in 8 of 11 tasks, further supporting the benefits of knowledge-structured semantic enhancement in real-world clinical settings.

At the individual lesion level, GloSAGE achieved the highest ROC-AUC in 7 of 11 tasks, including MH (0.9196), Crum (0.7863), ADH (0.8127), and MP (0.9919) (Table \ref{sm_glopro_recognition}). Importantly, despite the distribution shift from the curated GloAtlas dataset to real-world GloPro, GloSAGE maintained excellent discrimination for several lesions, particularly PAS(-), mesangial proliferation, endocapillary proliferation, and global glomerulosclerosis, all achieving ROC-AUCs above 0.99 or close to this threshold. Moreover, GloSAGE exhibited improved performance on mesangial proliferation and endocapillary proliferation compared with its internal GloAtlas evaluation, suggesting that its semantically enriched representations can generalize effectively to heterogeneous clinical samples. Collectively, these findings highlight the robustness and practical potential of GloSAGE for large-scale glomerular lesion assessment in real-world clinical settings.

\subsection{Knowledge-guided representations improve case-level diagnostic tasks}
Glomerular abnormalities are central to the pathogenesis and progression of numerous kidney diseases, and their histopathological characteristics provide important diagnostic and prognostic information beyond conventional clinical indicators \cite{srivastava2018prognostic}. Therefore, extending glomerulus-level representations to case-level clinical assessment is essential for evaluating their translational utility. Leveraging the deep semantic representations and lesion probability profiles derived from GloSAGE, we conducted comprehensive case-level evaluations on RenalXJ and RenalKPMP, spanning disease classification, kidney injury severity grading, and renal function biomarker prediction.

\subsubsection{Performance on RenalXJ}
\begin{figure}[htbp]
    \centering
    \includegraphics[width=\textwidth]{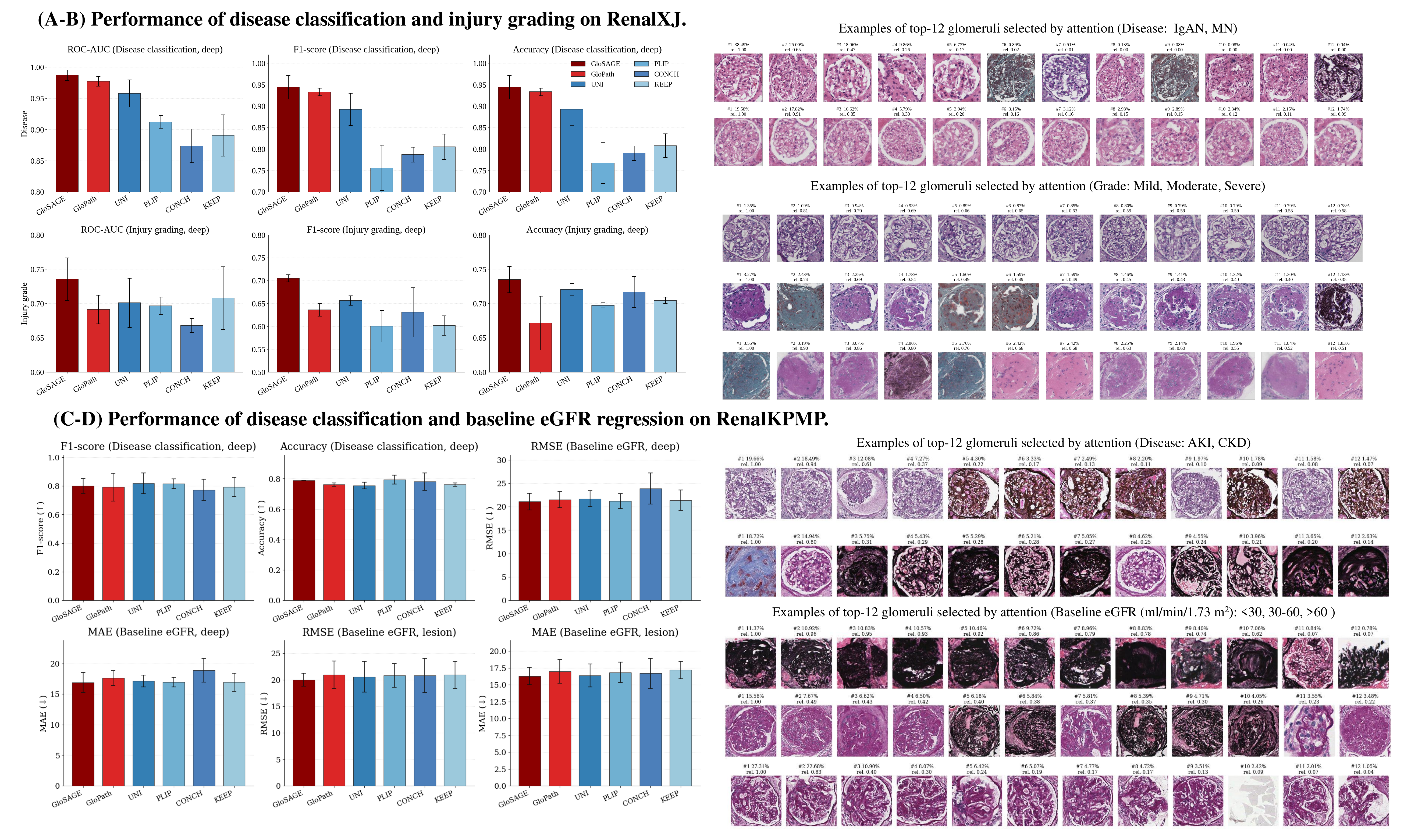}
    \caption{
    \textbf{(A)} Case-level performance comparison for disease classification (top row) and injury grading (bottom row) on RenalXJ.
    \textbf{(B)} Representative examples of the top 12 glomeruli ranked by attention weights in case-level disease classification and injury grading on RenalXJ.
    \textbf{(C)} Case-level performance comparison for disease classification (top row) and biomarker regression (bottom row) on RenalKPMP.
    \textbf{(D)} Representative examples of the top 12 glomeruli ranked by attention weights in case-level disease classification and biomarker regression on RenalKPMP.
    }
    \label{fig:case}
\end{figure}

Case-level diagnosis requires aggregating heterogeneous glomeruli across a biopsy while capturing both disease-specific morphology and tissue injury. We therefore employed attention-based multiple instance learning (MIL) using either glomerular deep features or lesion probability profiles (Fig.~\ref{fig:case}A-B). Disease classification involved 236 IgA nephropathy and 217 membranous nephropathy cases, while injury grading included 744 cases across grades 0, 1, and 2 (483, 162, and 99 cases, respectively).

\textbf{Deep feature-based evaluation.} GloSAGE achieved the highest disease classification ROC-AUC (0.9873~$\pm$~0.0085), weighted-F1 (0.9444~$\pm$~0.0274), and accuracy (0.9445~$\pm$~0.0273), exceeding GloPath, the strongest comparator, by 1.0, 1.1, and 1.1 percentage points, respectively. Its advantage was more pronounced over general-purpose pathology foundation models, with ROC-AUC improvements ranging from 2.91 to 11.33 percentage points. For injury grading, GloSAGE consistently achieved the best ROC-AUC (0.7358~$\pm$~0.0310), weighted-F1 (0.7054~$\pm$~0.0078), and accuracy (0.7352~$\pm$~0.0193), outperforming the strongest comparators by 2.78, 4.83, and 1.46 percentage points, respectively (Supplementary Tables~\ref{sm_renalxj_disease_deep} and~\ref{sm_renalxj_injury_deep}). These results demonstrate the effectiveness of GloSAGE's semantically enriched representations for integrating heterogeneous glomerular morphology into case-level diagnostic predictions.

\textbf{Lesion probability-based evaluation}. To further investigate the clinical utility of explicit pathological concepts, we performed case-level MIL using lesion probability profiles instead of deep features. For disease classification, GloSAGE achieved the highest weighted-F1 (0.8657~$\pm$~0.0017) and accuracy (0.8667~$\pm$~0.0015), while maintaining a competitive ROC-AUC of 0.9334~$\pm$~0.0258, slightly below GloPath (0.9458). For injury grading, GloSAGE achieved an ROC-AUC of 0.7433~$\pm$~0.0406 and accuracy of 0.7259~$\pm$~0.0231, remaining competitive with GloPath and CONCH, which achieved the highest ROC-AUC (0.7690) and accuracy (0.7398), respectively (Supplementary Tables~\ref{sm_renalxj_disease_lesion} and~\ref{sm_renalxj_injury_lesion}). Notably, GloSAGE achieved a slightly higher injury grading ROC-AUC using lesion probabilities than deep features (0.7433 vs. 0.7358), suggesting that explicit lesion-level information can provide complementary value for assessing kidney injury severity. Together, these findings support the utility of GloSAGE for case-level assessment through both high-dimensional semantic representations and interpretable lesion probability profiles.

Attention visualizations further illustrate how GloSAGE aggregates morphologically informative glomeruli for case-level prediction (Fig.~\ref{fig:case}B). For disease classification, the highest-attention glomeruli in representative IgA nephropathy cases exhibit prominent mesangial hypercellularity and mesangial matrix expansion, consistent with characteristic mesangial alterations. In contrast, highly attended glomeruli in membranous nephropathy cases predominantly show relatively preserved glomerular architecture with diffuse capillary wall thickening and limited mesangial proliferation, reflecting the characteristic morphological pattern of membranous nephropathy. For injury grading, attention highlights distinct patterns across increasing severity: mildly injured glomeruli generally retain recognizable capillary loops with relatively limited structural alterations; moderately injured glomeruli exhibit more pronounced mesangial expansion, capillary loop distortion, and segmental structural damage; and severely injured glomeruli are characterized by extensive sclerosis, extracellular matrix accumulation, and substantial loss of normal glomerular architecture. Notably, attention is concentrated on a small subset of representative glomeruli for disease classification, whereas injury grading involves a broader distribution of attention across glomeruli exhibiting heterogeneous degrees of damage. Although these visualizations do not establish lesion-level localization accuracy, they provide qualitative evidence that GloSAGE prioritizes pathologically relevant morphological patterns when integrating glomerulus-level representations into case-level predictions.

\subsubsection{Performance on RenalKPMP}

Acute kidney injury (AKI) and chronic kidney disease (CKD) exhibit distinct, albeit partially overlapping, renal histopathological characteristics, providing a morphological basis for case-level disease classification. Meanwhile, baseline estimated glomerular filtration rate (eGFR) is a key indicator of kidney function and CKD severity. Previous work has demonstrated associations between glomerular histomorphological features and renal biomarkers, providing a rationale for morphology-based prediction of kidney function \cite{holscher2023next}. We therefore evaluated GloSAGE on RenalKPMP for AKI/CKD classification using deep semantic features and baseline eGFR regression using both deep semantic features and lesion probability profiles, aiming to investigate the relationship between glomerular morphology and renal functional biomarkers.

For AKI/CKD classification, GloSAGE achieved an F1-score of 0.8003~$\pm$~0.0523 and an accuracy of 0.7885, outperforming GloPath (0.7915 and 0.7613, respectively) and achieving the second-highest accuracy among all models (Fig. \ref{fig:case}C). Although UNI and PLIP achieved slightly higher F1-scores (0.8185 and 0.8165, respectively), their advantages over GloSAGE were relatively modest. Notably, GloSAGE improved accuracy over GloPath by 2.72 percentage points, suggesting that semantic enhancement provides additional discriminative information beyond visual representations alone. The relatively small differences among the leading models may reflect the overlapping glomerular manifestations of AKI and CKD, as their distinction also depends on tubulointerstitial alterations and clinical disease duration, which are not fully captured by glomerulus-level representations.

For baseline eGFR prediction, GloSAGE demonstrated consistent advantages across both feature representations (Fig. \ref{fig:case}C). Using deep semantic features, GloSAGE achieved the lowest RMSE (21.0978~$\pm$~1.7840) and MAE (16.9080~$\pm$~1.6373), outperforming all five competing models. Importantly, lesion probability-based regression further improved its performance, reducing RMSE to 20.0111~$\pm$~1.2494 and MAE to 16.2814~$\pm$~1.2926, corresponding to reductions of 5.15\% and 3.71\%, respectively, compared with its deep feature-based results. GloSAGE again achieved the lowest errors among all competing models, outperforming the strongest comparator, UNI, by 2.56\% in RMSE and 0.61\% in MAE. These findings suggest that explicit lesion probability profiles may capture clinically relevant pathological information associated with renal function, complementing high-dimensional semantic representations. Collectively, the consistent performance of GloSAGE across both regression settings supports the potential of knowledge-guided glomerular representations for morphology-based renal function assessment.

To further investigate the morphological relevance of GloSAGE-derived representations, we visualized the top-12 glomeruli with the highest attention weights for representative cases in AKI/CKD classification and baseline eGFR prediction (Fig.~\ref{fig:case}D). For disease classification, highly attended glomeruli exhibited heterogeneous pathological patterns, with CKD cases showing more prominent chronic structural alterations, including glomerulosclerosis and architectural distortion, whereas AKI cases displayed relatively preserved glomerular architecture in several instances. For eGFR prediction, attention patterns also varied across renal function categories ($\textless$30, 30–60, and $\textgreater$60 mL/min/1.73 m²), with severely impaired renal function generally associated with more pronounced glomerular structural abnormalities, while higher eGFR cases exhibited comparatively preserved morphology. These visualizations suggest that GloSAGE-based MIL can prioritize morphologically informative glomeruli associated with disease identity and renal functional status, providing qualitative interpretability for case-level predictions.

\subsection{Ablation study}
To assess the contributions of GloSAGE's semantic design, we examined three components linking image--text pretraining to downstream lesion classification (Fig.~\ref{fig:ablation}A): KEYS, which constructs knowledge-structured pathological descriptions; EDGE, which filters candidate negative pairs according to observed lesion evidence; and STL, which provides the intermediate visual projection features for downstream prediction. We evaluated the removal of KEYS and EDGE, individually and jointly, and compared STL features with the output of the complete visual projection head.

\subsubsection{Separating knowledge structure from evidence-aware negatives}

\begin{figure}[H]
    \centering
    \includegraphics[width=\textwidth]{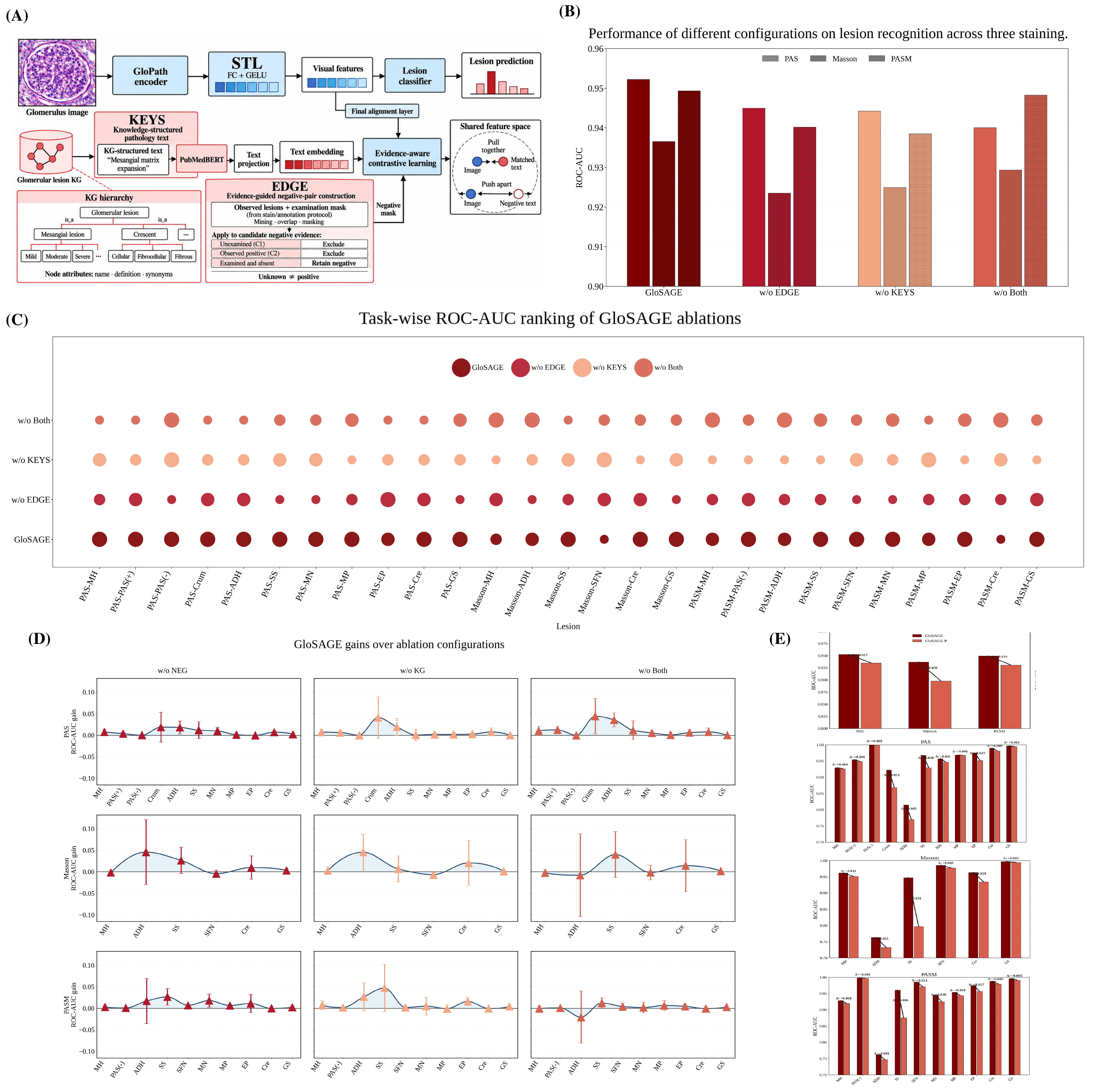}
    \caption{Ablation of the semantic components of GloSAGE. \textbf{(A)} GloSAGE architecture highlighting KEYS, EDGE and STL. \textbf{(B)} ROC-AUC rankings of the ablation configurations across individual lesions. \textbf{(C)} Mean ROC-AUC for each stain. \textbf{(D)} GloSAGE gains relative to each ablation configuration on lesion-recognition tasks. Rows correspond to PAS, Masson and PASM, respectively; columns correspond to w/o EDGE, w/o KEYS and w/o EDGE and KEYS, respectively. \textbf{(E)} Comparison of STL features with the complete projection-head output (GloSAGE-P). The upper-left panel shows mean ROC-AUC across the three stains; the upper-right, lower-left and lower-right panels show lesion-specific ROC-AUC for PAS, Masson and PASM, respectively.}
    \label{fig:ablation}
\end{figure}

To assess the contributions of EDGE and KEYS, we compared plain image--text alignment (w/o EDGE and KEYS), alignment with EDGE alone (w/o KEYS), alignment with KEYS alone (w/o EDGE), and the complete GloSAGE model across 27 lesion-recognition tasks. GloPath served as the vanilla visual-pretraining reference. Plain image--text alignment largely preserved the diagnostic performance of GloPath, yielding an overall mean ROC-AUC of 0.9407 versus 0.9404. Stain-specific means were 0.9400, 0.9293 and 0.9482 for PAS, Masson and PASM (Fig. \ref{fig:ablation}B), respectively, compared with 0.9410, 0.9287 and 0.9468 for GloPath. These mixed changes suggest that image--text supervision alone was insufficient to consistently improve the source visual representation.

Adding either EDGE or KEYS alone improved the PAS mean but reduced the Masson and PASM means, resulting in overall ROC-AUCs of 0.9378 and 0.9384, respectively, below the plain-alignment baseline (Fig. \ref{fig:ablation}A). Combining both components yielded the highest overall mean of 0.9476. GloSAGE exceeded EDGE alone by 0.98 percentage points (95\% CI, 0.41--1.55; Holm-adjusted \(p=0.0034\)) and KEYS alone by 0.92 percentage points (95\% CI, 0.49--1.36; Holm-adjusted \(p=0.0008\)). Relative to plain alignment, the improvement was 0.70 percentage points (95\% CI, 0.15--1.24; Holm-adjusted \(p=0.0144\)), supporting the joint configuration rather than uniform benefits from either component in isolation.

The largest gains over plain alignment included PAS Crum (0.8779 to 0.9220) and Masson segmental sclerosis (0.9077 to 0.9475); PASM endocapillary lesions also improved from 0.9701 to 0.9747 (Fig. \ref{fig:ablation}C-D). These patterns may reflect the value of combining structured descriptions that distinguish related concepts with evidence-guided negatives that avoid penalizing coexisting or incompletely examined lesions. Such supervision could be particularly helpful for focal and heterogeneous abnormalities. Benefits were not universal: plain alignment performed better for Masson and PASM capsular adhesion, whereas near-saturated tasks showed little variation (Supplementary Tables~\ref{sm_ablation_recognition_pas}--\ref{sm_ablation_recognition_pasm}).

\subsubsection{Comparison with the complete projection head}

We next compared three visual representations: the original GloPath features, the output of the complete projection head (GloSAGE-P), and the intermediate features from STL, comprising the first fully connected layer and GELU activation. GloPath achieved an overall mean lesion-recognition ROC-AUC of 0.9404. Using the complete projection output reduced this mean to 0.9250, with stain-specific means of 0.9348, 0.8977 and 0.9306 for PAS, Masson and PASM. In contrast, STL yielded the highest overall mean of 0.9476, exceeding GloSAGE-P by 2.26 percentage points (95\% CI, 0.99--3.53; Holm-adjusted \(p=0.0034\)) and outperforming it on all 27 tasks (Fig.~\ref{fig:ablation}E).

The differences were most pronounced for challenging lesions. For Masson segmental sclerosis, ROC-AUC decreased from 0.9100 with GloPath to 0.7966 with the complete projection head, but increased to 0.9475 with STL. A similar pattern occurred for PAS-Crum (0.8870, 0.8685 and 0.9220) and PASM-SS (0.9450, 0.8750 and 0.9609), whereas near-saturated tasks showed smaller differences. These results suggest that the final embedding optimized for cross-modal alignment may constrain information useful for lesion discrimination, while STL provides an intermediate representation that benefits from semantic training and retains greater diagnostic utility.

\subsection{GloSAGE connects glomerular morphology to language}

\begin{figure}[htbp]
    \centering
    \includegraphics[width=0.95\textwidth]{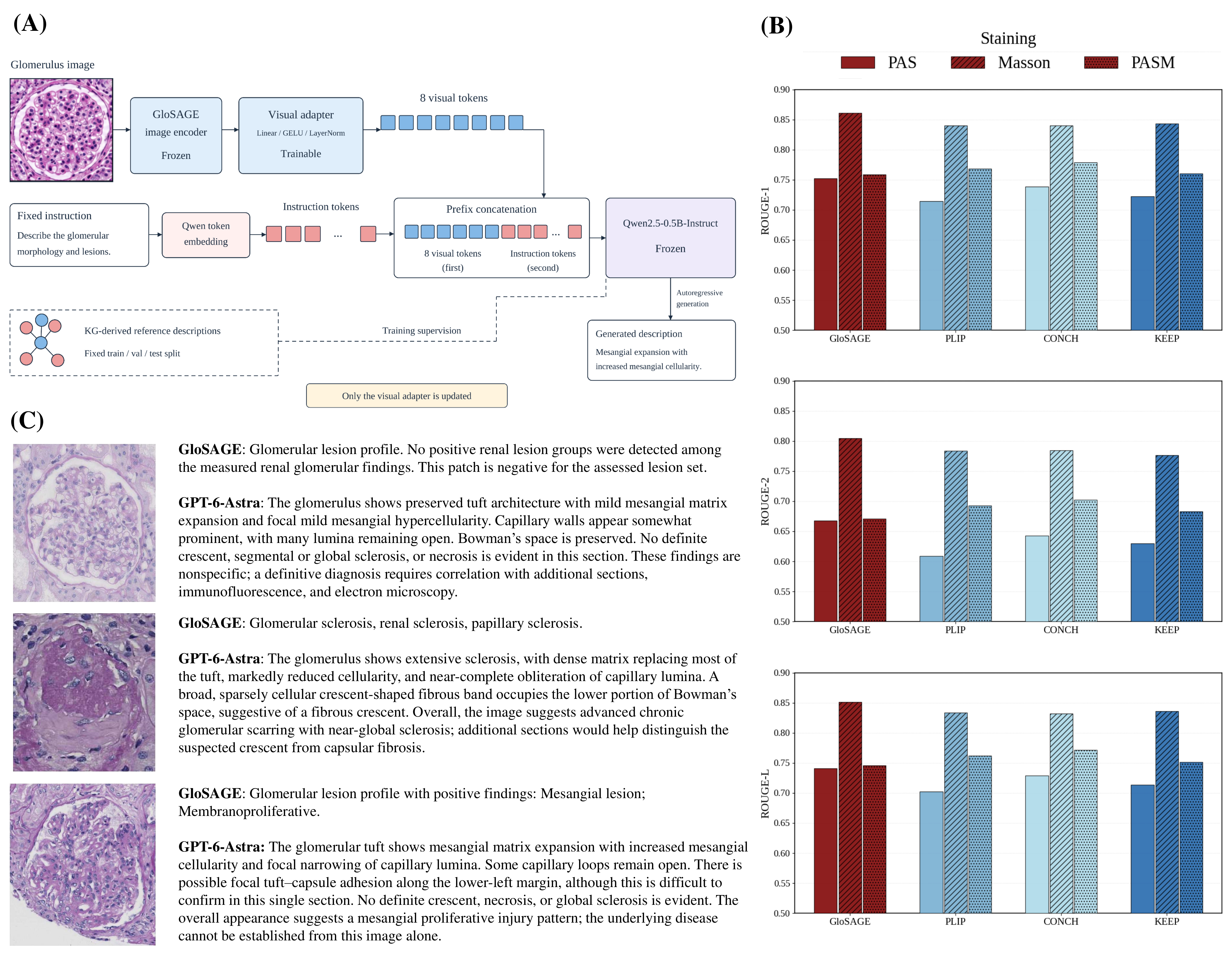}
    \caption{Image-conditioned generation of glomerular descriptions.
\textbf{(A)} A trainable visual-prefix adapter converts features from a frozen
visual encoder into eight tokens that condition a frozen language
model. Each image is paired with the same fixed descriptive
instruction; knowledge-graph-derived descriptions provide the
training references. \textbf{(B)} ROUGE-1, ROUGE-2, and ROUGE-L recall
scores against structured reference descriptions across PAS, Masson,
and PASM. \textbf{(C)} Selected generated descriptions from the
GloSAGE-based system alongside GPT-6-Astra outputs for the same
images. Panel C is qualitative and is not a quantitative comparison
between the two systems.}
    \label{fig:vqa}
\end{figure}

\subsubsection{Exploratory evaluation of image-conditioned description generation}

We explored whether GloSAGE representations could support language-based interpretation by coupling each visual encoder to a frozen Qwen2.5-0.5B-Instruct model through a trainable visual-prefix adapter. The adapter converted image features into eight visual tokens, which were prepended to the instruction embeddings for autoregressive generation (Fig.~\ref{fig:vqa}A). The visual encoder and language model remained frozen, and only the visual-prefix adapter was trained. Each test image was paired with a fixed instruction requesting a description of glomerular morphology and lesions; this setting therefore assessed image-conditioned description generation rather than unrestricted question answering. Across the three stains, GloSAGE achieved the highest mean ROUGE-1 (0.7906), ROUGE-2 (0.7143), and ROUGE-L (0.7791). Performance varied by stain: GloSAGE led all three ROUGE metrics on PAS and Masson, achieving a ROUGE-L of 0.8509 on Masson, whereas CONCH led all three metrics on PASM. These results suggest that GloSAGE features can support generation of descriptions with greater reference-text overlap on average (Supplementary Table \ref{sm_vqa_all_models}).

\subsubsection{Qualitative exploration and comparison with GPT-6-Astra}

Representative examples illustrate how the adapted model translates glomerular images into lesion descriptions. We also juxtaposed its outputs with those of GPT-6-Astra to explore similarities and differences in the descriptions produced by a domain-adapted model and a general-purpose multimodal model. The selected cases illustrate the outputs and limitations of the two systems but were not used for quantitative comparison, providing qualitative context for the overlap-based evaluation. This comparison was limited to selected examples and was not a systematic head-to-head benchmark. It therefore does not establish equivalent performance, superiority or comparable clinical reliability relative to GPT-6-Astra. Broader evaluation with standardized prompts, representative test cases and blinded expert assessment would be needed to determine the accuracy, completeness and clinical usefulness of the generated descriptions.

%% file: Section/discussion.tex
\section{Discussion}\label{discussion}

This study shows that the knowledge accumulated by a visual pathology foundation model can serve as a starting point for its semantic extension. Visual pretraining has produced transferable morphological representations, including general-purpose pathology models and the glomerulus-specific GloPath model \cite{chen2024uni,he2026glopath}. GloSAGE builds on this foundation by converting teacher-predicted lesion profiles into image-specific textual supervision and aligning them with frozen visual features. The resulting model supports both concept-based interaction and conventional diagnostic readouts without requiring a newly annotated image--text corpus or retraining the visual encoder. This suggests a complementary route for developing pathology foundation models: knowledge acquired during visual model development can be reorganized into representations that are more readily accessed through pathological concepts.

The semantic evaluations illuminate what this reorganization makes possible. Zero-shot lesion recognition uses descriptions as direct readouts of image features; bidirectional retrieval examines correspondence between images and structured lesion profiles; hard-negative discrimination and concept intervention probe whether similarity responds to differences in lesion meaning. Together, these analyses indicate that the learned space supports several forms of interaction beyond a task-specific classification head. Existing pathology vision--language models have established the value of image--text alignment for retrieval and zero-shot analysis \cite{lu2024conch,zhou2026keep}. GloSAGE extends this line of work at the glomerulus level by deriving image-specific text from an existing visual teacher and organizing it around a defined lesion vocabulary. Its semantic capability is therefore best understood as making previously acquired pathological knowledge accessible through a new interface, rather than creating pathological information absent from the source visual representation.

The provenance of that knowledge is central to interpreting the results. The GloPath-based lesion predictors used to generate GloSAGE's training descriptions had been trained on GloAtlas annotations. Their predictions subsequently supplied semantic supervision for selected GloMillion images; GloAtlas images were not used as the image--text pairs for semantic alignment. Accordingly, the GloAtlas experiments examine whether knowledge acquired by the visual teacher can be expressed and reused through language-aligned representations. In this setting, ``zero-shot'' denotes inference without fitting an additional lesion classifier, rather than concepts unseen throughout model development. This distinction clarifies the contribution of semantic self-distillation: it converts established visual diagnostic knowledge into a form that can be queried by text and transferred to downstream readouts. The separately assembled evaluation cohorts broaden the assessment beyond this internal knowledge-transfer setting.

The ablations provide a second insight into how this conversion works. KEYS specifies the lesion concepts represented in the text, whereas EDGE determines which candidate negatives are supported by the available observation record. Their combination performed better than plain alignment or either component in isolation, suggesting that informative semantic targets and defensible contrastive comparisons are mutually important when lesion profiles are partially observed. This differs from approaches that incorporate broader pathological knowledge during pretraining or distil information among visual models \cite{zhou2026keep,ma2026distillation}: the emphasis here is on using teacher-derived, entity-specific evidence to govern image--text alignment. The representation comparison adds a practical architectural finding. Although the final projection was trained for cross-modal matching, the intermediate STL features provided the stronger readout for lesion recognition. Semantic matching and diagnostic transfer can thus favor different points along the same projection pathway; the present ablation identifies a useful interface without assigning its advantage to a single architectural factor.

The diagnostic evaluations show how this interface operates across levels of pathological assessment. On individual glomeruli, semantic transfer improved aggregate lesion recognition and supported distinctions in severity and morphological subtype. Evaluations on GloGS-1, GloGS-2, GloPro and GloGrade then examined the representation across independently sourced images, routine clinical sampling and an experimental mouse dataset. At the biopsy level, multiple-instance learning connected glomerular representations to disease classification, injury grading and baseline renal function estimation. These applications differ in their labels and units of analysis, yet all draw on information carried by glomerular morphology. Recent whole-slide multimodal models have demonstrated the value of linking pathology representations to broader slide-level tasks \cite{ding2025titan}; our findings suggest that an explicitly defined anatomical entity can also provide a useful bridge from lesion concepts to case-level assessment.

Connecting GloSAGE features to a frozen language model offers another view of this bridge. With only a visual-prefix adapter trained, the system generated descriptions under a fixed instruction, demonstrating that the semantically organized representation can condition language output. This experiment is an exploratory extension of the visual representation rather than an evaluation of unrestricted visual question answering. Its overlap-based metrics assess agreement with structured reference descriptions, while the selected GPT-6-Astra examples illustrate differences in descriptive content without establishing comparative clinical accuracy. Together with advances in pathology vision--language and multimodal modeling \cite{lu2024conch,xiang2025musk}, the result points to a modular development strategy in which entity-level representations can be connected to language interfaces without replacing the underlying visual model.

Several considerations guide the next stage of this work. Teacher-derived supervision inherits the coverage and uncertainty of the teacher's lesion vocabulary, while broader clinical assessment will benefit from additional institutions, staining conditions and workflow settings. Glomerular representations also capture one important component of renal pathology rather than the full range of tissue compartments and clinical information considered in diagnosis. Finally, expert assessment of lesion accuracy, omissions and unsupported statements is needed to establish the usefulness of generated descriptions. These directions do not change the central finding: existing visual pathology models can supply image-specific knowledge for structured semantic alignment. Where suitable visual teachers, concept resources and observation records are available, this approach offers a path for extending foundation models from morphological encoding towards concept-based analysis and language-mediated interaction.

%% file: Section/methods.tex
\section{Methods}\label{methods}

\subsection{Dataset description}\label{sec:dataset_description}

We assembled eight cohorts from the Department of Pathology,
Xijing Hospital, Fourth Military Medical University, Xi'an, China and publicly available database. GloAtlas
provided expert-annotated glomerular images for lesion recognition and
grading; GloMillion provided the large-scale corpus used for visual
pretraining and semantic self-distillation; GloPro supported
prospective evaluation in routine practice; and RenalXJ and RenalKPMP linked
glomerular images with clinical and pathological reports for case-level
analysis.

Renal biopsy specimens were processed with hematoxylin and eosin (H\&E),
periodic acid--Schiff (PAS), Masson's trichrome (Masson), and periodic
acid--methenamine silver (PASM) staining. Whole-slide images (WSIs) were
screened at low resolution with a pretrained glomerular detector \cite{he2024identifying}, and the
detected glomeruli were extracted at $20\times$ magnification for model
development and evaluation. Full digitization, extraction, quality-control,
eligibility, annotation, and partitioning procedures are provided in
Supplementary Methods, Section~\ref{supp:data}.

\subsubsection{GloMillion}\label{dataset_glomillion}

GloMillion comprises 14,049 renal biopsy WSIs collected between 2022 and
2023 and contains more than one million extracted glomerular images. The
cohort underlies both the source visual representation and the semantic
self-distillation framework. GloPath-based lesion predictors generated
candidate lesion profiles, from which reliability-guided selection retained
more than 40,000 visual--semantic anchors. These lesion predictors had previously been trained using the
expert-annotated GloAtlas lesion labels. Although GloAtlas and
GloMillion comprised non-overlapping patients, GloAtlas therefore
contributed concept-level supervision indirectly through the teacher;
the GloMillion images used for semantic alignment were not the
GloAtlas evaluation images. GloAtlas and GloMillion contain
strictly non-overlapping patient cohorts.

\subsubsection{GloAtlas}\label{dataset_gloatlas}

GloAtlas comprises renal biopsy specimens from 870 cases collected between
2019 and 2021. Each case contributed six glass slides containing 12 serial
tissue sections across H\&E, PAS, Masson, and PASM staining. Two trained nephrologists independently annotated the
glomerular images, and a senior renal pathologist adjudicated disagreements.
GloAtlas served as the principal annotated cohort for lesion recognition and
severity grading. Dataset partitioning was performed at the patient level.

\subsubsection{GloPro}\label{dataset_gloprospect}

GloPro comprises 695 prospectively acquired PAS-stained renal biopsy slides collected through the routine diagnostic workflow at Xijing Hospital between 2022 and 2023. GloPro and GloMillion were mutually exclusive at the patient level:
no patient represented in GloPro contributed a slide or glomerular
image to GloMillion. Consequently, GloPro images were used in neither
GloPath visual pretraining nor GloSAGE semantic alignment. To facilitate real-world data acquisition, we established an integrated clinical workflow using the SQS 600P whole-slide scanner (Shenzhen Shengqiang Technology Co., Ltd., China) equipped with Innoviora-kidney, a deployed AI-assisted renal pathology analysis software. Following routine renal biopsy, tissue processing, and PAS staining, slides were digitized using the scanner and automatically processed by Innoviora-kidney, which performed glomerular detection and lesion classification using pretrained models. The resulting glomerular images and predictions were subsequently subjected to manual review and pathological assessment. In total, 13,749 glomerular images were collected and independently annotated by two nephrologists, with disagreements adjudicated by a senior renal pathologist. This prospective cohort provides a real-world benchmark for evaluating the robustness and clinical applicability of GloSAGE under routine diagnostic conditions.

\subsubsection{GloGS-1}\label{dataset_glogs1}

GloGS-1 was derived from the publicly available AIDPATH dataset hosted on Mendeley Data (\url{https://doi.org/10.17632/k7nvtgn2x6.3}, \cite{bueno2020data}) and was used to evaluate external generalization in glomerulosclerosis recognition. The dataset contains 2,340 individual glomerular images extracted from human PAS-stained kidney sections scanned at $20\times$ magnification, comprising 1,170 normal and 1,170 sclerotic glomeruli. The released image-level labels were retained for binary classification.

\subsubsection{GloGS-2}\label{dataset_glogs2}

GloGS-2 was constructed from the public dataset (\url{https://doi.org/10.5281/zenodo.4299694}) to assess glomerulosclerosis recognition across heterogeneous tissue sources and staining conditions. The source dataset comprises 78 annotated regions of interest from 20 human kidney WSIs originating from three medical centers and TCGA. Individual glomeruli were extracted using the released segmentation masks. Each mask-derived bounding box was expanded outward by 10\% to retain surrounding tissue context. The resulting dataset contained 1,397 glomerular images, including 272 globally sclerotic and 1,125 non-globally sclerotic glomeruli.

\subsubsection{GloGrade}\label{dataset_glograde}

GloGrade was derived from the public KidneyAI Dataset (\url{https://doi.org/10.5281/zenodo.17456927}, \cite{matsoukas2025streamlining}), released by investigators from AstraZeneca and collaborating institutions, to assess cross-species transfer to glomerular injury grading. The source dataset contains 3,928 PAS-stained glomerular images extracted from 30 WSIs in a BTBR ob/ob mouse model of diabetic nephropathy. Each image was independently assessed by three expert kidney pathologists using normal, mild, moderate, severe, and excluded categories. The final evaluation dataset comprised 3,668 images assigned to four ordered grades: normal ($n=757$), mild ($n=979$), moderate ($n=1,281$), and severe ($n=651$).

\subsubsection{RenalXJ}\label{dataset_renalxj}

RenalXJ comprises 744 renal biopsy cases with paired clinical and
pathological reports. Each case contributed 12 serial sections across four
stains, yielding 7,464 WSIs. Case-level labels represented IgA nephropathy
versus membranous nephropathy and mild, moderate, or severe renal tissue
injury. The cohort was partitioned at the case level into training,
validation, and test subsets in a 6:2:2 ratio.

\subsubsection{RenalKPMP}\label{dataset_renalkpmp}
RenalKPMP was derived from the Kidney Precision Medicine Project (KPMP), a multicenter initiative providing publicly accessible human kidney biopsy images and linked clinical data through the KPMP Repository (\url{https://atlas.kpmp.org/repository/}). This cohort was used to evaluate case-level discrimination between acute kidney injury (AKI) and chronic kidney disease (CKD), together with prediction of baseline estimated glomerular filtration rate (eGFR). The deep-feature analysis included 297 cases, comprising 62 AKI and 235 CKD cases. Baseline eGFR ranged from 22.18 to 141.64~mL/min/1.73~m$^2$, with a mean of $57.87 \pm 22.62$~mL/min/1.73~m$^2$ (mean $\pm$ standard deviation). The lesion-based analysis included 258 cases with available lesion-probability representations, comprising 57 AKI and 201 CKD cases. In this subset, baseline eGFR ranged from 25.93 to 141.64~mL/min/1.73~m$^2$, with a mean of $57.91 \pm 22.77$~mL/min/1.73~m$^2$. Glomerular representations were grouped by participant to form case-level bags, with disease category and baseline eGFR serving as the corresponding case-level targets.

\subsection{GloSAGE}\label{sec_glosage}

GloSAGE contains four connected components. Reliability-guided teacher
inference first identifies image--lesion profiles supported by the source
visual model in our previous research \cite{he2026glopath}. KEYS converts each retained profile into a structured
pathology description using the domain knowledge graph. EDGE then filters
candidate negative pairs according to the concepts actually assessed for
each image. Finally, PIRA aligns frozen morphological features with the
structured descriptions, and STL exposes an intermediate semantically
organized representation for supervised downstream analysis. The semantic
training corpus is derived from GloMillion, whereas GloAtlas, GloPro,
and RenalXJ and RenalKPMP provide complementary glomerular and case-level evaluation.

\subsubsection{Problem formulation}\label{sec:problem_formulation}

Glomeruli can contain multiple coexisting lesions, whereas any one staining
or annotation protocol observes only part of their pathological state. We
therefore formulated GloSAGE as knowledge-structured and evidence-aware
semantic self-distillation. A pretrained visual teacher supplies
image-specific lesion evidence; a nephropathology knowledge graph supplies
standardized concept descriptions; and an evidence-aware rule determines
which image--text pairs can validly contribute negative supervision.

For a test image $X$ with pathological state $Y$, define the frozen visual
representation $Z=f_{\omega}(X)$ and its normalized semantic projection
$U=g_{\theta}(Z)/\|g_{\theta}(Z)\|_2$. With the trained parameters held
fixed, these variables satisfy
\begin{equation}
Y\longrightarrow Z\longrightarrow U,
\qquad I(Y;U)\leq I(Y;Z),
\label{eq:representation_information}
\end{equation}
where $I(\cdot;\cdot)$ is mutual information. Semantic enrichment instead refers to reorganizing
the available information so that concept-based comparison and restricted
readouts can access it more effectively. This distinction allows improved
linear or cosine-similarity performance without implying improvement for an
unrestricted optimal predictor.

Let $\mathcal{D}=\{x_i\}_{i=1}^{N}$ denote glomerular images,
$f_{\omega}:\mathcal{X}\rightarrow\mathbb{R}^{d}$ a frozen visual encoder,
and $g_{\theta}:\mathbb{R}^{d}\rightarrow\mathbb{R}^{q}$ a trainable visual
projection. For each image, a pretrained lesion head produces scores
$\boldsymbol{p}_i$. Reliability-guided selection $\mathcal{R}$ retains a
positive lesion profile $\widehat{Y}_i$ within the observed concept set
$S_i$, and the knowledge-guided serializer $\mathcal{T}_{\mathcal{K}}$
converts that profile into text:
\begin{equation}
\begin{aligned}
\boldsymbol{p}_i &= h_{\eta}(f_{\omega}(x_i)),\\
(\widehat{Y}_i,r_i) &= \mathcal{R}(\boldsymbol{p}_i,S_i),
\qquad \widehat{Y}_i\subseteq S_i,\\
t_i &= \mathcal{T}_{\mathcal{K}}(\widehat{Y}_i),
\qquad
\mathcal{D}_{\mathrm{sem}}
=\{(x_i,t_i,S_i,\widehat{Y}_i):r_i=1\}.
\end{aligned}
\label{eq:semantic_supervision}
\end{equation}
Here, an unrecorded lesion is not assumed to be absent, and a selected
pseudo-label is not treated as manually verified ground truth. GloSAGE
reorganizes information already present in the frozen morphological
features into a concept-aligned space; a deterministic projection does not
increase the total information in the source representation, but can make
that information more accessible to cosine-similarity or restricted
downstream readouts.

Let $e_{\phi}$ be the text encoder and its text-side projection. The
normalized image and text representations define
$s_{ij}=\boldsymbol{u}_i^{\mathsf T}\boldsymbol{v}_j$, where
$\boldsymbol{v}_j=e_{\phi}(t_j)/\|e_{\phi}(t_j)\|_2$. A candidate text was
allowed to act as a negative only if the observation predicate supported
that relationship. This is essential in partially observed multi-label
pathology: a non-matching description can share a positive lesion with the
image or describe a lesion that was never examined.

The role of alignment can also be stated geometrically. Let
$\boldsymbol{v}^{+}$ and $\boldsymbol{v}^{-}$ be unit-normalized embeddings
of a correct description and a valid alternative, and let
$\delta=1-(\boldsymbol{v}^{+})^{\mathsf T}\boldsymbol{v}^{-}>0$. If a
unit-normalized image representation satisfies
$\|\boldsymbol{u}-\boldsymbol{v}^{+}\|_2\leq\varepsilon$, then
\begin{equation}
s(\boldsymbol{u},\boldsymbol{v}^{+})-
s(\boldsymbol{u},\boldsymbol{v}^{-})
\geq \delta-\varepsilon\sqrt{2\delta}>0
\quad\text{when}\quad
\varepsilon<\sqrt{\delta/2}.
\label{eq:semantic_margin_bound}
\end{equation}
This bound separates three empirical requirements addressed by
GloSAGE: reliable image--text correspondence, discriminative text targets,
and valid negative relations.

\subsubsection{Visual backbone}\label{sec:visual_backbone}

The visual backbone was GloPath, an entity-centric ViT-B model with 12
Transformer blocks pretrained with DINO on more than one million glomerular
images~\cite{he2026glopath,dosovitskiy2020image,caron2021emerging}. For image $x_i$, the frozen encoder and trainable projection
produce
\begin{equation}
\boldsymbol{z}_i=\operatorname{sg}[f_{\omega}(x_i)],
\qquad
\boldsymbol{u}_i=
\frac{g_{\theta}(\boldsymbol{z}_i)}
{\|g_{\theta}(\boldsymbol{z}_i)\|_2},
\label{eq:visual_backbone}
\end{equation}
where $\operatorname{sg}$ denotes stop-gradient. The source visual encoder
remained fixed during semantic alignment.

\subsubsection{Knowledge-structured pathology text synthesis}\label{sec:pseudo_text}

Knowledge-Structured Pathology Text Synthesis (KEYS) converted reliable
teacher predictions into structured descriptions. Candidate outputs from
the 27 stain-specific lesion predictors were mapped to an ordered vocabulary
of 12 lesion groups. Positive groups were linked to a nephropathology
knowledge graph containing 24 nodes and 60 edges. Each node stored a
standardized name, anatomical compartment, synonyms,
definition, and hierarchy path; graph relations represented hierarchy,
co-occurrence, candidate progression, and mutually exclusive findings.

The graph supplied domain-level semantic structure, whereas the teacher
supplied image-specific evidence. Definitions and synonyms therefore
described the selected concepts but were not treated as separately observed
attributes of the image. Likewise, graph relations were available to the
pair-construction procedure only when explicitly invoked; their presence in
the graph did not create additional positive findings or independent loss
terms.

For every retained profile, KEYS serialized the corresponding lesion names,
definitions, and up to three synonyms in a fixed order. Empty profiles were
described as containing no positive findings among the assessed lesion set,
rather than as normal glomeruli. Details are in Supplementary Sec. ~\ref{supp:keys}.

\subsubsection{Evidence-guided negative-pair construction}\label{sec:negative_construction}

Evidence-Guided Negative-pair Construction (EDGE) addresses false-negative
supervision in partially observed multi-label pathology. Standard
contrastive learning treats other descriptions in a batch as negatives,
but this assignment is unreliable when glomeruli share lesions or when a
candidate lesion was not included in the observation protocol. In contrast
to approaches that infer possible false negatives from embedding
similarity~\cite{huynh2022boosting}, EDGE uses the recorded observation
process before a pair enters the objective. For concept
$\ell$, the recorded state of image $i$ was
\begin{equation}
o_i(\ell)=
\begin{cases}
+1, & \ell\in\widehat{Y}_i,\\
-1, & \ell\in S_i\setminus\widehat{Y}_i,\\
\bot, & \ell\notin S_i,
\end{cases}
\label{eq:observation_states}
\end{equation}
where $\bot$ denotes an unavailable assessment. A candidate description was
retained as a negative for image $i$ only when every lesion coordinate used
to justify that assignment had been assessed and recorded as negative:
\begin{equation}
A_{ij}=
\mathbf{1}[\Lambda_{ij}\neq\varnothing]
\mathbf{1}[\Lambda_{ij}\subseteq S_i]
\mathbf{1}[\Lambda_{ij}\cap\widehat{Y}_i=\varnothing].
\label{eq:certified_negative_mask}
\end{equation}
Thus, candidate pairs based on an unexamined lesion or contradicted by a
recorded positive lesion were removed from repulsive supervision, but were
not promoted to positive pairs. Details are
provided in Supplementary Methods, Section~\ref{supp:edge}.

This distinction separates unsupported assignments from contradicted ones.
For example, a crescent-bearing description cannot be certified as a
negative for an image whose crescent status was not examined; if crescent is
recorded as positive, the proposed negative is directly contradicted. Only
an explicitly assessed negative crescent coordinate supports repulsion on
that basis. Certification is conditional on the available record and does
not eliminate errors already present in a pseudo-label.

\subsubsection{Pathology-informed representation alignment}\label{sec:contrastive_learning}

Pathology-Informed Representation Alignment (PIRA) aligned the projected
visual features with PubMedBERT text features~\cite{gu2021pubmedbert}.
For a mini-batch of $B$ paired images and descriptions, normalized features
$\boldsymbol{u}_i$ and $\boldsymbol{v}_j$ defined logits
$\ell_{ij}=\boldsymbol{u}_i^{\mathsf T}\boldsymbol{v}_j/\tau$. The comparison
mask retained the paired positives and EDGE-certified negatives. Training
used the symmetric masked CLIP objective~\cite{radford2021clip}:
\begin{equation}
\begin{aligned}
\mathcal{L}_{v\rightarrow t}
&=-\frac{1}{B}\sum_i
\log\frac{\exp(\ell_{ii})}{\sum_j M_{ij}\exp(\ell_{ij})},\\
\mathcal{L}_{t\rightarrow v}
&=-\frac{1}{B}\sum_i
\log\frac{\exp(\ell_{ii})}{\sum_j M_{ji}\exp(\ell_{ji})},\\
\mathcal{L}_{\mathrm{align}}
&=\tfrac{1}{2}(\mathcal{L}_{v\rightarrow t}+\mathcal{L}_{t\rightarrow v}).
\end{aligned}
\label{eq:clip_alignment_loss}
\end{equation}

Specifically, the mask used in the contrastive objective was
\begin{equation}
M_{ij}=\mathbb{1}[i=j]+\mathbb{1}[i\ne j]A_{ij},
\end{equation}
where $A_{ij}$ is the off-diagonal negative-pair admissibility
indicator defined in Eq.~(6). Thus, paired positives were always
retained. Excluded candidates did not enter the corresponding softmax denominator.
The knowledge graph and evidence mask acted through different mechanisms:
node text determined the semantic targets, whereas observation-aware
certification determined the admissible comparison set. Positive pairs
remained the paired image and description; shared or excluded off-diagonal
pairs were not automatically promoted to positives.

For an eligible image-to-text pair, the similarity gradient is the usual
softmax contrastive gradient; for a masked pair it is zero. EDGE therefore
changes the set of repulsive comparisons without changing the paired
positive target or adding a separate graph loss. The reverse text-to-image
term uses the transposed mask, so the same evidence decision is respected in
both directions. An anchor with no retained alternative contributes no
directional repulsion beyond its paired positive.

The Semantic Transfer Layer (STL), defined as the first visual projection
layer followed by GELU activation \cite{hendrycks2016gaussian}, supplied GloSAGE features to supervised
downstream tasks. Network dimensions, tokenization, sampling, optimization,
and the complete masked-gradient formulation are given in Supplementary
Methods, Section~\ref{supp:architecture}.

\subsection{Evaluation}

The evaluation followed a progression from semantic alignment to diagnostic
utility and language-mediated interpretation. Classifier-free experiments
first examined whether glomerular morphology corresponded to pathological
concepts in the joint embedding space. Supervised recognition and grading
then assessed whether semantic transfer preserved or strengthened
lesion-level discrimination, while case-level multiple-instance learning
(MIL) tested whether these features supported disease classification and
injury assessment across heterogeneous glomeruli. Ablations examined the
contributions of knowledge-structured text, evidence-aware negatives, and
the transfer layer. Finally, a language-model extension explored whether
the resulting representations could support image-conditioned descriptions.

\subsubsection{Comparator models}

We evaluated GloSAGE alongside UNI\cite{chen2024uni},
CONCH\cite{lu2024conch}, PLIP\cite{huang2023visual}, and
KEEP\cite{zhou2026keep}. UNI served as an image-only baseline, whereas
CONCH and PLIP supplied pretrained visual--language representations.
KEEP was included as a knowledge-enhanced comparator because it uses
a disease knowledge graph to organize image--text pairs into semantic
groups and guide vision--language pretraining. It therefore provided a
relevant comparison for evaluating glomerulus-specific semantic
self-distillation against broader knowledge-enhanced pretraining.
GloPath was retained as the source visual baseline to isolate the
contribution of semantic transfer\cite{he2026glopath}. All models used
the same stain-specific task definitions; image--text evaluations used
the corresponding paired encoders where available. Checkpoint,
preprocessing, and model-adaptation details are provided in Supplementary
Methods, Section~\ref{supp:tasks}.

\subsubsection{Semantic understanding}

Semantic understanding was evaluated on GloAtlas across PAS, Masson, and
PASM through four complementary tasks, examining concept recognition,
cross-modal correspondence, fine-grained specificity, and sensitivity
to changes in pathological meaning.

Zero-shot lesion recognition assigned independent lesion scores using
cosine similarities between normalized image embeddings and predefined
lesion-description embeddings, testing concept recognition without a
trained classifier. Here, zero-shot refers to inference without fitting an additional
lesion classifier to the evaluated images. It does not imply that the
lesion concepts were unseen during model development: the visual
teacher used to generate alignment targets had been trained on
GloAtlas lesion annotations. For the GloPath visual-only control, lesion descriptions were encoded independently with PubMedBERT; masked mean pooling of token representations yielded 768-dimensional text vectors. These vectors and the 768-dimensional GloPath image features were separately normalized and compared directly by cosine similarity, without a trained image--text alignment layer. Bidirectional retrieval ranked descriptions for
an image query and images for a text query, with semantic-group identity
defining valid matches, to assess correspondence across modalities.
Hard-negative discrimination tested whether a lesion-positive image
ranked its matching description above five related but incorrect
alternatives, probing discrimination between closely related concepts.
Concept intervention kept the image fixed and replaced the lesion-bearing
description, measuring whether the similarity response distinguished
lesion-positive from lesion-negative images.

GloPath was included only in zero-shot recognition as an unaligned visual-only negative control; it was excluded from the other semantic analyses because it has no jointly trained image--text embedding space. Full candidate construction, aggregation, and metric definitions are
given in Supplementary Methods, Section~\ref{supp:semantic_eval}.
Zero-shot recognition, retrieval, and hard-negative analyses used the
available stain-specific manifests, including their assigned training,
validation, and test records; concept intervention used the test
partition. These analyses required no additional classifier fitting,
but not all constitute independent held-out evaluation.

\subsubsection{Glomerular and case-level diagnosis}

Diagnostic evaluation assessed whether semantic transfer supported clinically relevant assessment at both the individual-glomerulus and biopsy-case levels. On GloAtlas, 27 lesion-recognition tasks (11 PAS, six Masson, and ten PASM) tested the identification of coexisting morphological abnormalities. Ten grading tasks assessed severity or clinically meaningful lesion subtypes: mesangial hypercellularity severity, crescent composition, and segmental versus global involvement of membranoproliferative and endocapillary lesions. Task-specific classification heads were trained using the corresponding training partitions, with checkpoint selection based on validation ROC-AUC. Together, these tasks evaluated the morphological distinctions that underpin renal biopsy interpretation.

Case-level evaluation on RenalXJ comprised IgA nephropathy versus membranous nephropathy classification and three-class renal injury grading, testing whether glomerular representations could support disease differentiation and integrated assessment of tissue injury. Two forms of instance-level representation were evaluated: \emph{deep features}, extracted directly by the foundation model, and \emph{lesion-based features}, comprising glomerular lesion-probability vectors predicted by lesion classifiers developed from the foundation model. Each representation was grouped into case bags and aggregated separately by attention-based MIL (implemented at \url{https://github.com/lingxitong/MIL\_BASELINE}), allowing heterogeneous glomerular findings to contribute to a biopsy-level prediction. Feature extraction and lesion-probability inference were performed with the corresponding models frozen. 3-fold validation was performed at the case level, preventing images from the same case from appearing across partitions. Attention weights were used to inspect high-contribution glomeruli rather than as independently validated localization annotations. Complete task definitions, classifier protocols, and MIL settings are reported in Supplementary Methods, Section~\ref{supp:architecture}.

\subsubsection{Ablation study}

Ablations on GloAtlas examined the intended contributions of the three
semantic components. EDGE was designed to exclude negative pairs
unsupported or contradicted by observed lesion evidence, whereas KEYS
provided structured pathological descriptions grounded in the lesion
knowledge graph. Removing EDGE (w/o EDGE), KEYS (w/o KEYS), or both
(w/o EDGE and KEYS) tested their individual and combined contributions
under the same downstream protocol. An architectural ablation compared
STL, comprising the first visual projection layer and GELU activation,
with the complete projection head (GloSAGE-Fullproj), testing whether
the intermediate representation more effectively transferred semantic
alignment to supervised diagnosis.

\subsection{Language-related task extension}

An exploratory language-model extension assessed image-conditioned
generation of glomerular morphology and lesion descriptions. A trainable
adapter comprising a linear layer, GELU, and LayerNorm mapped each image
representation to eight visual-prefix embeddings, which were prepended
to the instruction embeddings of Qwen2.5-0.5B-Instruct. The visual encoder
and language model remained frozen; only the adapter was optimized using
autoregressive cross-entropy on reference-answer tokens, with prefix,
instruction, and padding positions excluded from the loss. Images were
paired with a fixed descriptive instruction and knowledge-graph-derived
reference descriptions, preserving the predefined data partitions.
Checkpoints were selected by validation loss, and test descriptions
were generated deterministically. Selected examples were additionally compared qualitatively with
GPT-6-Astra using the same images and fixed descriptive question,
without additional task-specific prompts or reference descriptions.
This exploratory comparison examined lesion-description content
rather than establishing quantitative performance equivalence.
Adapter training, decoding settings, and the GPT-6-Astra comparison
procedure are detailed in Supplementary Methods,
Section~\ref{sm_gpt6}.

\subsubsection{Quantification and statistical analysis}

Diagnostic discrimination was quantified by ROC-AUC. Retrieval used
recall at ranks 1, 5, and 10. Hard-negative accuracy was the proportion
of images for which the matching description ranked first.
Concept-intervention effect was the similarity difference between
original and perturbed descriptions; intervention-discrimination
ROC-AUC assessed whether this difference separated lesion-positive
from lesion-negative images. Generated descriptions were evaluated with ROUGE-1, ROUGE-2, and ROUGE-L, which quantify lexical and sequence overlap with reference descriptions.

Repeated-run results were summarized as means and standard deviations
under the selection rules stated for each analysis. Comparisons across
matched lesion tasks used nonparametric paired analyses, with
multiplicity control where reported. Details are provided in Supplementary Methods,
Section~\ref{supp:statistics}. Training hyperparameters and the
hardware and software environment are provided in
Sections~\ref{supp:architecture} and~\ref{supp:hardware}, respectively.

%% file: Section/appendix.tex
\section*{Acknowledgements}

This study was supported by National Natural Science Foundation of China (82572986), the Research Funds of Jinfeng Laboratory (JFLKYXM202503AZ-203), National Natural Science Foundation of China (82430062), National High Level Hospital Clinical Research Funding (BJ-2024-155), Natural Science Foundation of Fujian Province (2026J01310023), Shenzhen Engineering Research Centre (XMHT20230115004), Beijing Municipal Health Commission Research Ward Excellence Clinical Research Program (BRWEP2024WO32240114), and the Fuzhou University Talent-Introduction
Scientific Startup Fund (XRC-25119). We also acknowledge the creators of the
AIDPATH glomerular dataset, the publicly released kidney glomerular ROI
dataset, and the KidneyAI dataset for making their data available. The
results reported here are in part based on data generated by the Kidney
Precision Medicine Project (KPMP) and codes generated by \url{https://github.com/lingxitong/MIL_BASELINE}.

\subsection*{Ethics approval}
The use of institutional renal biopsy images and associated clinical data in this study was approved by the Medical Ethics Committee of the First Affiliated Hospital of the Fourth Military Medical University (approval no. KY20243325-1). The study was conducted in accordance with the approved research protocol and informed consent documents. Publicly available datasets were used in accordance with their respective access conditions and original ethical approvals.

\section*{Author contributions}

Qiming He, Jing Li, Zihan Wang, Tian Guan, Chao He, Zhe Wang, and Yonghong He
conceived and designed the study. Chao He, Zhe Wang, and Yonghong He
supervised the project and provided overall guidance. Qiming He performed
data processing and model development. Jing Li led dataset construction,
pathological annotation, and result interpretation. Qiming He, Jing Li,
and Tian Guan drafted the manuscript; Tian Guan also revised and polished
the text. Yifei Ma and Zimo Zhao contributed to code development and
experimental analysis. Xitong Ling, Minxi Ouyang and Yujie Xie contributed to model
validation. Yanxia Wang and Hongjing Chen contributed to dataset
construction and annotation. Yexing Zhang, Yizhi Wang, Xinrui Chen,
Lianghui Zhu, Yiqing Liu, and Junru Cheng assisted with experiments and
code organization. Qingxia Hou, Shuyan Zhao, Xiaoqin Wang, and Lili Ma
contributed to dataset curation. Qiang Huang, Zhiyuan Shen,
and Siqi Zeng provided equipment and research resources. All authors
reviewed and approved the manuscript.

\section*{Competing interests}
The authors declare no competing interests.

\section*{Data availability}
Access for the institutional datasets may be requested from the
corresponding author and is subject to approval by the data-holding
institution, applicable ethics requirements, and a data-use agreement.
The public datasets are available from their respective repositories:
the AIDPATH glomerular image dataset (GloGS-1) at
\url{https://data.mendeley.com/datasets/k7nvtgn2x6/3};
the kidney glomerular ROI dataset used to derive GloGS-2 at
\url{https://zenodo.org/records/4299694};
the KidneyAI dataset used for GloGrade at
\url{https://zenodo.org/records/17456927}; and
Kidney Precision Medicine Project (KPMP) data used for RenalKPMP at
\url{https://atlas.kpmp.org/repository}.

\section*{Code availability}

Code and model weights are released at \hyperlink{https://github.com/jiangnansss/Code-for-GloSAGE}{https://github.com/jiangnansss/Code-for-GloSAGE}.

\section*{Supplementary information}

See Supplementary Materials for details of Sections, Figures and Tables.

%% file: Section/supplementary.tex
\appendix
\clearpage

\section{Supplementary Methods}\label{supp:methods}

\subsection{Cohort eligibility, slide digitization, and image quality control}\label{supp:data}


All institutional cohorts were assembled at the Department of Pathology,
Xijing Hospital, Air Force Medical University. Renal biopsy sections were stained with H\&E, PAS, Masson's trichrome and PASM to provide complementary views of renal tissue morphology. H\&E reveals tissue architecture and cellular changes; PAS highlights mesangial matrix and basement membranes; Masson's trichrome distinguishes collagen deposition and fibrosis; and PASM delineates glomerular basement membranes, facilitating assessment of capillary-wall alterations. Together, these stains support comprehensive evaluation of glomerular and tubulointerstitial lesions. Slides were digitized with an SQS-600P slide imaging system of Shenzhen Shengqiang
Technology Co., Ltd., China. For glomerulus-level analysis, WSIs were read
with OpenSdpc (\hyperlink{https://github.com/WonderLandxD/opensdpc}{https://github.com/WonderLandxD/opensdpc}) and downsampled to an effective magnification of $5\times$ ($1.68\,\mu\mathrm{m/pixel}$). A pretrained glomerular detector localized
candidate glomeruli, and the corresponding regions were extracted at
$20\times$ ($0.42\,\mu\mathrm{m/pixel}$). This entity-level extraction kept
the complete detected glomerulus as the unit used for representation
learning and downstream analysis. Samples assigned to validation or test
sets were excluded from the corresponding training procedures, and all
partitions were constructed at the patient or WSI level specified below to
avoid image-level leakage.

GloAtlas included 870 renal biopsy cases collected from 2019 to 2021.
Each case contributed six glass slides containing 12 serial tissue
sections, yielding 5,220 digitized slides. The cohort encompassed diabetic
nephropathy, focal segmental glomerulosclerosis, antineutrophil cytoplasmic
antibody-associated glomerulonephritis, membranous nephropathy,
obesity-related glomerulopathy, anti-glomerular basement membrane disease,
lupus nephritis, endocapillary proliferative glomerulonephritis,
membranoproliferative glomerulonephritis, tubulointerstitial nephritis,
crescentic glomerulonephritis, IgA nephropathy, and IgA vasculitis nephritis. Glomeruli were localized in whole-slide images using a pretrained
glomerular detector and extracted at $20\times$ magnification for
lesion annotation. The annotation scheme comprised 27 stain--lesion
combinations: 11 for PAS (MH, PAS(+), PAS(-), Crum, ADH, SS, MN, MP,
EP, Cre, and GS), six for Masson's trichrome (MH, ADH, SS, SFN, Cre,
and GS), and ten for PASM (MH, PAS(-), ADH, SS, SFN, MN, MP, EP,
Cre, and GS). Where applicable, annotations additionally distinguished
mesangial hypercellularity severity, segmental versus global involvement
of membranoproliferative and endocapillary lesions, and crescent subtypes.
The morphological definitions summarized in
Supplementary Table~\ref{sm_lesion_definitions} reference the Renal
Pathology Society consensus, the Oxford classification, and the revised
ISN/RPS definitions for crescent composition \cite{roberts2009oxford, haas2020consensus, bajema2018revision}.
Yanxia Wang and Xiaoqin Wang independently annotated glomerular lesions
using the Kidney Labeling Tool according to standardized diagnostic
criteria. Jing Li, a senior renal pathologist, reviewed disagreements
and established the final adjudicated reference labels. 

GloMillion included 14,049 renal biopsy WSIs collected from 2022 to 2023
and more than one million detected glomeruli. The slides were organized
chronologically and did not carry detailed disease-subtype annotations.
GloMillion supplied the unlabeled morphology corpus for visual pretraining
and the candidate images for semantic self-distillation. GloMillion and
GloAtlas were temporally separated.

GloPro included 695 PAS-stained renal biopsy slides and detected 13,749 glomerular
images acquired prospectively in the routine diagnostic workflow. Glomeruli were detected
and extracted with the same procedure used for GloAtlas. Yanxia Wang and
Xiaoqin Wang independently annotated the images, and Jing Li adjudicated
disagreements. The cohort was not used to construct semantic anchors and
was reserved for prospective assessment of lesion recognition.

RenalXJ included 744 cases, each represented by 12 serial sections:
four H\&E, four PAS, two Masson, and two PASM sections. Its 7,464 WSIs were digitized at a reported resolution of $0.105 \mu m/pixel$. Clinical and pathological reports were linked at the case level. Pathological diagnosis was restricted to IgA nephropathy and membranous nephropathy for binary disease classification;
renal tissue injury was categorized as mild, moderate, or severe.

\subsection{Knowledge graph and structured-text construction}
\label{supp:keys}

Recent studies have explored medical knowledge graphs as structured
sources of supervision for language-based interpretation. MedKA
transformed graph-encoded medical knowledge into a question--answer
corpus for language-model training~\cite{deng2025medka}, while KG-LLaVA
retrieved domain-specific graph knowledge to support natural-language
explanations of medical images~\cite{hamza2025llava}. These approaches
motivate the conversion of organized medical concepts into explicit
textual supervision. In GloSAGE, KEYS performs this conversion at the
glomerular lesion level, combining standardized concept names,
definitions, and synonyms into structured descriptions conditioned on
teacher-derived lesion profiles.

The nephropathology knowledge graph was represented as
$\mathcal{G}=(\mathcal{V},\mathcal{E},\mathcal{A})$, where $\mathcal{V}$
denotes the nodes, $\mathcal{E}$ the relation edges, and $\mathcal{A}$
the node attributes. It was stored as node, edge, and metadata records
and contained 24 nodes and 60 edges. Each node included a unique
identifier, names, anatomical compartment,
synonyms, definition, and hierarchy path. The root,
\texttt{KG\_Renal\_ROOT}, represented glomerular lesions; descendants
represented lesion groups and selected subtypes, including mesangial
grades and cellular, fibrocellular, and fibrous crescents. A fixed
profile-to-node map linked the 12 lesion groups to their concept
records. For example, \texttt{membranous\_nephropathy} mapped to
\texttt{KG\_Renal\_P11}.

Four graph relations were retained. \texttt{is\_a} encoded the
annotation hierarchy. \texttt{co\_occurs} represented statistical
lesion associations screened using lift, multiplicity-adjusted
significance, support, and sample size. \texttt{progresses\_to}
stored candidate directional relations; directions lacking
longitudinal support were marked as unverifiable and were not
interpreted causally. Mutually exclusive edges recorded the subtype
of the exclusion rule, including structural impossibility,
mutually exclusive trait, and suspect. Suspect relations were kept
distinct from hard structural exclusions. The graph topology
supported pair construction, whereas node names, definitions, and
synonyms supplied the text content; relation edges were not
automatically verbalized as image findings.

For each positive group, KEYS retrieved its English name, stored
definition, and first three synonyms. Let $\phi(c)$ map lesion
concept $c$ to its graph node, and let $n(v)$, $d(v)$, and $s_3(v)$
denote the English name, definition, and first three synonyms of
node $v$, respectively. For the positive lesion profile
$\mathcal{P}_i$ defined in Eq.~\ref{eq:semantic_supervision}, structured
text was constructed as
\begin{equation}
    t_i =
    \begin{cases}
        \mathcal{T}\!\left(N_i,D_i,S_i\right),
        & \mathcal{P}_i \neq \varnothing,\\
        t_{\mathrm{empty}},
        & \mathcal{P}_i = \varnothing,
    \end{cases}
    \qquad
    \begin{aligned}
        N_i &= \operatorname{Join}_{c\in\mathcal{P}_i}
                n\!\left(\phi(c)\right),\\
        D_i &= \operatorname{Join}_{c\in\mathcal{P}_i}
                d\!\left(\phi(c)\right),\\
        S_i &= \operatorname{Join}_{c\in\mathcal{P}_i}
                s_3\!\left(\phi(c)\right).
    \end{aligned}
    \label{eq:keys_text}
\end{equation}
Here, $\operatorname{Join}$ concatenates records in the fixed
vocabulary order, and $\mathcal{T}$ inserts the resulting fields
into the following template:
\begin{quote}
Glomerular lesion profile with positive findings:
\texttt{<lesion names>}.
Definitions: \texttt{<definitions>}.
Synonyms: \texttt{<synonyms>}.
\end{quote}

For an empty profile, $t_{\mathrm{empty}}$ was the exact template:
\begin{quote}
Glomerular lesion profile. No positive lesion groups were detected
among the measured renal glomerular findings. This patch is negative
for the assessed lesion set.
\end{quote}
This sentence refers only to the assessed coordinates and does not
label the glomerulus as globally normal. Because subthreshold
teacher probabilities do not establish verified absence, the
empty-profile wording should be understood as a description of
the recorded profile rather than independent confirmation of
biological negativity.

As an example, a profile containing mesangial lesion and
endocapillary proliferation was rendered as the two English
concept names followed by their stored definitions and the three
stored synonyms for each concept. The same structured text bank was used to construct
the image--text targets and the language-model reference answers.

\subsection{Evidence-guided negative construction}
\label{supp:edge}

Negative-pair construction requires particular care when medical images
share pathological findings or have partially observed lesion labels.
Recent work has addressed false negatives in medical vision--language
learning using graph-informed semantic targets~\cite{wei2025relaxing},
while partial-label learning has highlighted the need to distinguish
unannotated positives from genuine negatives~\cite{lyu2025addressing}.
Motivated by these issues, EDGE certifies candidate negative pairs
against the available lesion evidence, excluding assignments that are
unsupported by observation or contradicted by recorded positive findings.

For image $i$, EDGE represented each lesion coordinate $\ell$ as
recorded positive, assessed negative, or unexamined. Let $S_i$ denote
the set of assessed coordinates and $\widehat{Y}_i\subseteq S_i$ the
set of recorded positive coordinates. A candidate negative $(i,j)$
was associated with a non-empty evidence set $\Lambda_{ij}$ containing
the coordinates on which its negative assignment depended. Its
admissibility was defined as
\begin{equation}
    a_{ij}
    =
    \mathbf{1}\!\left[\Lambda_{ij}\subseteq S_i\right]
    \mathbf{1}\!\left[
        \Lambda_{ij}\cap\widehat{Y}_i=\varnothing
    \right],
    \label{eq:edge_admissibility}
\end{equation}
where $\mathbf{1}[\cdot]$ denotes the indicator function.
Two failure modes were distinguished. An unsupported negative used
at least one coordinate absent from $S_i$; a contradicted negative
used a coordinate already positive in $\widehat{Y}_i$. The unexamined
check was applied first so that the two diagnostic categories
remained disjoint.

Pair-level elimination evaluated each ordered image--description
candidate separately. Profile-level elimination evaluated the
description profile against the same observation rule and removed
every pair generated from a profile when certification failed.
Sequential application was monotonic: a pair removed by the first
rule could not be reintroduced by the second. Unknown coordinates
supported neither disagreement nor overlap. Removed pairs
contributed neither attraction nor repulsion and were never
relabeled as positives. Structural reachability and mutually
exclusive relations were computed at the profile level;
suspect-exclusive relations were excluded from the default
hard-exclusion matrix. These rules also applied to empty and
partially observed profiles without treating missing values
as negatives.

\subsection{Network architecture and optimization}\label{supp:architecture}

The visual branch used the frozen 768-dimensional GloPath ViT-B feature \cite{he2026glopath,dosovitskiy2020image}.
Its projection head was a 768-to-1,024 linear layer, GELU activation \cite{hendrycks2016gaussian}, and a
1,024-to-512 linear layer. The text branch used PubmedBERT\cite{gu2021pubmedbert}; text
was tokenized with the associated WordPiece tokenizer \cite{song2021fast}, truncated to 64
tokens, and mapped through a projection head with the same 1,024-dimensional
hidden and 512-dimensional output configuration. Image and text outputs
were $L_2$ normalized. The visual encoder remained frozen, and semantic
alignment optimized the projection heads.

Contrastive pretraining which adopt CLIP \cite{radford2021clip} ran for 10 epochs with learning rate $1\times10^{-3}$, weight decay $1\times10^{-4}$, and temperature 0.07. Each
batch sampled 16 semantic groups and four images per group, giving 64 images.
The symmetric image-to-text and text-to-image loss used the same
EDGE-derived comparison mask, transposed for the reverse direction. A
masked candidate was assigned no probability mass in the corresponding
softmax and therefore produced zero direct pairwise gradient. When every
off-diagonal candidate was retained, the objective reduced to the standard
CLIP loss.

For downstream lesion recognition, grading, and recognition ablations,
images were resized to $224\times224$ pixels and normalized by ImageNet
channel statistics \cite{deng2009imagenet}. Augmentation comprised horizontal and vertical flips,
rotation, shift--scale--rotation, and contrast adjustment. Adam optimization
used learning rate $1\times10^{-5}$, weight decay $1\times10^{-4}$, batch
size eight, and at most 50 epochs. Validation ROC-AUC selected checkpoints.
The standard GloSAGE representation was the 1,024-dimensional activation
after the first visual-projection linear layer and GELU; GloSAGE-P used the
complete 512-dimensional projection output.

For case-level tasks, the ABMIL \cite{ilse2018attention} feature layer had width 512, the attention
network had width 128, and dropout was 0.1. The model processed one case bag
per optimization step and was trained with Adam for 30 epochs using learning
rate $2\times10^{-4}$ and weight decay $1\times10^{-5}$. 3-fold
validation used seed 2024; 20\% of each fold's
training portion was reserved for validation. Two instance-level representations were evaluated separately. For the \emph{deep-feature} representation, each preprocessed glomerular image was passed through the corresponding frozen foundation model to extract its visual embedding; GloSAGE used the STL output. For the \emph{lesion-based} representation, each glomerulus was represented by a 15-dimensional vector comprising a three-dimensional one-hot stain indicator (PAS, Masson, or PASM) and 12 lesion probabilities predicted by the corresponding foundation-model-based lesion classifiers. The lesion coordinates followed the fixed order MH, PAS(+), PAS($-$), Crum, ADH, SS, SFN, MN, MP, EP, Cre, and GS. Glomerular vectors from the same biopsy case formed a bag, which was supplied to ABMIL for case-level prediction. Attention weights were used
only to visualize representative glomeruli and were not treated as
independently validated localization labels.

\subsection{Evaluation models}\label{supp:tasks}

The comparison included GloSAGE, its source visual model GloPath,
UNI, CONCH, PLIP, and KEEP. Together, these models represented
glomerulus-specific visual pretraining, general-purpose pathology
visual pretraining, and pathology vision--language pretraining.
Parameter counts below indicate approximate encoder sizes rather
than task-specific trainable parameters.

GloPath~\cite{he2026glopath} used a DINO-pretrained ViT-B/16
visual backbone, with approximately 86 million parameters,
trained on 1,017,879 glomerular images extracted from 14,049
renal biopsy WSIs. GloSAGE retained this visual backbone and
used reliable teacher-derived lesion profiles from GloMillion
to construct knowledge-structured text supervision. Its text
pathway used PubMedBERT, a BERT-base encoder with approximately
110 million parameters. Semantic alignment trained the image
and text projection heads while keeping both encoders frozen.

UNI~\cite{chen2024uni} served as a general-purpose image-only
pathology baseline. It used a ViT-L/16 backbone with approximately
300 million parameters and was pretrained using DINOv2 on
Mass-100K, comprising approximately 100 million tissue patches
from more than 100,000 WSIs across 20 major tissue types.
UNI was included in supervised and MIL evaluations but excluded
from cross-modal semantic experiments because it does not
provide a pretrained joint image--text embedding space.

CONCH~\cite{lu2024conch} used a CoCa-based \cite{yu2022coca} architecture trained
with contrastive and captioning objectives on approximately
1.17 million human histopathology image--caption pairs.
The released model combined a ViT-B/16 visual encoder
(approximately 90 million parameters) with a 12-layer,
768-dimensional text encoder (approximately 110 million
parameters). We used the \texttt{conch\_ViT-B-16} architecture
and its contrastive projection for semantic matching.

PLIP~\cite{huang2023visual} was obtained by adapting a pretrained
CLIP model to OpenPath, a dataset containing 208,414 pathology
image--text pairs collected from public online sources,
including medical Twitter. Its released architecture comprised
a ViT-B/32 visual encoder with approximately 88 million
parameters and a 12-layer, 512-dimensional text Transformer
with approximately 63 million parameters. We used the released
CLIP processor and image and text projections.

KEEP~\cite{zhou2026keep} provided a knowledge-enhanced
vision--language comparator. It used a UNI-initialized
ViT-L/16 visual encoder with approximately 300 million
parameters and a PubMedBERT-based text encoder with
approximately 110 million parameters. Its disease knowledge
graph contained 11,454 disease entities and 139,143 attributes,
and guided the organization of image--text data from OpenPath
and Quilt-1M into approximately 143,000 semantic groups.
The text encoder underwent disease-knowledge pretraining
before image--text alignment. We used the released image
transformation and text tokenizer.

Semantic experiments used each multimodal model's released
image and text pathways with model-specific preprocessing.
For GloSAGE, semantic matching used the complete
512-dimensional image and text projection heads, whereas
supervised downstream tasks used STL. Visual representations
were evaluated under the corresponding supervised and
case-level MIL protocols described below.

\subsection{Semantic evaluation protocols}
\label{supp:semantic_eval}

Semantic evaluation examined concept recognition, bidirectional retrieval,
discrimination between related concepts, and sensitivity to targeted
textual changes. Following the joint-embedding evaluation paradigm used
in pathology vision--language models~\cite{lu2024conch}, image and text
representations were normalized before matching. For image $x$ and
description $t$, the matching score was defined as
\begin{equation}
    s(x,t)
    =
    \frac{f_{\mathrm{img}}(x)^{\top}f_{\mathrm{text}}(t)}
    {\|f_{\mathrm{img}}(x)\|_2
     \|f_{\mathrm{text}}(t)\|_2},
    \label{eq:semantic_similarity}
\end{equation}
where $f_{\mathrm{img}}$ and $f_{\mathrm{text}}$ denote the
model-specific image and text embedding pathways.

\paragraph{Zero-shot lesion recognition.}
Every image included in the evaluation was compared with the
stain-specific lesion descriptions using Eq.~\ref{eq:semantic_similarity}.
Each lesion received an independent similarity score, allowing
coexisting findings to be evaluated without imposing mutually
exclusive predictions. No additional classifier or decision threshold
was fitted. Lesion-wise ROC-AUC was computed from the binary reference
labels, and stain-level performance was the unweighted mean over the
standard recognition panel.

\paragraph{Bidirectional image--text retrieval.}
Image-to-text retrieval ranked candidate semantic-group descriptions
for each image. Text-to-image retrieval ranked images for each
description, with every image sharing the query's semantic-group
identifier counted as a valid match. Recall at ranks 1, 5, and 10
was computed in each direction, with a query counted as successful
when at least one valid match appeared among the top-$k$ candidates.
Zero-shot recognition and retrieval used the stain-specific manifests
and their available training, validation, and test records. These
analyses were classifier-free evaluations of the semantic space
rather than estimates from an independent held-out cohort.

\paragraph{Hard-negative discrimination.}
Each lesion-positive image was paired with its matching description
and five related but incorrect descriptions. Top-1 accuracy was the
proportion of images for which the matching description ranked first
by cosine similarity; chance performance was $1/6$. The hard-negative
panels also contained explicitly labelled stress-test concepts beyond
the standard 27-task recognition panel, as identified in the
corresponding Supplementary Tables.

\paragraph{Concept intervention.}
The image was held fixed while the original lesion-bearing
description was replaced with a competing concept description.
The intervention effect for image $x$ was defined as
\begin{equation}
    \Delta s(x)
    =
    s(x,t_{\mathrm{original}})
    -
    s(x,t_{\mathrm{intervened}}).
    \label{eq:concept_intervention}
\end{equation}
A positive value indicated higher similarity to the original
description than to the intervened description. Mean effect size
was calculated on images positive for the original lesion.
Intervention-discrimination ROC-AUC assessed whether $\Delta s$
distinguished lesion-positive from lesion-negative images.
It therefore quantified discrimination using the similarity-change
score, rather than a difference between two diagnostic ROC-AUC
values. Concept intervention used the test partition, and lesion-
and stain-level summaries averaged only valid interventions.

\subsection{Ablation protocols}\label{supp:ablation}

Component ablations examined the contributions of evidence-guided
negative construction and knowledge-structured text synthesis by
changing only the semantic-alignment pretraining configuration.
GloSAGE w/o EDGE disabled EDGE while retaining KEYS;
GloSAGE w/o KEYS replaced knowledge-structured text while retaining
EDGE; and GloSAGE w/o EDGE and KEYS removed both components. Each
configuration loaded its own pretrained projection checkpoint and
was evaluated using the same STL downstream interface, with
three training repetitions per lesion-recognition task.

The architectural ablation compared STL with the complete visual
projection head, termed GloSAGE-P. Both configurations
loaded the same GloSAGE projection weights. STL supplied the
output of the first linear layer followed by GELU activation
to the downstream classifier, whereas GloSAGE-P supplied
the complete 512-dimensional projection output. The downstream
preprocessing and optimization settings were otherwise unchanged.

\subsection{Language-model extension protocols}
\label{supp:vqa_ablation}

We explored image-conditioned generation of glomerular morphology
and lesion descriptions by connecting each visual encoder to
Qwen2.5-0.5B-Instruct~\cite{qwen2024qwen25} through a trainable
visual-prefix adapter. The fixed prompt was
``Describe the renal glomerular morphology and identified lesions
in this image. Answer:''.
The corresponding knowledge-graph-derived structured description
served as the reference answer. This fixed-prompt protocol
evaluated descriptive generation rather than unrestricted
question answering.

\paragraph{Visual-prefix adaptation.}
For image $x$, the visual encoder produced a feature vector
$z=f_{\mathrm{img}}(x)$. The adapter applied a linear mapping,
GELU activation, and layer normalization, then reshaped the
output into eight visual-prefix embeddings:
\begin{equation}
    V_{\theta}(x)
    =
    \operatorname{Reshape}_{8\times H}
    \left[
        \operatorname{LayerNorm}
        \left(
            \operatorname{GELU}(Wz+b)
        \right)
    \right],
    \label{eq:vqa_prefix}
\end{equation}
where $H$ is the language-model embedding dimension and
$\theta$ denotes the adapter parameters. The visual embeddings
were prepended to the prompt-token embeddings, allowing the
language model to generate the answer autoregressively.
Both the visual encoder and language model remained frozen;
only the adapter was optimized.

\paragraph{Training and decoding.}
Training minimized autoregressive cross-entropy over answer
tokens:
\begin{equation}
    \mathcal{L}_{\mathrm{ans}}
    =
    -\frac{1}{|\mathcal{A}|}
    \sum_{k\in\mathcal{A}}
    \log p_{\psi}
    \left(
        a_k \mid V_{\theta}(x),q,a_{<k}
    \right),
    \label{eq:vqa_loss}
\end{equation}
where $q$ denotes the prompt, $a_k$ the $k$-th answer token,
$\mathcal{A}$ the set of non-padding answer positions,
and $\psi$ the frozen language-model parameters.
Visual-prefix, prompt, and padding positions were excluded
from the loss.

The predefined manifest training, validation, and test assignments
were preserved without reshuffling. AdamW optimization ran for
10 epochs with a learning rate of $1\times10^{-4}$, weight decay
of 0.01, batch size of eight, and seed 3407. Prompt--answer
sequences were truncated to 192 tokens. The checkpoint with
the lowest validation loss was selected for testing.
Test answers were generated without access to the reference
answer, using deterministic decoding with sampling disabled
and a maximum of 128 new tokens.

\paragraph{Generation metrics.}
Generated text was lowercased and tokenized into word tokens for evaluation. ROUGE-1 and ROUGE-2 used reference-normalized unigram and bigram overlap, respectively, and ROUGE-L used reference-normalized longest common subsequence length. All three metrics were reported as recall-based scores. These measures quantified overlap with reference descriptions rather than independently established clinical correctness.

\paragraph{Exploratory comparison with GPT-6-Astra.}\label{sm_gpt6}
GPT-6-Astra was included as a general-purpose multimodal language-model
comparator for qualitative inspection of selected examples.
For each example, a single glomerular image was submitted in a fresh
conversation together with the fixed question:
``Describe the renal glomerular morphology and identified lesions
in this image.''
No preceding conversational context, additional task-specific
instructions, lesion labels, knowledge-graph content, or reference
answers were supplied.

The same image and descriptive question were used for the
GloSAGE-based language-model extension. Its inference interface
appended the standard ``Answer:'' delimiter used during training.
Responses were inspected alongside the reference lesion information,
with attention to agreement on lesion identity, coverage of
coexisting findings, and unsupported statements. The comparison
was restricted to selected examples and did not involve aggregate
scoring, statistical testing, or a systematic assessment of
equivalence between the models.

\subsection{Quantification and statistical analysis}\label{supp:statistics}

The lesion-related tasks used ROC-AUC, and case-level injury
grading used macro-averaged one-versus-rest ROC-AUC. Accuracy, macro-F1, and
class-frequency-weighted F1 were reported where applicable. Repeated runs
were summarized by their arithmetic mean and sample standard deviation.
For comparisons over the same lesion tasks, models constituted related
samples. An omnibus comparison among more than two models used the Friedman
test. Planned pairwise comparisons used two-sided Wilcoxon signed-rank tests
on task-wise metric differences. Family-wise multiplicity was controlled by
the Holm step-down procedure, and adjusted $p$ values are reported where
specified. The significance threshold was two-sided $p<0.05$. Reported 95\%
confidence intervals for mean task-wise gains were obtained by nonparametric
resampling of the matched task differences and recomputing their mean. No
independence was assumed among images within a task for these task-level
comparisons. Missing or undefined task metrics were excluded pairwise rather
than imputed, and the number of matched tasks was retained with each test.

\subsection{Hardware and software}\label{supp:hardware}

Evaluation experiments were performed on a workstation running
Ubuntu 20.04.6 LTS, equipped with an AMD EPYC 7551P 32-core
processor and four NVIDIA GeForce RTX 3090 GPUs, each with
24~GB of memory. The environment used Python 3.8.20,
PyTorch 2.4.1, torchvision 0.19.1, and CUDA 12.1.
Model implementation and pretrained-model integration used
timm 1.0.9
(\url{https://github.com/huggingface/pytorch-image-models})
and Transformers 4.46.3
(\url{https://github.com/huggingface/transformers}).
Image preprocessing and augmentation used OpenCV 4.8.0
and Albumentations 1.2.0, whereas numerical analysis and
diagnostic metric computation used NumPy 1.24.4 and
scikit-learn 1.3.2. Reference-normalized ROUGE-1, ROUGE-2, and ROUGE-L were implemented in the evaluation code.

Comparator models were implemented using their official
repositories: UNI
(\url{https://github.com/mahmoodlab/UNI}),
CONCH
(\url{https://github.com/mahmoodlab/CONCH}),
PLIP
(\url{https://github.com/PathologyFoundation/plip}),
and KEEP
(\url{https://github.com/MAGIC-AI4Med/KEEP}).
The language-model extension used the released
\texttt{Qwen/Qwen2.5-0.5B-Instruct} checkpoint
(\url{https://huggingface.co/Qwen/Qwen2.5-0.5B-Instruct}).
The case-level implementation followed the public
\texttt{MIL\_BASELINE} framework
(\url{https://github.com/lingxitong/MIL_BASELINE}),
using attention-based multiple-instance learning
\cite{ilse2018attention}.

\clearpage
\section{Supplementary Figures}\label{supp:figures}
\setcounter{figure}{0}
\renewcommand{\thefigure}{S\arabic{figure}}
\makeatletter
\let\theHfigure\thefigure
\makeatother

\begin{figure}[htbp]
    \centering
    \includegraphics[width=\textwidth]{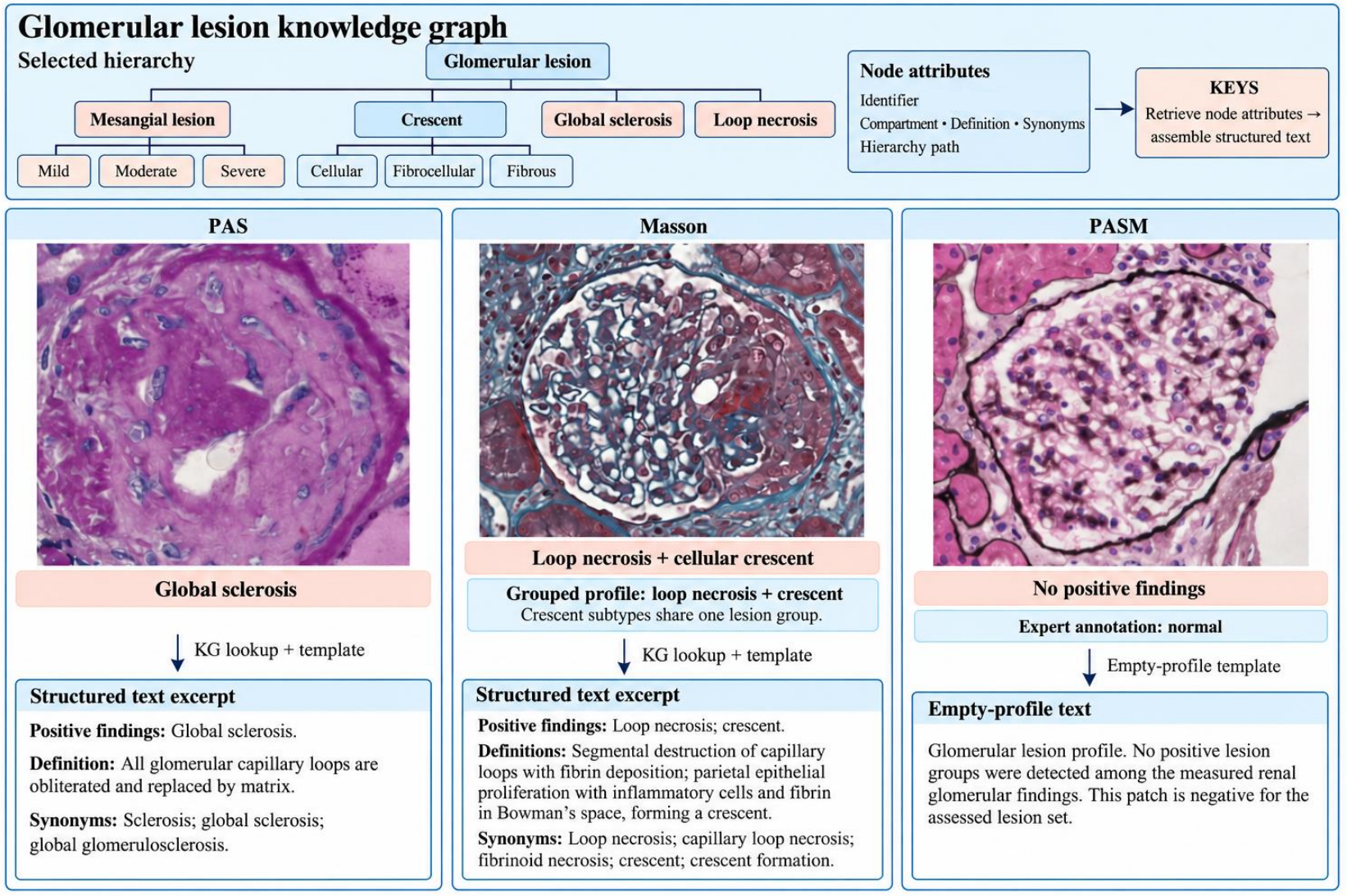}
    \caption{Glomerular lesion knowledge graph and structured-text construction. The selected hierarchy and node attributes illustrate how KEYS retrieves concept names, definitions and synonyms to assemble pathological descriptions. Representative examples show global sclerosis in PAS, loop necrosis with a cellular crescent in Masson, and no positive findings in PASM. Crescent subtypes are consolidated into a single lesion group during profile construction. The empty-profile description is restricted to the assessed lesion set.}
    \label{sm_zeroshot_auc_radar}
\end{figure}

\begin{figure}[htbp]
    \centering
    \includegraphics[width=\textwidth]{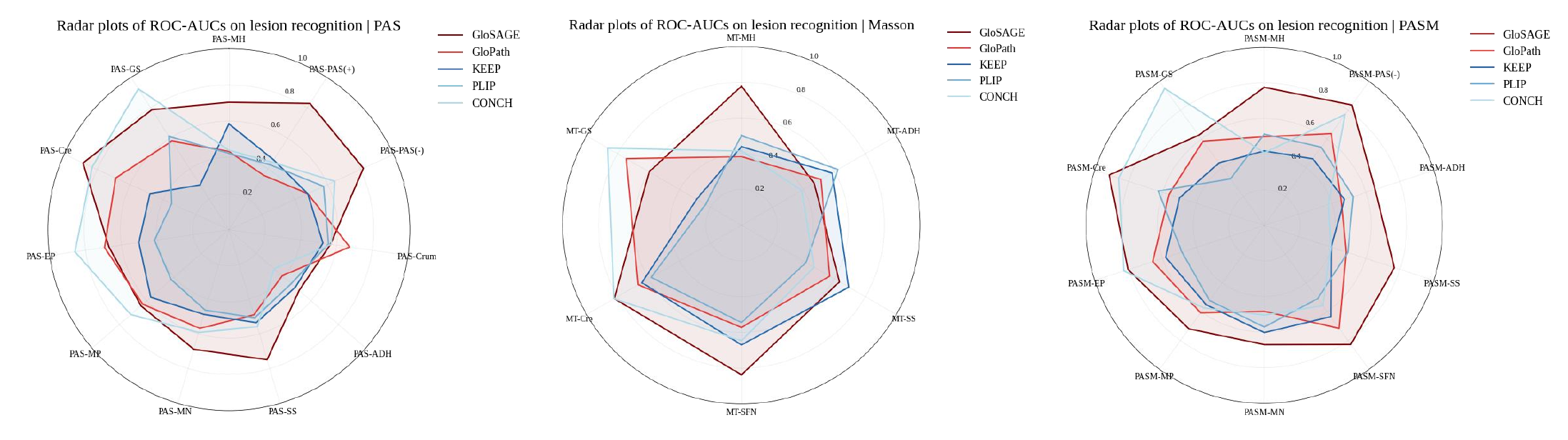}
    \caption{Comparison of models on zero-shot tasks on three stains.}
    \label{sm_zeroshot_auc_radar}
\end{figure}

\begin{figure}[htbp]
    \centering
    \includegraphics[width=\textwidth]{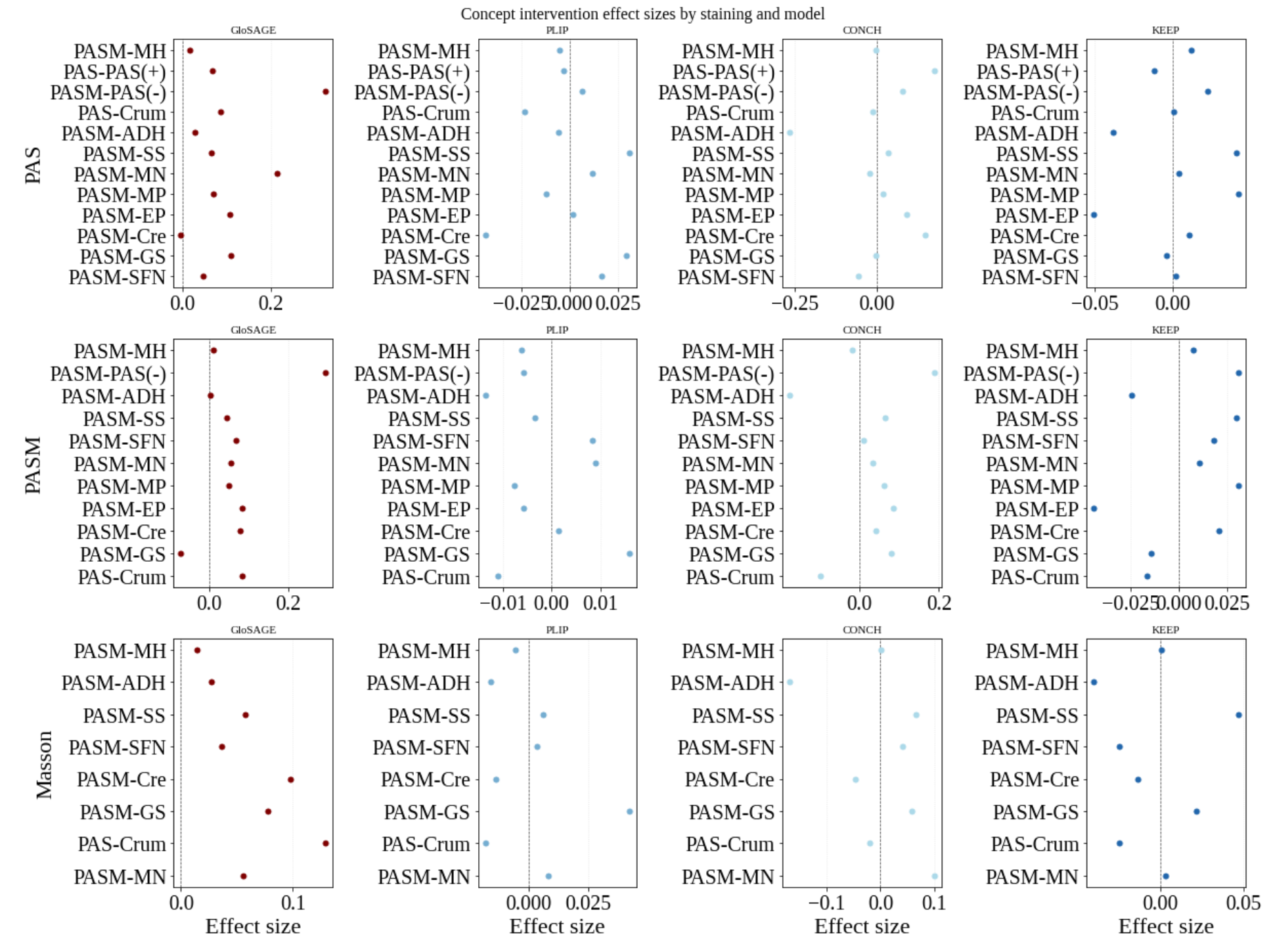}
    \caption{Lesion-wise concept-intervention effect sizes: mean change in image--text similarity on images positive for the original lesion, after replacing its description with a competing concept.}
    \label{sm_concept_all_size}
\end{figure}

\begin{figure}[htbp]
    \centering
    \includegraphics[width=\textwidth]{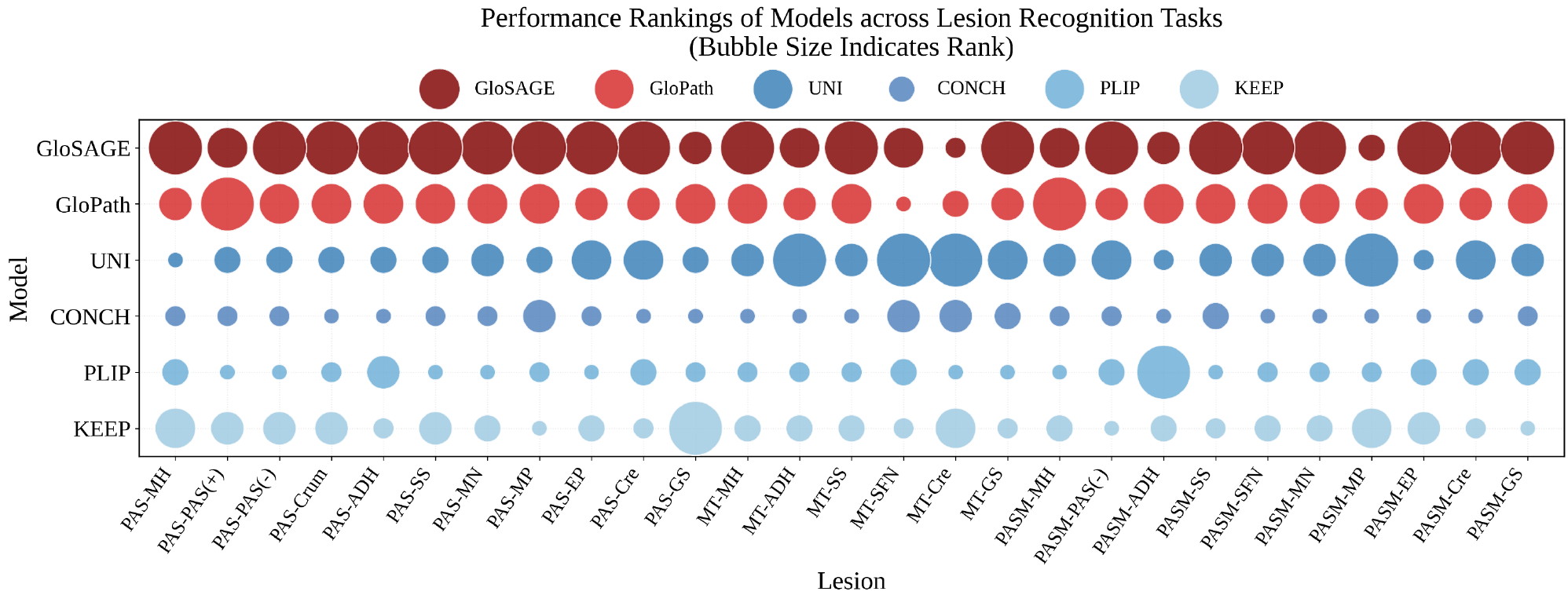}
    \caption{Performance rankings of models across 27 lesion recognition tasks.}
    \label{sm_lesion_rec_bubble}
\end{figure}

\clearpage
\section{Supplementary Tables}\label{supp:tables}
\setcounter{table}{0}
\renewcommand{\thetable}{S\arabic{table}}
\makeatletter
\let\theHtable\thetable
\makeatother

\begin{table}[htbp]
\centering
\caption{ROC-AUC of zero-shot task on lesion recognition (PAS)}
\label{sm_zeroshot_auc_pas}
\begin{tabular}{lccccc}
\toprule
\textbf{Lesion} & \textbf{GloSAGE} & \textbf{GloPath} &
\textbf{KEEP} & \textbf{PLIP} & \textbf{CONCH} \\
\midrule
PAS-MH     & \textbf{0.703} & 0.432 & 0.584 & 0.421 & 0.439 \\
PAS-PAS(+) & \textbf{0.828} & 0.356 & 0.458 & 0.425 & 0.443 \\
PAS-PAS(-) & \textbf{0.816} & 0.479 & 0.478 & 0.575 & 0.643 \\
PAS-Crum   & 0.567 & \textbf{0.674} & 0.525 & 0.556 & 0.571 \\
PAS-ADH    & \textbf{0.513} & 0.389 & 0.486 & 0.455 & 0.330 \\
PAS-SS     & \textbf{0.749} & 0.491 & 0.537 & 0.508 & 0.559 \\
PAS-MN     & \textbf{0.689} & 0.569 & 0.489 & 0.464 & 0.592 \\
PAS-MP     & 0.643 & 0.629 & 0.571 & 0.422 & \textbf{0.717} \\
PAS-EP     & 0.669 & 0.694 & 0.503 & 0.416 & \textbf{0.859} \\
PAS-Cre    & \textbf{0.886} & 0.687 & 0.479 & 0.347 & 0.830 \\
PAS-GS     & 0.787 & 0.582 & 0.294 & 0.614 & \textbf{0.925} \\
\bottomrule
\end{tabular}
\end{table}

\begin{table}[htbp]
\centering
\caption{ROC-AUC of zero-shot task on lesion recognition (Masson)}
\label{sm_zeroshot_auc_masson}
\begin{tabular}{lccccc}
\toprule
\textbf{Lesion} & \textbf{GloSAGE} & \textbf{GloPath} &
\textbf{KEEP} & \textbf{PLIP} & \textbf{CONCH} \\
\midrule
MT-MH  & \textbf{0.778} & 0.384 & 0.440 & 0.502 & 0.416 \\
MT-ADH & 0.470 & 0.513 & 0.585 & \textbf{0.623} & 0.390 \\
MT-SS  & 0.633 & 0.570 & \textbf{0.693} & 0.417 & 0.469 \\
MT-SFN & \textbf{0.837} & 0.572 & 0.670 & 0.545 & 0.647 \\
MT-Cre & 0.821 & 0.667 & 0.643 & 0.583 & \textbf{0.823} \\
MT-GS  & 0.596 & 0.745 & 0.290 & 0.232 & \textbf{0.865} \\
\bottomrule
\end{tabular}
\end{table}

\begin{table}[htbp]
\centering
\caption{ROC-AUC of zero-shot task on lesion recognition (PASM)}
\label{sm_zeroshot_auc_pasm}
\begin{tabular}{lccccc}
\toprule
\textbf{Lesion} & \textbf{GloSAGE} & \textbf{GloPath} &
\textbf{KEEP} & \textbf{PLIP} & \textbf{CONCH} \\
\midrule
PASM-MH     & \textbf{0.776} & 0.498 & 0.418 & 0.513 & 0.408 \\
PASM-PAS(-) & \textbf{0.837} & 0.638 & 0.463 & 0.543 & 0.771 \\
PASM-ADH    & \textbf{0.654} & 0.454 & 0.473 & 0.525 & 0.378 \\
PASM-SS     & \textbf{0.768} & 0.484 & 0.395 & 0.493 & 0.392 \\
PASM-SFN    & \textbf{0.827} & 0.715 & 0.634 & 0.509 & 0.558 \\
PASM-MN     & \textbf{0.670} & 0.483 & 0.603 & 0.571 & 0.504 \\
PASM-MP     & \textbf{0.719} & 0.608 & 0.552 & 0.522 & 0.574 \\
PASM-EP     & 0.803 & 0.658 & 0.582 & 0.486 & \textbf{0.828} \\
PASM-Cre    & \textbf{0.915} & 0.562 & 0.500 & 0.626 & 0.859 \\
PASM-GS     & 0.626 & 0.585 & 0.433 & 0.323 & \textbf{0.952} \\
\bottomrule
\end{tabular}
\end{table}

\begin{table*}[htbp]
\centering
\caption{Retrieval performance on PAS-stained images}
\label{sm_ret_pas}
\begin{tabular}{lcccccc}
\toprule
& \multicolumn{3}{c}{\textbf{Image-to-Text}}
& \multicolumn{3}{c}{\textbf{Text-to-Image}} \\
\cmidrule(lr){2-4}\cmidrule(lr){5-7}
\textbf{Model}
& \textbf{R@1} & \textbf{R@5} & \textbf{R@10}
& \textbf{R@1} & \textbf{R@5} & \textbf{R@10} \\
\midrule
CONCH
& 0.0052 & 0.0411 & 0.1021
& 0.0108 & 0.0215 & 0.0323 \\

GloSAGE
& \textbf{0.1694} & \textbf{0.2723} & \textbf{0.3292}
& \textbf{0.0215} & \textbf{0.0645} & \textbf{0.0753} \\

KEEP
& 0.0109 & 0.0415 & 0.0863
& 0.0108 & 0.0323 & 0.0538 \\

PLIP
& 0.0081 & 0.1210 & 0.2126
& 0.0108 & 0.0430 & \textbf{0.0753} \\
\bottomrule
\end{tabular}
\end{table*}

\begin{table*}[htbp]
\centering
\caption{Retrieval performance on Masson-stained images}
\label{sm_ret_masson}
\begin{tabular}{lcccccc}
\toprule
& \multicolumn{3}{c}{\textbf{Image-to-Text}}
& \multicolumn{3}{c}{\textbf{Text-to-Image}} \\
\cmidrule(lr){2-4}\cmidrule(lr){5-7}
\textbf{Model}
& \textbf{R@1} & \textbf{R@5} & \textbf{R@10}
& \textbf{R@1} & \textbf{R@5} & \textbf{R@10} \\
\midrule
CONCH
& 0.0033 & 0.0658 & 0.2242
& 0.0500 & 0.0500 & 0.2000 \\

GloSAGE
& \textbf{0.2867} & 0.4826 & 0.6007
& \textbf{0.1000} & \textbf{0.1500} & \textbf{0.3000} \\

KEEP
& 0.0076 & 0.0397 & 0.1094
& 0.0500 & \textbf{0.1500} & 0.2000 \\

PLIP
& 0.2671 & \textbf{0.7193} & \textbf{0.8069}
& \textbf{0.1000} & \textbf{0.1500} & 0.2000 \\
\bottomrule
\end{tabular}
\end{table*}

\begin{table*}[htbp]
\centering
\caption{Retrieval performance on PASM-stained images}
\label{sm_ret_pasm}
\begin{tabular}{lcccccc}
\toprule
& \multicolumn{3}{c}{\textbf{Image-to-Text}}
& \multicolumn{3}{c}{\textbf{Text-to-Image}} \\
\cmidrule(lr){2-4}\cmidrule(lr){5-7}
\textbf{Model}
& \textbf{R@1} & \textbf{R@5} & \textbf{R@10}
& \textbf{R@1} & \textbf{R@5} & \textbf{R@10} \\
\midrule
CONCH
& 0.0062 & 0.0355 & 0.0719
& 0.0392 & \textbf{0.1176} & 0.1373 \\

GloSAGE
& \textbf{0.1083} & 0.2233 & 0.2655
& \textbf{0.0588} & 0.0784 & \textbf{0.1569} \\

KEEP
& 0.0110 & 0.0652 & 0.1112
& 0.0000 & 0.0392 & 0.0784 \\

PLIP
& 0.0585 & \textbf{0.2530} & \textbf{0.4528}
& 0.0196 & 0.0588 & 0.0784 \\
\bottomrule
\end{tabular}
\end{table*}

\begin{table}[htbp]
\centering
\caption{Mean hard-negative description-discrimination accuracy by staining. Each lesion-positive image is matched to one correct and five closely related incorrect descriptions; chance top-1 accuracy is 0.1667. Means assign equal weight to each evaluated lesion (12 PAS, 8 Masson and 11 PASM), including additional stress-test lesions beyond the 27 canonical zero-shot recognition tasks. GloPath is omitted here to match the models displayed in Fig.~\ref{fig:semantic}C.}
\label{sm_hard_negative}
\begin{tabular}{lccc}
\toprule
\textbf{Model} & \textbf{PAS} & \textbf{Masson} & \textbf{PASM} \\
\midrule
GloSAGE & 0.283 & 0.201 & 0.179 \\
PLIP & 0.106 & 0.139 & 0.116 \\
CONCH & 0.176 & 0.083 & 0.179 \\
KEEP & 0.226 & 0.120 & 0.188 \\
\bottomrule
\end{tabular}
\end{table}

\begin{table}[htbp]
\centering
\caption{Lesion-wise hard-negative top-1 accuracy for PAS. The final row is the unweighted mean across all 12 evaluated lesion targets; PAS-SFN is an additional stress-test target beyond the standard 11-lesion recognition panel.}
\label{sm_hard_negative_pas}
\begin{tabular}{lcccc}
\toprule
\textbf{Lesion} & \textbf{GloSAGE} & \textbf{PLIP} & \textbf{CONCH} & \textbf{KEEP} \\
\midrule
PAS-MH & 0.002 & 0.008 & 0.003 & 0.021 \\
PAS-PAS($+$) & 0.244 & 0.000 & 0.962 & 0.000 \\
PAS-PAS($-$) & 0.827 & 0.013 & 0.000 & 0.680 \\
PAS-Crum & 0.161 & 0.000 & 0.000 & 0.032 \\
PAS-ADH & 0.014 & 0.000 & 0.000 & 0.000 \\
PAS-SS & 0.092 & 0.618 & 0.066 & 0.895 \\
PAS-MN & 0.782 & 0.059 & 0.000 & 0.000 \\
PAS-MP & 0.053 & 0.000 & 0.011 & 0.968 \\
PAS-EP & 0.406 & 0.000 & 0.319 & 0.000 \\
PAS-Cre & 0.000 & 0.000 & 0.691 & 0.091 \\
PAS-GS & 0.635 & 0.570 & 0.023 & 0.008 \\
PAS-SFN & 0.179 & 0.000 & 0.036 & 0.018 \\
\midrule
Mean & 0.283 & 0.106 & 0.176 & 0.226 \\
\bottomrule
\end{tabular}
\end{table}

\begin{table}[htbp]
\centering
\caption{Lesion-wise hard-negative top-1 accuracy for Masson. The final row is the unweighted mean across all eight evaluated lesion targets; MT-Crum and MT-MN are additional stress-test targets beyond the standard six-lesion recognition panel.}
\label{sm_hard_negative_masson}
\begin{tabular}{lcccc}
\toprule
\textbf{Lesion} & \textbf{GloSAGE} & \textbf{PLIP} & \textbf{CONCH} & \textbf{KEEP} \\
\midrule
MT-MH & 0.333 & 0.004 & 0.046 & 0.007 \\
MT-ADH & 0.051 & 0.000 & 0.000 & 0.000 \\
MT-SS & 0.233 & 0.000 & 0.067 & 0.900 \\
MT-SFN & 0.030 & 0.104 & 0.060 & 0.000 \\
MT-Cre & 0.184 & 0.000 & 0.000 & 0.000 \\
MT-GS & 0.375 & 1.000 & 0.492 & 0.055 \\
MT-Crum & 0.000 & 0.000 & 0.000 & 0.000 \\
MT-MN & 0.400 & 0.000 & 0.000 & 0.000 \\
\midrule
Mean & 0.201 & 0.139 & 0.083 & 0.120 \\
\bottomrule
\end{tabular}
\end{table}

\begin{table}[htbp]
\centering
\caption{Lesion-wise hard-negative top-1 accuracy for PASM. The final row is the unweighted mean across all 11 evaluated lesion targets; PASM-Crum is an additional stress-test target beyond the standard ten-lesion recognition panel.}
\label{sm_hard_negative_pasm}
\begin{tabular}{lcccc}
\toprule
\textbf{Lesion} & \textbf{GloSAGE} & \textbf{PLIP} & \textbf{CONCH} & \textbf{KEEP} \\
\midrule
PASM-MH & 0.030 & 0.000 & 0.000 & 0.017 \\
PASM-PAS($-$) & 0.384 & 0.000 & 0.233 & 0.178 \\
PASM-ADH & 0.000 & 0.000 & 0.000 & 0.000 \\
PASM-SS & 0.029 & 0.000 & 0.088 & 0.794 \\
PASM-SFN & 0.263 & 0.289 & 0.237 & 0.184 \\
PASM-MN & 0.343 & 0.051 & 0.000 & 0.056 \\
PASM-MP & 0.205 & 0.000 & 0.455 & 0.598 \\
PASM-EP & 0.406 & 0.000 & 0.359 & 0.000 \\
PASM-Cre & 0.026 & 0.105 & 0.158 & 0.237 \\
PASM-GS & 0.238 & 0.829 & 0.439 & 0.000 \\
PASM-Crum & 0.040 & 0.000 & 0.000 & 0.000 \\
\midrule
Mean & 0.179 & 0.116 & 0.179 & 0.188 \\
\bottomrule
\end{tabular}
\end{table}

\begin{table*}[htbp]
\centering
\caption{Intervention-discrimination ROC-AUC under PAS staining. The score is the change in image--text similarity between the original and intervened descriptions; ROC-AUC evaluates its separation of lesion-positive versus lesion-negative images. It is not a difference between two diagnostic ROC-AUC values. Each lesion entry averages its valid interventions; Overall averages valid interventions across lesions.}
\label{sm_concept_pas_delta_auc}
\resizebox{\textwidth}{!}{
\begin{tabular}{lcccccccccccc}
\toprule
Model
& MH & PAS($+$) & PAS($-$) & Crum & ADH & SS
& MN & MP & EP & Cre & GS & Overall \\
\midrule
GloSAGE
& 0.523
& \textbf{0.664}
& \textbf{0.884}
& 0.551
& 0.441
& \textbf{0.696}
& \textbf{0.721}
& 0.660
& 0.564
& 0.662
& \textbf{0.710}
& \textbf{0.651} \\

PLIP
& 0.502
& 0.482
& 0.361
& 0.515
& \textbf{0.501}
& 0.607
& 0.513
& 0.537
& 0.409
& 0.428
& 0.577
& 0.492 \\

CONCH
& 0.455
& 0.495
& 0.529
& 0.477
& 0.360
& 0.343
& 0.560
& 0.581
& \textbf{0.645}
& \textbf{0.808}
& 0.334
& 0.509 \\

KEEP
& \textbf{0.589}
& 0.383
& 0.517
& \textbf{0.583}
& 0.396
& 0.512
& 0.554
& \textbf{0.682}
& 0.310
& 0.604
& 0.708
& 0.529 \\
\bottomrule
\end{tabular}
}
\end{table*}

\begin{sidewaystable*}[p]
\centering
\caption{Glomerular lesion categories, morphological criteria,
and grading categories in GloAtlas.}
\label{sm_lesion_definitions}
\small
\setlength{\tabcolsep}{4pt}
\renewcommand{\arraystretch}{1.25}

\begin{tabular}{|p{0.8cm}|p{3cm}|p{1.3cm}|p{9cm}|p{2.5cm}|p{1.8cm}|}
\hline
\textbf{No.}
& \textbf{Lesion}
& \textbf{Abbrev.}
& \textbf{Morphological criteria}
& \textbf{Grading categories}
& \textbf{Stains} \\
\hline

1--3
& Mesangial hypercellularity
& MH
& Increased mesangial cellularity, assessed away from the vascular
pole. Mild, moderate, and severe involvement correspond to 4--5,
6--7, and $\geq 8$ mesangial cells per mesangial area,
respectively.
& Absent; mild; moderate; severe
& PAS, Masson, PASM \\
\hline

5
& PAS-positive glomerular finding
& PAS(+)
& PAS-positive glomerular lesions show magenta-staining matrix expansion or hyaline material.
& Not applicable
& PAS \\
\hline

6
& PAS-negative glomerular finding
& PAS($-$)
& PAS-negative lesions show pale areas of tissue injury or abnormal material accumulation.
& Not applicable
& PAS, PASM \\
\hline

7
& Ischemic-type glomerular capillary collapse
& Crum
& Wrinkled and folded glomerular basement membranes with narrowing
or closure of capillary lumina, without overlying podocyte
hyperplasia.
& Not applicable
& PAS \\
\hline

8
& Glomerular tuft adhesion to Bowman's capsule
& ADH
& Localized continuity between the glomerular tuft and Bowman's
capsule, distinguished from an associated crescent or sclerotic
segment.
& Not applicable
& PAS, Masson, PASM \\
\hline

9
& Segmental glomerulosclerosis
& SS
& Matrix-associated obliteration of capillary lumina involving
part of the tuft, with preservation of a nonsclerotic portion.
& Not applicable
& PAS, Masson, PASM \\
\hline

10
& Segmental fibrinoid necrosis
& SFN
& Segmental tuft destruction with fibrin deposition and
basement-membrane disruption, potentially accompanied by
nuclear debris.
& Not applicable
& Masson, PASM \\
\hline

11
& Membranous-pattern glomerular lesion
& MN
& Capillary-wall thickening with basement-membrane spikes or
intramembranous holes, particularly evident on silver staining;
these features may be inconspicuous in early lesions.
& Not applicable
& PAS, PASM \\
\hline

12--13
& Membranoproliferative-pattern lesion
& MP
& Proliferative changes accompanied by capillary-wall remodeling,
including basement-membrane duplication and mesangial interposition.
Segmental and global categories describe the extent of tuft
involvement.
& Segmental; global
& PAS, PASM \\
\hline

14--15
& Endocapillary hypercellularity
& EP
& Increased intracapillary cells causing narrowing or occlusion
of capillary lumina. Segmental and global categories describe
the extent of tuft involvement.
& Segmental; global
& PAS, PASM \\
\hline

16--18
& Glomerular crescent
& Cre
& An extracapillary lesion in Bowman's space. Cellular crescents
contain $>75\%$ cells and fibrin; fibrocellular crescents contain
25--75\% cells and fibrin; fibrous crescents contain $>75\%$
fibrous matrix.
& Cellular; fibrocellular; fibrous
& PAS, Masson, PASM \\
\hline

19
& Global glomerulosclerosis
& GS
& Sclerotic matrix replacement and obliteration of capillary
lumina involving the entire glomerular tuft.
& Not applicable
& PAS, Masson, PASM \\
\hline

\end{tabular}
\end{sidewaystable*}

\begin{table}[htbp]
\centering
\caption{Intervention-discrimination ROC-AUC under Masson staining. See Supplementary Table~\ref{sm_concept_pas_delta_auc} for the score definition and aggregation.}
\label{sm_concept_masson_delta_auc}
\begin{tabular}{lccccccc}
\toprule
Model
& MH & ADH & SS & SFN & Cre & GS & Overall \\
\midrule
GloSAGE
& 0.518
& 0.488
& \textbf{0.654}
& 0.529
& \textbf{0.683}
& 0.621
& \textbf{0.584} \\

PLIP
& \textbf{0.551}
& 0.411
& 0.584
& 0.345
& 0.561
& \textbf{0.741}
& 0.507 \\

CONCH
& 0.404
& 0.421
& 0.332
& \textbf{0.604}
& 0.637
& 0.336
& 0.481 \\

KEEP
& 0.454
& \textbf{0.514}
& 0.480
& 0.570
& 0.502
& 0.578
& 0.475 \\
\bottomrule
\end{tabular}
\end{table}

\begin{table*}[htbp]
\centering
\caption{Intervention-discrimination ROC-AUC under PASM staining. See Supplementary Table~\ref{sm_concept_pas_delta_auc} for the score definition and aggregation.}
\label{sm_concept_pasm_delta_auc}
\resizebox{\textwidth}{!}{
\begin{tabular}{lccccccccccc}
\toprule
Model
& MH & PAS($-$) & ADH & SS & SFN
& MN & MP & EP & Cre & GS & Overall \\
\midrule
GloSAGE
& 0.527
& \textbf{0.840}
& 0.437
& \textbf{0.665}
& 0.636
& 0.595
& 0.545
& 0.616
& 0.721
& 0.549
& \textbf{0.613} \\

PLIP
& \textbf{0.562}
& 0.436
& 0.432
& 0.381
& 0.414
& 0.505
& 0.496
& 0.543
& 0.548
& 0.527
& 0.485 \\

CONCH
& 0.417
& 0.653
& 0.428
& 0.334
& 0.506
& \textbf{0.617}
& 0.497
& \textbf{0.732}
& \textbf{0.773}
& 0.456
& 0.532 \\

KEEP
& 0.451
& 0.620
& \textbf{0.556}
& 0.480
& \textbf{0.674}
& 0.608
& \textbf{0.580}
& 0.375
& 0.725
& \textbf{0.559}
& 0.559 \\
\bottomrule
\end{tabular}
}
\end{table*}

\begin{table*}[htbp]
\centering
\caption{Performance of lesion recognition across all the models (PAS)}
\label{sm_lesion_rec_pas}
\resizebox{\textwidth}{!}{
\begin{tabular}{lccccccccccc}
\toprule
Model
& MH
& PAS(+)
& PAS($-$)
& Crum
& ADH
& SS
& MN
& MP
& EP
& Cre
& GS \\
\midrule
GloSAGE
& \textbf{0.930 $\pm$ 0.005}
& 0.954 $\pm$ 0.002
& \textbf{1.000 $\pm$ 0.000}
& \textbf{0.922 $\pm$ 0.024}
& \textbf{0.815 $\pm$ 0.014}
& \textbf{0.967 $\pm$ 0.004}
& \textbf{0.957 $\pm$ 0.001}
& \textbf{0.968 $\pm$ 0.003}
& \textbf{0.975 $\pm$ 0.002}
& \textbf{0.989 $\pm$ 0.003}
& 0.997 $\pm$ 0.001 \\

GloPath
& 0.909 $\pm$ 0.038
& \textbf{0.955 $\pm$ 0.003}
& 1.000 $\pm$ 0.000
& 0.887 $\pm$ 0.037
& 0.779 $\pm$ 0.033
& 0.962 $\pm$ 0.006
& 0.949 $\pm$ 0.013
& 0.965 $\pm$ 0.005
& 0.968 $\pm$ 0.015
& 0.980 $\pm$ 0.005
& 0.997 $\pm$ 0.001 \\

UNI
& 0.879 $\pm$ 0.092
& 0.936 $\pm$ 0.008
& 1.000 $\pm$ 0.000
& 0.842 $\pm$ 0.108
& 0.762 $\pm$ 0.016
& 0.924 $\pm$ 0.019
& 0.947 $\pm$ 0.006
& 0.952 $\pm$ 0.008
& 0.969 $\pm$ 0.004
& 0.984 $\pm$ 0.005
& 0.996 $\pm$ 0.001 \\

CONCH
& 0.884 $\pm$ 0.032
& 0.934 $\pm$ 0.004
& 0.997 $\pm$ 0.001
& 0.777 $\pm$ 0.157
& 0.720 $\pm$ 0.027
& 0.919 $\pm$ 0.011
& 0.923 $\pm$ 0.006
& 0.954 $\pm$ 0.006
& 0.965 $\pm$ 0.002
& 0.963 $\pm$ 0.008
& 0.993 $\pm$ 0.001 \\

PLIP
& 0.889 $\pm$ 0.005
& 0.924 $\pm$ 0.004
& 0.996 $\pm$ 0.002
& 0.842 $\pm$ 0.048
& 0.765 $\pm$ 0.016
& 0.915 $\pm$ 0.024
& 0.868 $\pm$ 0.012
& 0.950 $\pm$ 0.003
& 0.943 $\pm$ 0.004
& 0.967 $\pm$ 0.007
& 0.994 $\pm$ 0.002 \\

KEEP
& 0.926 $\pm$ 0.002
& 0.936 $\pm$ 0.005
& 1.000 $\pm$ 0.000
& 0.861 $\pm$ 0.020
& 0.760 $\pm$ 0.015
& 0.951 $\pm$ 0.012
& 0.923 $\pm$ 0.011
& 0.946 $\pm$ 0.004
& 0.966 $\pm$ 0.011
& 0.966 $\pm$ 0.005
& \textbf{0.998 $\pm$ 0.000} \\
\bottomrule
\end{tabular}
}
\end{table*}

\begin{table*}[htbp]
\centering
\caption{Performance of lesion recognition across all the models (Masson)}
\label{sm_lesion_rec_masson}
\resizebox{\textwidth}{!}{
\begin{tabular}{lcccccc}
\toprule
Model
& MH
& ADH
& SS
& SFN
& Cre
& GS \\
\midrule
GloSAGE
& \textbf{0.962 $\pm$ 0.005}
& 0.763 $\pm$ 0.024
& \textbf{0.948 $\pm$ 0.013}
& 0.986 $\pm$ 0.004
& 0.963 $\pm$ 0.002
& \textbf{0.997 $\pm$ 0.001} \\

GloPath
& 0.961 $\pm$ 0.003
& 0.750 $\pm$ 0.050
& 0.910 $\pm$ 0.054
& 0.982 $\pm$ 0.007
& 0.974 $\pm$ 0.018
& 0.995 $\pm$ 0.003 \\

UNI
& 0.952 $\pm$ 0.005
& \textbf{0.773 $\pm$ 0.080}
& 0.868 $\pm$ 0.060
& \textbf{0.990 $\pm$ 0.002}
& \textbf{0.986 $\pm$ 0.005}
& 0.995 $\pm$ 0.000 \\

CONCH
& 0.928 $\pm$ 0.007
& 0.704 $\pm$ 0.038
& 0.709 $\pm$ 0.076
& 0.985 $\pm$ 0.001
& 0.983 $\pm$ 0.008
& 0.993 $\pm$ 0.001 \\

PLIP
& 0.929 $\pm$ 0.003
& 0.708 $\pm$ 0.099
& 0.710 $\pm$ 0.042
& 0.984 $\pm$ 0.006
& 0.960 $\pm$ 0.018
& 0.992 $\pm$ 0.002 \\

KEEP
& 0.930 $\pm$ 0.005
& 0.723 $\pm$ 0.008
& 0.786 $\pm$ 0.077
& 0.984 $\pm$ 0.006
& 0.984 $\pm$ 0.004
& 0.992 $\pm$ 0.002 \\
\bottomrule
\end{tabular}
}
\end{table*}

\begin{table*}[htbp]
\centering
\caption{Performance of lesion recognition across all the models (PASM)}
\label{sm_lesion_rec_pasm}
\resizebox{\textwidth}{!}{
\begin{tabular}{lcccccccccc}
\toprule
Model
& MH
& PAS($-$)
& ADH
& SS
& SFN
& MN
& MP
& EP
& Cre
& GS \\
\midrule
GloSAGE
& 0.928 $\pm$ 0.001
& \textbf{0.998 $\pm$ 0.001}
& 0.762 $\pm$ 0.030
& \textbf{0.961 $\pm$ 0.010}
& \textbf{0.986 $\pm$ 0.002}
& \textbf{0.946 $\pm$ 0.012}
& 0.953 $\pm$ 0.001
& \textbf{0.975 $\pm$ 0.001}
& \textbf{0.988 $\pm$ 0.000}
& \textbf{0.996 $\pm$ 0.002} \\

GloPath
& \textbf{0.928 $\pm$ 0.004}
& 0.997 $\pm$ 0.001
& 0.768 $\pm$ 0.038
& 0.945 $\pm$ 0.018
& 0.982 $\pm$ 0.004
& 0.944 $\pm$ 0.008
& 0.955 $\pm$ 0.008
& 0.972 $\pm$ 0.009
& 0.982 $\pm$ 0.010
& 0.995 $\pm$ 0.001 \\

UNI
& 0.924 $\pm$ 0.003
& 0.998 $\pm$ 0.001
& 0.747 $\pm$ 0.022
& 0.943 $\pm$ 0.031
& 0.977 $\pm$ 0.011
& 0.929 $\pm$ 0.013
& \textbf{0.969 $\pm$ 0.001}
& 0.971 $\pm$ 0.006
& 0.983 $\pm$ 0.009
& 0.992 $\pm$ 0.001 \\

CONCH
& 0.912 $\pm$ 0.005
& 0.993 $\pm$ 0.004
& 0.695 $\pm$ 0.013
& 0.933 $\pm$ 0.018
& 0.945 $\pm$ 0.014
& 0.894 $\pm$ 0.014
& 0.943 $\pm$ 0.007
& 0.953 $\pm$ 0.006
& 0.938 $\pm$ 0.004
& 0.990 $\pm$ 0.002 \\

PLIP
& 0.905 $\pm$ 0.005
& 0.993 $\pm$ 0.002
& \textbf{0.790 $\pm$ 0.049}
& 0.902 $\pm$ 0.034
& 0.946 $\pm$ 0.005
& 0.920 $\pm$ 0.009
& 0.953 $\pm$ 0.005
& 0.971 $\pm$ 0.002
& 0.968 $\pm$ 0.019
& 0.991 $\pm$ 0.002 \\

KEEP
& 0.912 $\pm$ 0.006
& 0.984 $\pm$ 0.021
& 0.749 $\pm$ 0.057
& 0.914 $\pm$ 0.012
& 0.955 $\pm$ 0.015
& 0.927 $\pm$ 0.002
& 0.957 $\pm$ 0.004
& 0.972 $\pm$ 0.007
& 0.944 $\pm$ 0.041
& 0.989 $\pm$ 0.001 \\
\bottomrule
\end{tabular}
}
\end{table*}

\begin{table}[htbp]
\centering
\caption{Relative reduction in ROC-AUC deficit of GloSAGE compared with baseline models on PAS staining.}
\label{tab:relative_auc_deficit_pas}
\small
\setlength{\tabcolsep}{7pt}
\renewcommand{\arraystretch}{1.15}
\begin{tabular}{lccccc}
\toprule
\textbf{Lesion} & \textbf{GloPath} & \textbf{UNI} &
\textbf{CONCH} & \textbf{PLIP} & \textbf{KEEP} \\
\midrule
MH      & +22.38 & +41.96 & +39.41 & +36.34 & +5.18 \\
PAS(+)  & -1.70 & +28.35 & +30.52 & +39.78 & +27.62 \\
PAS(-)  & +100.00 & +100.00 & +100.00 & +100.00 & +100.00 \\
Crum    & +30.81 & +50.71 & +64.99 & +50.81 & +43.83 \\
ADH     & +15.98 & +22.09 & +33.75 & +21.06 & +22.69 \\
SS      & +13.59 & +56.85 & +59.77 & +61.70 & +32.94 \\
MN      & +15.88 & +19.68 & +44.76 & +67.59 & +44.46 \\
MP      & +9.62 & +33.87 & +32.16 & +36.66 & +41.63 \\
EP      & +19.73 & +17.91 & +27.73 & +55.34 & +25.98 \\
Cre     & +45.42 & +35.16 & +71.47 & +68.07 & +68.14 \\
GS      & -4.66 & +11.87 & +49.26 & +43.24 & -48.12 \\
\midrule
\textbf{Overall} &
\textbf{+18.72} & \textbf{+35.03} &
\textbf{+45.87} & \textbf{+44.46} & \textbf{+31.34} \\
\bottomrule
\end{tabular}
\end{table}

\begin{table}[htbp]
\centering
\caption{Relative reduction in ROC-AUC deficit of GloSAGE compared with baseline models on Masson staining.}
\label{tab:relative_auc_deficit_masson}
\small
\setlength{\tabcolsep}{7pt}
\renewcommand{\arraystretch}{1.15}
\begin{tabular}{lccccc}
\toprule
\textbf{Lesion} & \textbf{GloPath} & \textbf{UNI} &
\textbf{CONCH} & \textbf{PLIP} & \textbf{KEEP} \\
\midrule
MH      & +2.41 & +21.48 & +47.76 & +46.80 & +46.20 \\
ADH     & +5.19 & -4.37 & +20.03 & +18.75 & +14.28 \\
SS      & +41.96 & +60.13 & +81.95 & +81.91 & +75.44 \\
SFN     & +23.17 & -34.63 & +7.61 & +12.74 & +13.30 \\
Cre     & -42.09 & -160.00 & -113.14 & +7.66 & -123.56 \\
GS      & +48.72 & +45.71 & +59.87 & +66.23 & +63.45 \\
\midrule
\textbf{Overall} &
\textbf{+11.17} & \textbf{+12.76} &
\textbf{+45.50} & \textbf{+46.88} & \textbf{+36.60} \\
\bottomrule
\end{tabular}
\end{table}

\begin{table}[htbp]
\centering
\caption{Relative reduction in ROC-AUC deficit of GloSAGE compared with baseline models on PASM staining.}
\label{tab:relative_auc_deficit_pasm}
\small
\setlength{\tabcolsep}{7pt}
\renewcommand{\arraystretch}{1.15}
\begin{tabular}{lccccc}
\toprule
\textbf{Lesion} & \textbf{GloPath} & \textbf{UNI} &
\textbf{CONCH} & \textbf{PLIP} & \textbf{KEEP} \\
\midrule
MH      & -0.40 & +5.79 & +18.59 & +24.33 & +18.39 \\
PAS(-)  & +32.15 & +29.44 & +75.50 & +74.80 & +88.92 \\
ADH     & -2.35 & +6.25 & +22.00 & -12.90 & +5.47 \\
SS      & +28.37 & +31.15 & +41.98 & +60.01 & +54.35 \\
SFN     & +20.90 & +37.48 & +74.11 & +73.49 & +68.23 \\
MN      & +3.17 & +23.46 & +48.47 & +31.72 & +25.54 \\
MP      & -2.66 & -47.63 & +18.98 & +0.20 & -7.09 \\
EP      & +8.53 & +13.52 & +46.09 & +11.71 & +9.80 \\
Cre     & +34.16 & +29.93 & +80.80 & +62.08 & +78.51 \\
GS      & +26.93 & +48.75 & +60.14 & +57.11 & +62.48 \\
\midrule
\textbf{Overall} &
\textbf{+4.70} & \textbf{+10.86} &
\textbf{+37.04} & \textbf{+23.21} & \textbf{+27.28} \\
\bottomrule
\end{tabular}
\end{table}


\begin{table*}[!t]
\centering
\caption{Performance comparison of models on lesion grading tasks. Values are ROC-AUC (mean $\pm$ standard deviation).}
\label{sm_lesion_grading}
\resizebox{\textwidth}{!}{
\begin{tabular}{lcccccc}
\hline
Task & GloSAGE & GloPath & UNI & PLIP & CONCH & KEEP \\
\hline
PAS-MH-Grading
& \textbf{0.897 $\pm$ 0.005}
& 0.896 $\pm$ 0.004
& 0.894 $\pm$ 0.004
& 0.873 $\pm$ 0.005
& 0.861 $\pm$ 0.004
& 0.890 $\pm$ 0.009 \\

MT-MH-Grading
& \textbf{0.920 $\pm$ 0.009}
& 0.907 $\pm$ 0.011
& 0.901 $\pm$ 0.007
& 0.883 $\pm$ 0.006
& 0.852 $\pm$ 0.012
& 0.856 $\pm$ 0.015 \\

PASM-MH-Grading
& 0.892 $\pm$ 0.005
& \textbf{0.893 $\pm$ 0.006}
& 0.877 $\pm$ 0.005
& 0.885 $\pm$ 0.011
& 0.834 $\pm$ 0.014
& 0.877 $\pm$ 0.012 \\

PAS-Cre-Grading
& 0.976 $\pm$ 0.016
& \textbf{0.982 $\pm$ 0.008}
& 0.946 $\pm$ 0.011
& 0.949 $\pm$ 0.031
& 0.932 $\pm$ 0.028
& 0.964 $\pm$ 0.017 \\

MT-Cre-Grading
& \textbf{0.961 $\pm$ 0.015}
& 0.908 $\pm$ 0.046
& 0.915 $\pm$ 0.032
& 0.840 $\pm$ 0.028
& 0.881 $\pm$ 0.074
& 0.922 $\pm$ 0.043 \\

PASM-Cre-Grading
& 0.925 $\pm$ 0.016
& \textbf{0.937 $\pm$ 0.016}
& 0.936 $\pm$ 0.029
& 0.915 $\pm$ 0.005
& 0.892 $\pm$ 0.036
& 0.892 $\pm$ 0.054 \\

PAS-MP-Grading
& \textbf{0.925 $\pm$ 0.005}
& 0.903 $\pm$ 0.033
& 0.910 $\pm$ 0.031
& 0.847 $\pm$ 0.024
& 0.878 $\pm$ 0.039
& 0.881 $\pm$ 0.062 \\

PASM-MP-Grading
& 0.857 $\pm$ 0.009
& 0.836 $\pm$ 0.015
& \textbf{0.866 $\pm$ 0.034}
& 0.841 $\pm$ 0.010
& 0.836 $\pm$ 0.030
& 0.842 $\pm$ 0.020 \\

PAS-EP-Grading
& \textbf{0.887 $\pm$ 0.102}
& 0.813 $\pm$ 0.118
& 0.830 $\pm$ 0.121
& 0.834 $\pm$ 0.127
& 0.835 $\pm$ 0.110
& 0.810 $\pm$ 0.112 \\

PASM-EP-Grading
& \textbf{0.876 $\pm$ 0.014}
& 0.844 $\pm$ 0.098
& 0.788 $\pm$ 0.045
& 0.845 $\pm$ 0.100
& 0.808 $\pm$ 0.100
& 0.827 $\pm$ 0.070 \\
\hline
\end{tabular}
}
\end{table*}

\begin{table*}[htbp]
\centering
\caption{Performance on 11 PAS-related lesion recognition tasks on GloPro.}
\label{sm_glopro_recognition}
\resizebox{\textwidth}{!}{
\begin{tabular}{lccccccccccc}
\toprule
Model & MH & PAS(+) & PAS(-) & Crum & ADH & SS & MN & MP & EP & Cre & GS \\
\midrule
GloSAGE
& \textbf{0.9196} & 0.8462 & \textbf{0.9900}
& \textbf{0.7863} & \textbf{0.8127} & 0.9142
& 0.8807 & \textbf{0.9919} & \textbf{0.9995}
& \textbf{0.9680} & 0.9974 \\

GloPath
& 0.8969 & \textbf{0.8565} & 0.9830
& 0.7467 & 0.7904 & \textbf{0.9205}
& \textbf{0.9194} & 0.9817 & 0.9987
& 0.9542 & 0.9966 \\

UNI
& 0.8869 & 0.8226 & 0.9858
& 0.7458 & 0.7706 & 0.8660
& 0.8914 & 0.9620 & 0.9937
& 0.9462 & \textbf{0.9976} \\

PLIP
& 0.8591 & 0.8126 & 0.9645
& 0.7649 & 0.7739 & 0.8722
& 0.8791 & 0.9052 & 0.9956
& 0.9363 & 0.9963 \\

CONCH
& 0.8373 & 0.7979 & 0.9570
& 0.6999 & 0.7478 & 0.8177
& 0.8686 & 0.9426 & 0.9641
& 0.9074 & 0.9954 \\

KEEP
& 0.8788 & 0.8365 & 0.9653
& 0.7769 & 0.7560 & 0.8099
& 0.8685 & 0.9534 & 0.9930
& 0.8851 & 0.9966 \\
\bottomrule
\end{tabular}
}
\end{table*}

\begin{table}[htbp]
\centering
\caption{Performance comparison on disease classification on RenalXJ (deep feature based). Values are mean $\pm$ sample standard deviation across 3 folds. The task involves IgA nephropathy versus membranous nephropathy classification.}
\label{sm_renalxj_disease_deep}
\begin{tabular}{lccc}
\toprule
\textbf{Model} & \textbf{ROC-AUC} & \textbf{F1-score} & \textbf{Accuracy} \\
\midrule
GloSAGE
& $\mathbf{0.9873 \pm 0.0085}$
& $\mathbf{0.9444 \pm 0.0274}$
& $\mathbf{0.9445 \pm 0.0273}$ \\
GloPath
& $0.9777 \pm 0.0077$
& $0.9335 \pm 0.0087$
& $0.9336 \pm 0.0087$ \\
UNI
& $0.9582 \pm 0.0216$
& $0.8925 \pm 0.0378$
& $0.8933 \pm 0.0375$ \\
PLIP
& $0.9123 \pm 0.0102$
& $0.7562 \pm 0.0531$
& $0.7676 \pm 0.0476$ \\
CONCH
& $0.8740 \pm 0.0270$
& $0.7870 \pm 0.0175$
& $0.7905 \pm 0.0171$ \\
KEEP
& $0.8908 \pm 0.0330$
& $0.8056 \pm 0.0299$
& $0.8081 \pm 0.0275$ \\
\bottomrule
\end{tabular}
\end{table}

\begin{table}[htbp]
\centering
\caption{Performance comparison on injury grading on RenalXJ (deep feature based). Values are mean $\pm$ sample standard deviation across 3 folds. The task involves three kidney injury grades (0, 1, and 2).}
\label{sm_renalxj_injury_deep}
\begin{tabular}{lccc}
\toprule
\textbf{Model} & \textbf{ROC-AUC} & \textbf{F1-score} & \textbf{Accuracy} \\
\midrule
GloSAGE
& $\mathbf{0.7358 \pm 0.0310}$
& $\mathbf{0.7054 \pm 0.0078}$
& $\mathbf{0.7352 \pm 0.0193}$ \\
GloPath
& $0.6915 \pm 0.0209$
& $0.6361 \pm 0.0137$
& $0.6718 \pm 0.0393$ \\
UNI
& $0.7013 \pm 0.0358$
& $0.6571 \pm 0.0104$
& $0.7206 \pm 0.0090$ \\
PLIP
& $0.6968 \pm 0.0126$
& $0.6007 \pm 0.0342$
& $0.6974 \pm 0.0036$ \\
CONCH
& $0.6681 \pm 0.0104$
& $0.6312 \pm 0.0539$
& $0.7169 \pm 0.0228$ \\
KEEP
& $0.7080 \pm 0.0457$
& $0.6022 \pm 0.0213$
& $0.7048 \pm 0.0047$ \\
\bottomrule
\end{tabular}
\end{table}

\begin{table}[htbp]
\centering
\caption{Performance comparison on disease classification on RenalXJ (lesion-based).}
\label{sm_renalxj_disease_lesion}
\begin{tabular}{lccc}
\toprule
Model & F1-score & AUC & Acc \\
\midrule
GloSAGE & \textbf{0.8657 $\pm$ 0.0017} & 0.9334 $\pm$ 0.0258 & \textbf{0.8667 $\pm$ 0.0015} \\
GloPath & 0.8432 $\pm$ 0.0585 & \textbf{0.9458 $\pm$ 0.0088} & 0.8448 $\pm$ 0.0559 \\
UNI & 0.8367 $\pm$ 0.0117 & 0.9291 $\pm$ 0.0193 & 0.8377 $\pm$ 0.0118 \\
CONCH & 0.7839 $\pm$ 0.0038 & 0.9004 $\pm$ 0.0185 & 0.7852 $\pm$ 0.0045 \\
PLIP & 0.7703 $\pm$ 0.0353 & 0.8966 $\pm$ 0.0171 & 0.7743 $\pm$ 0.0316 \\
KEEP & 0.7504 $\pm$ 0.0699 & 0.8797 $\pm$ 0.0221 & 0.7632 $\pm$ 0.0549 \\
\bottomrule
\end{tabular}
\end{table}

\begin{table}[htbp]
\centering
\caption{Performance comparison on injury classification on RenalXJ (lesion-based).}
\label{sm_renalxj_injury_lesion}
\begin{tabular}{lcc}
\toprule
Model & AUC & Acc \\
\midrule
GloSAGE & 0.7433 $\pm$ 0.0406 & 0.7259 $\pm$ 0.0231 \\
GloPath & \textbf{0.7690 $\pm$ 0.0609} & 0.6963 $\pm$ 0.0039 \\
UNI & 0.7252 $\pm$ 0.0518 & 0.6974 $\pm$ 0.0044 \\
CONCH & 0.7533 $\pm$ 0.0785 & \textbf{0.7398 $\pm$ 0.0058} \\
PLIP & 0.7322 $\pm$ 0.0674 & 0.6963 $\pm$ 0.0039 \\
KEEP & 0.7238 $\pm$ 0.0546 & 0.6926 $\pm$ 0.0073 \\
\bottomrule
\end{tabular}
\end{table}

\begin{table*}[htbp]
\centering
\caption{Ablation results for lesion recognition on PAS-stained images. Values are reported as mean $\pm$ standard deviation. Bold indicates the highest mean among the five GloSAGE configurations, including ties; the GloPath baseline is excluded from this comparison.}
\label{sm_ablation_recognition_pas}
\resizebox{\textwidth}{!}{
\begin{tabular}{lccccccccccc}
\toprule
Model & MH & PAS($+$) & PAS($-$) & Crum & ADH & SS & MN & MP & EP & Cre & GS \\
\midrule
GloSAGE
& \textbf{0.9296 $\pm$ 0.0048}
& \textbf{0.9539 $\pm$ 0.0019}
& \textbf{1.0000 $\pm$ 0.0000}
& \textbf{0.9220 $\pm$ 0.0239}
& \textbf{0.8146 $\pm$ 0.0139}
& \textbf{0.9673 $\pm$ 0.0038}
& \textbf{0.9574 $\pm$ 0.0014}
& \textbf{0.9685 $\pm$ 0.0031}
& 0.9746 $\pm$ 0.0018
& \textbf{0.9893 $\pm$ 0.0026}
& \textbf{0.9967 $\pm$ 0.0006} \\

w/o EDGE
& 0.9222 $\pm$ 0.0042
& 0.9503 $\pm$ 0.0016
& \textbf{1.0000 $\pm$ 0.0000}
& 0.9034 $\pm$ 0.0254
& 0.7965 $\pm$ 0.0057
& 0.9555 $\pm$ 0.0189
& 0.9476 $\pm$ 0.0077
& 0.9674 $\pm$ 0.0019
& \textbf{0.9749 $\pm$ 0.0012}
& 0.9823 $\pm$ 0.0024
& 0.9943 $\pm$ 0.0018 \\

w/o KEYS
& 0.9226 $\pm$ 0.0049
& 0.9479 $\pm$ 0.0042
& \textbf{1.0000 $\pm$ 0.0000}
& 0.8813 $\pm$ 0.0418
& 0.7955 $\pm$ 0.0138
& 0.9667 $\pm$ 0.0128
& 0.9554 $\pm$ 0.0023
& 0.9666 $\pm$ 0.0034
& 0.9721 $\pm$ 0.0026
& 0.9818 $\pm$ 0.0086
& 0.9964 $\pm$ 0.0005 \\

w/o EDGE and KEYS
& 0.9192 $\pm$ 0.0097
& 0.9416 $\pm$ 0.0080
& \textbf{1.0000 $\pm$ 0.0000}
& 0.8779 $\pm$ 0.0336
& 0.7789 $\pm$ 0.0090
& 0.9558 $\pm$ 0.0217
& 0.9521 $\pm$ 0.0053
& 0.9675 $\pm$ 0.0050
& 0.9688 $\pm$ 0.0070
& 0.9817 $\pm$ 0.0085
& 0.9965 $\pm$ 0.0003 \\

GloSAGE-P
& 0.9260 $\pm$ 0.0082
& 0.9484 $\pm$ 0.0031
& 0.9998 $\pm$ 0.0003
& 0.8685 $\pm$ 0.0823
& 0.7692 $\pm$ 0.0096
& 0.9297 $\pm$ 0.0332
& 0.9464 $\pm$ 0.0085
& 0.9676 $\pm$ 0.0015
& 0.9512 $\pm$ 0.0229
& 0.9814 $\pm$ 0.0072
& 0.9946 $\pm$ 0.0027 \\
\midrule
Vanilla visual pretraining (GloPath)
& 0.909 $\pm$ 0.038
& 0.955 $\pm$ 0.003
& 1.000 $\pm$ 0.000
& 0.887 $\pm$ 0.037
& 0.779 $\pm$ 0.033
& 0.962 $\pm$ 0.009
& 0.949 $\pm$ 0.013
& 0.965 $\pm$ 0.005
& 0.968 $\pm$ 0.015
& 0.980 $\pm$ 0.005
& 0.997 $\pm$ 0.001 \\
\bottomrule
\end{tabular}
}
\end{table*}

\begin{table*}[htbp]
\centering
\caption{Ablation results for lesion recognition on Masson-stained images. Values are reported as mean $\pm$ standard deviation. Bold indicates the highest mean among the five GloSAGE configurations; the GloPath baseline is excluded from this comparison.}
\label{sm_ablation_recognition_masson}
\resizebox{\textwidth}{!}{
\begin{tabular}{lcccccc}
\toprule
Model & MH & ADH & SS & SFN & Cre & GS \\
\midrule
GloSAGE
& 0.9622 $\pm$ 0.0045
& 0.7630 $\pm$ 0.0243
& \textbf{0.9475 $\pm$ 0.0130}
& 0.9859 $\pm$ 0.0035
& \textbf{0.9633 $\pm$ 0.0016}
& \textbf{0.9972 $\pm$ 0.0005} \\

w/o EDGE
& 0.9643 $\pm$ 0.0027
& 0.7174 $\pm$ 0.0711
& 0.9212 $\pm$ 0.0277
& 0.9907 $\pm$ 0.0076
& 0.9535 $\pm$ 0.0268
& 0.9941 $\pm$ 0.0025 \\

w/o KEYS
& 0.9597 $\pm$ 0.0080
& 0.7177 $\pm$ 0.0341
& 0.9409 $\pm$ 0.0272
& \textbf{0.9927 $\pm$ 0.0036}
& 0.9428 $\pm$ 0.0517
& 0.9959 $\pm$ 0.0012 \\

w/o EDGE and KEYS
& \textbf{0.9648 $\pm$ 0.0052}
& \textbf{0.7710 $\pm$ 0.0928}
& 0.9077 $\pm$ 0.0518
& 0.9877 $\pm$ 0.0164
& 0.9491 $\pm$ 0.0602
& 0.9955 $\pm$ 0.0002 \\

GloSAGE-P
& 0.9509 $\pm$ 0.0034
& 0.7324 $\pm$ 0.0883
& 0.7966 $\pm$ 0.0270
& 0.9772 $\pm$ 0.0101
& 0.9345 $\pm$ 0.0351
& 0.9947 $\pm$ 0.0019 \\
\midrule
Vanilla visual pretraining (GloPath)
& 0.961 $\pm$ 0.003
& 0.750 $\pm$ 0.050
& 0.910 $\pm$ 0.054
& 0.982 $\pm$ 0.007
& 0.974 $\pm$ 0.018
& 0.995 $\pm$ 0.003 \\
\bottomrule
\end{tabular}
}
\end{table*}

\newpage

\begin{table*}[!t]
\centering
\caption{Ablation results for lesion recognition on PASM-stained images. Values are reported as mean $\pm$ standard deviation. Bold indicates the highest mean among the five GloSAGE configurations; the GloPath baseline is excluded from this comparison.}
\label{sm_ablation_recognition_pasm}
\resizebox{\textwidth}{!}{
\begin{tabular}{lcccccccccc}
\toprule
Model & MH & PAS($-$) & ADH & SS & SFN & MN & MP & EP & Cre & GS \\
\midrule
GloSAGE
& 0.9281 $\pm$ 0.0012
& \textbf{0.9982 $\pm$ 0.0007}
& 0.7623 $\pm$ 0.0299
& \textbf{0.9609 $\pm$ 0.0102}
& \textbf{0.9857 $\pm$ 0.0017}
& \textbf{0.9456 $\pm$ 0.0119}
& 0.9535 $\pm$ 0.0014
& \textbf{0.9747 $\pm$ 0.0014}
& 0.9880 $\pm$ 0.0004
& \textbf{0.9961 $\pm$ 0.0020} \\

w/o EDGE
& 0.9251 $\pm$ 0.0030
& 0.9974 $\pm$ 0.0007
& 0.7455 $\pm$ 0.0430
& 0.9344 $\pm$ 0.0167
& 0.9789 $\pm$ 0.0032
& 0.9278 $\pm$ 0.0095
& 0.9472 $\pm$ 0.0032
& 0.9635 $\pm$ 0.0210
& 0.9882 $\pm$ 0.0044
& 0.9935 $\pm$ 0.0019 \\

w/o KEYS
& 0.9207 $\pm$ 0.0101
& 0.9964 $\pm$ 0.0020
& 0.7358 $\pm$ 0.0129
& 0.9138 $\pm$ 0.0536
& 0.9838 $\pm$ 0.0045
& 0.9408 $\pm$ 0.0168
& \textbf{0.9543 $\pm$ 0.0093}
& 0.9588 $\pm$ 0.0093
& 0.9884 $\pm$ 0.0007
& 0.9919 $\pm$ 0.0006 \\

w/o EDGE and KEYS
& \textbf{0.9286 $\pm$ 0.0053}
& 0.9974 $\pm$ 0.0013
& \textbf{0.7830 $\pm$ 0.0522}
& 0.9486 $\pm$ 0.0059
& 0.9818 $\pm$ 0.0078
& 0.9438 $\pm$ 0.0033
& 0.9470 $\pm$ 0.0111
& 0.9701 $\pm$ 0.0045
& \textbf{0.9887 $\pm$ 0.0046}
& 0.9932 $\pm$ 0.0017 \\

GloSAGE-P
& 0.9198 $\pm$ 0.0063
& 0.9969 $\pm$ 0.0022
& 0.7466 $\pm$ 0.1000
& 0.8750 $\pm$ 0.0677
& 0.9714 $\pm$ 0.0047
& 0.9253 $\pm$ 0.0063
& 0.9439 $\pm$ 0.0060
& 0.9572 $\pm$ 0.0092
& 0.9790 $\pm$ 0.0041
& 0.9907 $\pm$ 0.0026 \\
\midrule
Vanilla visual pretraining (GloPath)
& 0.928 $\pm$ 0.004
& 0.997 $\pm$ 0.001
& 0.768 $\pm$ 0.038
& 0.945 $\pm$ 0.018
& 0.982 $\pm$ 0.004
& 0.944 $\pm$ 0.008
& 0.955 $\pm$ 0.008
& 0.972 $\pm$ 0.009
& 0.982 $\pm$ 0.010
& 0.995 $\pm$ 0.001 \\
\bottomrule
\end{tabular}
}
\end{table*}

\begin{table}[htbp]
\centering
\caption{Performance of image-conditioned glomerular description generation.
ROUGE-1, ROUGE-2, and ROUGE-L are recall-based overlap scores
against knowledge-graph-derived reference descriptions, averaged
across PAS, Masson, and PASM.}
\label{sm_vqa_all_models}
\begin{tabular}{lccc}
\toprule
Model & ROUGE-1 & ROUGE-2 & ROUGE-L \\
\midrule
GloSAGE & \textbf{0.7906} & \textbf{0.7143} & \textbf{0.7791} \\
PLIP    & 0.7744 & 0.6950 & 0.7660 \\
CONCH   & 0.7859 & 0.7099 & 0.7776 \\
KEEP    & 0.7757 & 0.6963 & 0.7671 \\
\bottomrule
\end{tabular}
\end{table}